\documentclass[10pt,twocolumn,letterpaper]{article}

\usepackage[pagenumbers]{cvpr} % To force page numbers, e.g. for an arXiv version
\usepackage{microtype}

\usepackage{multirow}
\usepackage{booktabs}
\usepackage{longtable}
\usepackage{graphicx}
\usepackage[table,xcdraw]{xcolor}
\usepackage[normalem]{ulem}
\useunder{\uline}{\ul}{}%%
\usepackage{float}
\definecolor{cvprblue}{rgb}{0.21,0.49,0.74}
\usepackage[pagebackref,breaklinks,colorlinks,allcolors=cvprblue]{hyperref}

\def\paperID{*****} % *** Enter the Paper ID here
\def\confName{CVPR}
\def\confYear{2026}

\title{FastMMoE: Accelerating Multimodal Large Language Models through Dynamic Expert Activation and Routing-Aware Token Pruning}

\author{
Guoyang Xia$^{1,2}$\quad
Yifeng Ding$^{2}$\quad
Fengfa Li$^{2}$\quad
Lei Ren$^{2}$\thanks{Project Leader}\quad
Wei Chen$^{2}$\quad
Fangxiang Feng$^{1}$\thanks{Corresponding author}\quad
Xiaojie Wang$^{1}$ \\
$^1$Beijing University of Posts and Telecommunications \quad
$^2$Li Auto \\
{\tt\small \{xiaguoyang, fxfeng\}@bupt.edu.cn \quad \{dingyifeng, renlei3\}@lixiang.com}
}

\begin{document}
\maketitle
\begin{abstract}
Multimodal large language models (MLLMs) have achieved impressive performance, but high-resolution visual inputs result in long sequences of visual tokens and substantial inference latency. Reducing redundant visual tokens is critical to ease computational/memory burdens while preserving performance, enabling MLLM deployment in resource-constrained or latency-sensitive scenarios. Current visual token pruning methods mainly rely on attention-based redundancy analysis and are tailored to dense architectures. We propose \textbf{Fast Multimodal Mixture-of-Experts (FastMMoE)}, a training-free acceleration framework for mixture-of-experts (MoE) based MLLMs, developed from a routing analysis perspective. FastMMoE combines two complementary strategies: (i) expert activation reduction for visual tokens to minimize unnecessary expert computation; and (ii) routing-aware token pruning that leverages similarity in routing probability distributions to identify and remove highly redundant visual tokens. Experiments on large-scale MoE-MLLMs such as DeepSeek-VL2 and InternVL3.5 demonstrate that FastMMoE can reduce FLOPs by up to 55.0\% while retaining approximately 95.5\% of the original performance, consistently outperforming dense-model pruning baselines including FastV and SparseVLM across multiple retention rates. Code is released at https://github.com/MindVLA-Team/FastMMoE.
\end{abstract}    
\section{Introduction}
\label{sec:intro}

With the rapid advancement of large language model (LLM) backbones, multimodal large language models (MLLMs), represented by vision-language models (VLMs), have achieved remarkable performance across a wide spectrum of tasks. Among various LLM architectures, the mixture-of-experts (MoE) paradigm has emerged as a mainstream choice in industrial applications due to its dual advantages of performance and efficiency. This trend has given rise to a surge of MoE-based MLLMs, such as DeepSeek-VL2~\cite{wu2024deepseek}, MolmoE~\cite{deitke2024molmo} and InternVL3.5~\cite{wang2025internvl3}. However, in order to achieve fine-grained vision-language understanding, most high-performing VLMs process images at increasingly higher resolutions, leading to a large number of visual tokens, which significantly reduces inference efficiency. For example, InternVL3.5 adopts a dynamic cropping strategy that splits a high-resolution image into multiple sub-images while also processing a downsampled global thumbnail. This enables the model to jointly capture global context and local details, but processing thousands of visual tokens significantly increases its computational overhead.

% but it results in the processing of thousands of visual tokens—substantially increasing its computational overhead.

A common approach to improve the computational efficiency of VLMs is to analyze visual token redundancy and prune tokens that carry less information, thereby achieving a better trade-off between efficiency and performance. Representative works include FastV~\cite{chen2024image} and SparseVLM~\cite{zhang2024sparsevlm}, which use attention-based analysis to reveal a high level of visual token redundancy. For example, removing up to 50\% of visual tokens leads to only marginal performance degradation. However, existing methods suffer from three limitations: (1) they mainly evaluate on relatively short sequences (e.g., LLaVA-1.5~\cite{liu2024improvedbaselinesvisualinstruction}), making their effectiveness on industrial-scale models with thousands of tokens unclear; (2) they are designed for dense architectures, leaving their applicability to MoE-based MLLMs unexplored; and (3) they do not leverage the inherent properties of MoE models, such as the number of experts activated per token. These limitations motivate us to design a visual token pruning method specifically tailored to industrial-scale MoE-based MLLMs.

In MoE models, pruning redundant visual tokens can be interpreted as setting their expert activation counts to zero. This inspires us to apply a sparser expert activation strategy for visual tokens in MoE-MLLMs, thereby further enhancing their efficiency advantage. Conceptually, this can be seen as assigning fewer experts to “easy” tokens~\cite{huang2024harder}. Our experiments reveal that, starting from the third transformer layer, halving the number of experts activated for all visual tokens—without differentiating among them—has a negligible impact on average model performance. Notably, this requires no additional training data. Through ablation studies, we draw two key observations: (1) experts with lower activation weights tend to carry more redundant information; and (2) visual modality expert outputs exhibit norm concentration, meaning that after re-normalizing outputs following expert reduction, vector norm changes remain small. As a result, the post-reduction outputs only deviate slightly in angular terms from the original outputs—a phenomenon for which we also provide theoretical justifications.

\begin{figure}[!h]
\centering
  \includegraphics[width=\columnwidth,keepaspectratio]{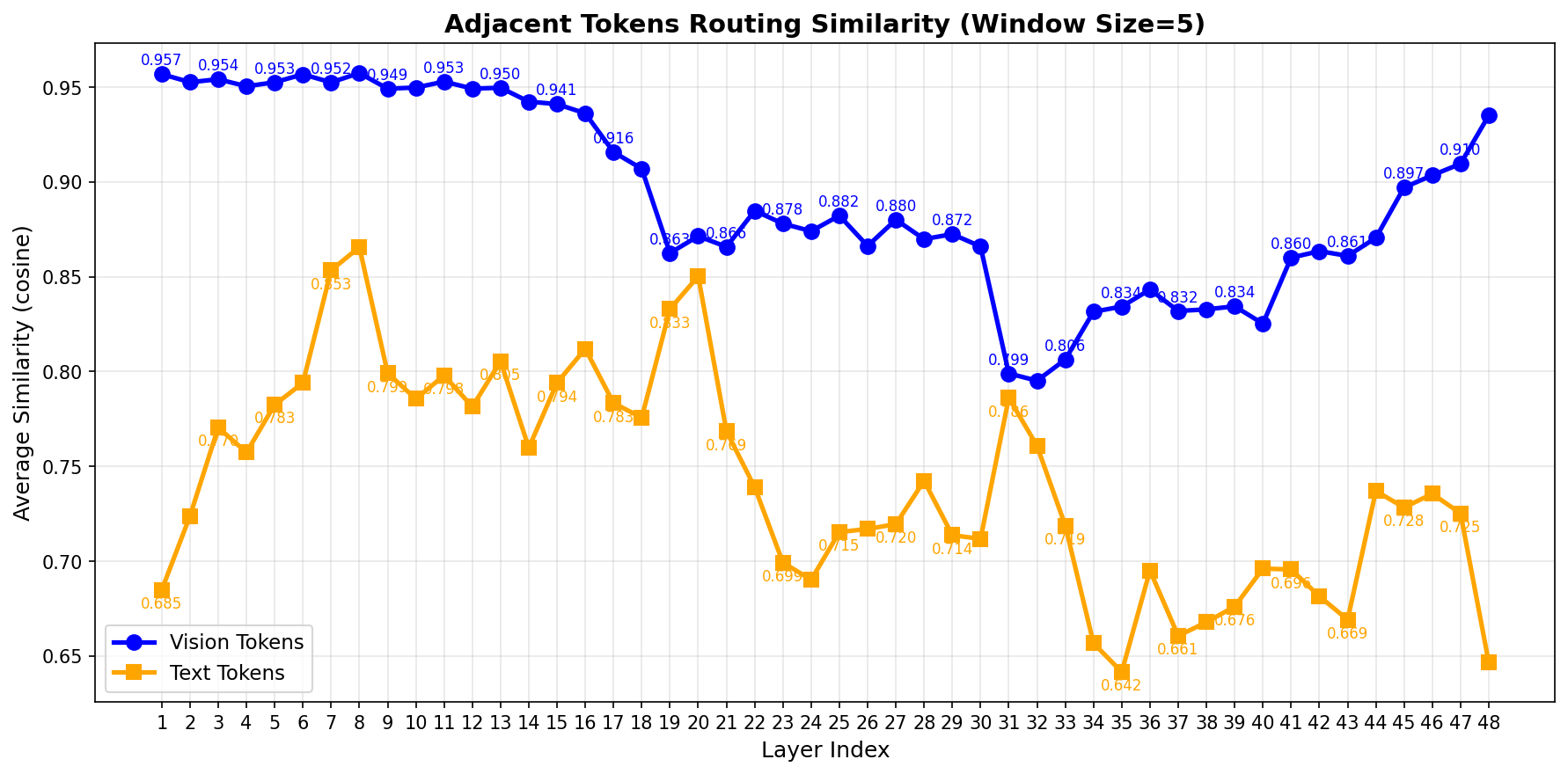}
  \caption{Similarity of adjacent tokens routing probability (Eq.\ref{con:adjacent_sim}) across layers in InternVL3.5.}
  \label{fig:internvl_window_routing_sim}
\end{figure}

Beyond optimizing expert activation for visual tokens, we further exploit the intrinsic nature of MoE routing—computing inner-product similarities between each token and a set of expert “centroids.” Tokens that activate the same experts tend to be more semantically similar. By analyzing the routing probability distributions in  InternVL3.5 and DeepSeek-VL2, we find that adjacent visual tokens have much higher similarity than adjacent text tokens. As shown in Figure~\ref{fig:internvl_window_routing_sim}, in the shallow layers of the Qwen3-30B-A3B backbone within InternVL3.5, the cosine similarity of routing probability distributions between consecutive visual tokens can exceed 95\%. This observation motivates a window-based grouping and fusion strategy for visual token pruning: within a window, higher routing similarity combined with lower attention weights indicates higher redundancy, making the window a strong candidate for removal or merging into a unified “Merged Token.” By incorporating routing distribution similarity into redundancy evaluation, our method consistently outperforms dense-model pruning baselines like FastV and SparseVLM under various retention rates (75\%, 50\%, 25\%).

Our contributions can be summarized as follows:
\begin{itemize}
    \item To the best of our knowledge, we present the first systematic exploration and theoretical analysis of expert activation reduction strategies for visual tokens in MoE-based MLLMs.  
    \item Also, we are the first to incorporate adjacent visual token routing-probability similarity into redundancy evaluation, proposing a novel window-based visual token pruning mechanism.
    \item We demonstrate the effectiveness of our approach on two industrial-scale MoE-based MLLMs, consistently outperforming mainstream dense-model pruning baselines; the combined strategy achieves significant computational savings while retaining most of the original performance.
\end{itemize}
\section{Related Works}
\label{sec:related_work}

\subsection{MoE-based MLLMs}
With the increasing adoption of MoE architectures in industrial-scale LLMs, many follow the multimodal transfer learning paradigm popularized since LLaVA-1.5~\cite{liu2024improvedbaselinesvisualinstruction}. Consequently, a growing number of high-performance MoE-based MLLMs have emerged.  
For example, the MoE version of InternVL3.5~\cite{wang2025internvl3} adopts Qwen3-30B-A3B~\cite{yang2025qwen3} as its backbone. It contains 128 routed experts without shared experts and also activates eight experts per token, aligning with the prevalent fine-grained MoE designs. Its visual encoder is InternViT-300M and retains the Dynamic High-Resolution strategy introduced in InternVL1.5~\cite{chen2024fargpt4vclosinggap}, which likewise results in extremely large numbers of visual tokens.

Similarly, DeepSeek-VL2~\cite{wu2024deepseek} is built on the DeepSeek-MoE~\cite{liu2024deepseek} backbone, which incorporates two shared experts and seventy-two routed experts. For each token, eight experts are activated—including the shared ones—to process its representation. The design intention is to allow shared experts to learn general-purpose knowledge, while routed experts focus on specialized domains. The model employs a SigLIP~\cite{zhai2023sigmoidlosslanguageimage} visual encoder and integrates a dynamic image segmentation strategy~\cite{dai2024nvlm,liu2024llavanext}, potentially leading to  long sequences of visual tokens.

\subsection{Vision Token Pruning for MLLMs}

Visual token pruning is typically guided by cross-modal attention patterns. \textbf{FastV}~\cite{chen2024image} analyzes LLaVA-1.5 and finds that the attention assigned to visual tokens rapidly diminishes in deeper layers—showing high redundancy in visual features. It therefore assumes that self-attention aggregates image representations onto several “anchor” text tokens and prunes low-attention visual tokens accordingly. \textbf{SparseVLM}~\cite{zhang2024sparsevlm} follows a similar idea but first computes cross-modal similarities between text and vision embeddings and applies a clustering-based aggregation to fuse redundant tokens, aiming to reduce information loss.

Other works, such as \textbf{VisionZip}~\cite{yang2025visionzip}, perform early token pruning before the feature stream enters the LLMs. However, most existing methods are designed for dense architectures like LLaVA-1.5 and fail to leverage the characteristics of mixture-of-experts models. Our experiments further show that SparseVLM’s fusion strategy can disturb expert routing in MoE-based MLLMs, highlighting the need for MoE-aware pruning approaches.
% \section{Methods}

% We propose \textbf{FastMMoE}, a training-free acceleration framework tailored for MoE-based MLLMs, which integrates two complementary strategies to reduce inference cost without sacrificing performance. First, we \emph{reduce the number of activated experts} for vision tokens, motivated by the observation that many visual tokens are less critical and can be processed with fewer experts while maintaining accuracy. Second, we design a \emph{routing-aware token pruning} scheme that leverages the high routing-probability similarity among neighboring visual tokens to identify and merge or discard redundant tokens. Both components operate at run time and can be seamlessly integrated into existing MoE-LVLM architectures. An overview of our method is illustrated in Fig.~\ref{fig:overview}, where (i) the expert activation controller sparsifies expert selection for vision tokens from a specified layer onward, and (ii) the sliding-window pruning module computes redundancy scores based on routing similarity and attention importance to perform token merging and removal. In the following, we describe each component in detail.
\section{Methods}

\begin{figure*}
  \centering
  \includegraphics[width=\textwidth,keepaspectratio]{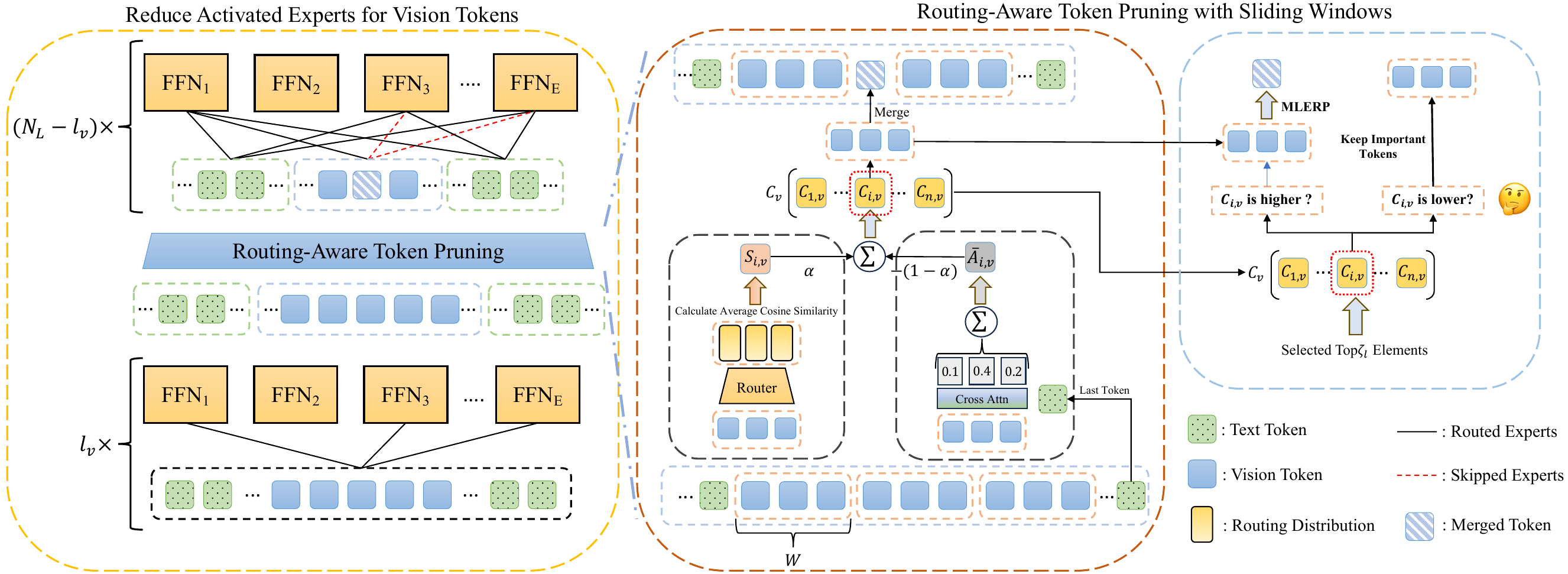}
\caption{
\textbf{FastMMoE overview.} 
(\emph{Left}) Vision-token expert activation reduction: from layer $l_v$ onward, vision tokens (blue) activate fewer experts (red dashed arrows), while text tokens (green) keep full routing. 
(\emph{Right}) Routing-aware token pruning: vision tokens are grouped into sliding windows ($W$), with routing-probability similarity $S_{i,v}$ and attention importance $\bar{A}_{i,v}$ combined into redundancy score $C_v$. 
High-redundancy windows are merged, and low-importance high-redundancy windows are pruned.
}

\label{fig:main_method}
\end{figure*}

We propose \textbf{FastMMoE}, a training-free acceleration framework for MoE-based MLLMs. It combines two complementary strategies to improve inference efficiency while maintaining performance: (1) \emph{reducing activated experts} for vision tokens, based on the observation that many visual tokens can be processed with fewer experts; and (2) a \emph{routing-aware token pruning} mechanism that exploits routing-probability similarity among neighboring vision tokens to merge or discard redundant ones. Both modules operate during inference and can be seamlessly integrated into existing MoE-MLLMs architectures. As shown in \cref{fig:main_method}, the expert activation controller sparsifies vision-token expert selection from a specified layer, while the sliding-window pruning module estimates token redundancy using routing similarity and attention guidance. We detail each component below.

\subsection{Reduce Activated Experts for Vision Tokens}
\label{sec:reduce_activation}
In this section, we present the first core component of FastMMoE—\emph{vision-token expert activation reduction}. 
The method leverages the MoE routing mechanism to selectively activate fewer experts for vision tokens, 
reducing computation without retraining while keeping full expert routing for text tokens. 
We next describe its implementation via routing probability modeling and gated expert selection.

For a single vision token $t_v \in \mathbb{R}^H$ with hidden dimension $H$, its similarity to expert centroid embeddings is computed by the routing network $\mathcal{G}$ as:
\begin{equation}
    Z_v = \mathcal{G}(t_v) = W_E t_v \in \mathbb{R}^{E},
\end{equation}
where $E$ denotes the number of non-shared experts, and $W_E \in \mathbb{R}^{E\times H}$ is the centroid embedding matrix of all experts (the only trainable parameters of the routing network). The corresponding routing probability distribution is then given by:
\begin{equation}
    P_v = \text{softmax}(Z_v) = \frac{\exp(Z_v)}{\sum_{i=1}^{E} \exp(Z_{i,v})} \in \mathbb{R}^E,
\end{equation}
where $Z_{i,v}$ denotes the similarity between the centroid embeddings of expert $i$ and $t_v$, and $P_{i,v}$ represents the routing probability of token $t_v$ to expert $i$.

Let $K$ be the original number of activated experts in the model. We introduce a vision expert activation ratio $0 \le p \le 1$ (if shared experts exist, $p$ can be set to $0$) to control the actual number of experts activated for vision tokens, $K_v = pK$. The $K_v$ experts with the highest routing probabilities are selected, and their index set is defined as:
\begin{equation}
    Q_v = \{h_1, h_2, ..., h_{K_v}\} \subseteq \{1,2,...,E\}.
\end{equation}

In a standard MoE layer, the feed-forward network (FFN) output for the vision token is computed as:
\begin{equation}
    o_v = \sum_{i=1}^{N_s}\text{FFN}^s_i(t_v) + \sum_{j=1}^{E} g_j \text{FFN}^e_j(t_v),
    \label{con:moe_cal}
\end{equation}
where $N_s$ is the number of shared experts, $\text{FFN}^s_i$ denotes the FFN of the $i$-th shared expert, and $\text{FFN}^e_j$ denotes that of the $j$-th non-shared expert. The gating weight $g_j$ is defined as:
\begin{equation}
    g_j=
    \begin{cases}
        \dfrac{P_{j,v}}{\sum_{k\in Q_v} P_{k,v}}, & j \in Q_v,\\[8pt]
        0, & \text{otherwise}.
    \end{cases}
\end{equation}

If $g_j=0$, the corresponding expert is not activated and incurs no computation. This mechanism enables us to assign different numbers of activated experts to vision tokens and text tokens within the same layer. As shown in \cref{fig:main_method}, we further introduce a hyper-parameter $l_v$ to indicate the first layer from which the reduced expert activation strategy for vision tokens is applied (inclusive). Experimental results show that this strategy can substantially reduce computation while preserving most model performance, even without differentiating among vision tokens or performing any retraining.

\subsection{Routing-Aware Token Pruning with Sliding Windows}
\label{sec:routing_aware_pruning}
We now introduce the second core component of FastMMoE—\emph{routing-aware sliding-window vision-token pruning}.
This module follows a three-stage pipeline: 
(1) extract per-token routing probability distributions and cross-modal attention weights for the vision sequence, 
which jointly serve as redundancy indicators; 
(2) segment the vision tokens into sliding windows, and compute within-window routing-probability similarity and attention-based importance, 
yielding a redundancy score for each window; 
and (3) reduce sequence length by merging highly redundant windows to preserve information, 
and removing low-attention redundant windows outright. 
The following three subsections correspond to these stages and provide the detailed methodology.
\subsubsection{Obtaining Routing Distributions and Cross-Modal Attention Weights}
Let the full input sequence be $T = [T_v, T_t]$, where $T_t \in \mathbb{R}^{N_t \times H}$ is the sequence of text tokens and $T_v \in \mathbb{R}^{N_v \times H}$ is the sequence of vision tokens. Using a modality mask, we extract the routing probability distributions for the vision part:
\begin{equation}
    R_v = \text{softmax}(\mathcal{G}(T_v)) \in \mathbb{R}^{N_v \times E},
\end{equation}

We also obtain the cross-modal attention matrix $A_{v,t} \in \mathbb{R}^{N_v \times N_t}$ averaged over all attention heads. Following FastV, we take the attention weights from the last text token to all vision tokens:
\begin{equation}
    \hat{A}_v = A_{v,t}[:, -1] \in \mathbb{R}^{N_v}.
\end{equation}

\subsubsection{Sliding-Window Similarity Analysis}

Given a window size $W$, the sequence of vision tokens is divided into $n = \lceil \frac{N_v}{W} \rceil$ windows, denoted as $\hat{T}_v \in \mathbb{R}^{n \times W \times H}$, where $\hat{T}_{i,v} \in \mathbb{R}^{W\times H}$ denotes the tokens in the $i$-th window. To measure the routing distribution similarity within each window, we compute:
% \begin{equation}
%     S_{i,v} = \frac{1}{\binom{W}{2}} \sum_{a=1}^{W}\sum_{b<a}^{W}\frac{R_{a,v}R_{b,v}^T}{\|R_{a,v}\|\|R_{b,v}\|}, \label{con:adjacent_sim}
% \end{equation}
\begin{equation}
    S_{i,v} = \frac{1}{\binom{W}{2}} \sum_{1 \le b < a \le W} \frac{R_{a,v}R_{b,v}^T}{\|R_{a,v}\|\|R_{b,v}\|}, \label{con:adjacent_sim}
\end{equation}
which gives the average cosine similarity $S_{v,i}$ for the $i$-th window. To reduce complexity, this can be approximated by first computing the average routing distribution $\bar{R}_v$ within the window:
\begin{equation}
    \bar{R}_v = \frac{1}{W} \sum_{i=1}^{W} R_{i,v},
\end{equation}

\begin{equation}
    S_{i,v} = \frac{1}{W}\sum_{a=1}^{W}\frac{R_{a,v}\bar{R}_v^T}{\|R_{a,v}\|\|\bar{R}_v\|}, \quad 
    S_v= [{S}_{1,v},...,{S}_{n,v}].
\end{equation}

The attention importance scores of vision tokens are summed within each window and then normalized:
\begin{equation}
    \bar{A}_{i,v} = \sum_{a=1}^{W} \hat{A}_a, \quad \bar{A}_v = [\bar{A}_{1,v},...,\bar{A}_{n,v}],
\end{equation}
\begin{equation}
    \bar{A}_v = \frac{\bar{A}_v}{\max_j \bar{A}_{j,v}} \in \mathbb{R}^{n}.
\end{equation}

The redundancy score for each window is finally defined as:
\begin{equation}
    C_v = \alpha S_v - (1-\alpha) \bar{A}_v,
\end{equation}
where $ 0 \leq \alpha \leq 1$ is a hyper-parameter controlling the trade-off between routing similarity and attention importance in the redundancy score. In essence, \textbf{windows with high routing similarity and low attention weights are considered more redundant and are better candidates for merging or removal.}

\subsubsection{Window Merging Based on Redundancy Scores}

Let $\eta_l$ be the target number of vision tokens to retain at layer $l$ (typically $\eta_l < N_v$). The number of tokens to be pruned is $\epsilon_l = N_v - \eta_l$. If $\epsilon_l > W$, the target number of windows to select is $\zeta_l = \lfloor \frac{\epsilon_l}{W} \rfloor$. To preserve important information, we define a merging rate $0 < \gamma < 1$, yielding the number of tokens to merge:
\begin{equation}
    m_v^l = \eta_l \gamma. \label{con:merge_rate}
\end{equation}

We select the $m_v^l$ windows with the highest redundancy scores $C_v$ to form the set:
\begin{equation}
    M = \{ c_1, c_2, ..., c_{m_v^l} \} \subseteq \{1, 2, ..., n\}.
\end{equation}
For each $c_i \in M$, the tokens within the window are merged via averaging:
\begin{equation}
    \bar{T}_{c_i,v} = \text{mean}(\hat{T}_{c_i,v}).
\end{equation}

Inspired by MLERP~\cite{kim2024token}, we further normalize the merged tokens to enhance their compatibility with the expert networks:
\begin{equation}
    \bar{x}_{c_i,v} = \frac{1}{W}\sum_{t_{j,v}\in\hat{T}_{c_i,v}} t_{j,v},
\end{equation}
\begin{equation}
    \text{mean}(\hat{T}_{c_i,v}) = \frac{\bar{x}_{c_i,v}}{\|\bar{x}_{c_i,v}\|} \times \|t_{m,v}\|, 
    \quad \|t_{m,v}\| \ge \|t_{j\neq m,v}\in \hat{T}_{c_i,v}\|.
\end{equation}

% \usepackage{booktabs}
% \usepackage{multirow}
% \usepackage{graphicx}
% \usepackage[normalem]{ulem}
\begin{table*}[!t]
\caption{\textbf{Peformance of all methods across different benchmarks for InternVL3.5.} FastMMoE$^*$ denotes that we only apply token pruning method without reducing activated experts. FastMMoE$^\dagger$ denotes that we apply token pruning with reducing activated experts. The best results and second best results are indicated by \textbf{boldface} and {\ul underline}, respectively.}
\label{tab:internvl_main_table}
\centering
\resizebox{\textwidth}{!}{%
\begin{tabular}{@{}lccccccccc@{}}
\toprule
\multicolumn{1}{l|}{\textbf{Method}} &
  \textbf{MMMU} &
  \textbf{SQA$^\text{I}$} &
  \textbf{MMBench} &
  \textbf{OCRBench} &
  \textbf{HallusionBench} &
  \multicolumn{1}{c|}{\textbf{AI2D}} &
  \multicolumn{1}{c|}{\textbf{Avg.$\uparrow$}} &
  \multicolumn{1}{c|}{\textbf{Drop.$\downarrow$}} &
  \textbf{Saving FLOPs. (\%)} \\ \midrule
\multicolumn{1}{l|}{\textbf{Baseline}} &
  \textbf{60.67} &
  \textbf{98.96} &
  \textbf{86.43} &
  \textbf{88.60} &
  \textbf{53.14} &
  \multicolumn{1}{c|}{\textbf{87.14}} &
  \multicolumn{1}{c|}{\textbf{79.16}} &
  \multicolumn{1}{c|}{\textbf{0}} &
  \textbf{0} \\ \midrule
\multicolumn{10}{c}{\cellcolor[HTML]{EFEFEF}\textit{\textbf{Retain 75\% vision tokens after pruning}}} \\
\multicolumn{1}{l|}{FastV} &
  60.22 &
  98.56 &
  {\ul 86.08} &
  79.60 &
  51.55 &
  \multicolumn{1}{c|}{{\ul 85.59}} &
  \multicolumn{1}{c|}{76.93} &
  \multicolumn{1}{c|}{2.22} &
   \\ \cmidrule(r){1-9}
\multicolumn{1}{l|}{SparseVLM} &
  58.89 &
  97.03 &
  83.76 &
  47.30 &
  49.60 &
  \multicolumn{1}{c|}{83.81} &
  \multicolumn{1}{c|}{70.06} &
  \multicolumn{1}{c|}{9.09} &
   \\ \cmidrule(r){1-9}
\multicolumn{1}{l|}{FastMMoE$^*$} &
  {\ul 60.67} &
  \textbf{98.91} &
  \textbf{86.60} &
  \textbf{80.70} &
  {\ul 52.21} &
  \multicolumn{1}{c|}{\textbf{86.27}} &
  \multicolumn{1}{c|}{\textbf{77.56}} &
  \multicolumn{1}{c|}{\textbf{1.60}} &
  \multirow{-3}{*}{21.99} \\ \midrule
\multicolumn{1}{l|}{\begin{tabular}[c]{@{}l@{}}FastMMoE$^\dagger$\\ ($l_v=6, K_v=5$)\end{tabular}} &
  \textbf{61.11} &
  {\ul 98.86} &
  85.82 &
  {\ul 79.90} &
  \textbf{52.31} &
  \multicolumn{1}{c|}{86.37} &
  \multicolumn{1}{c|}{{\ul 77.40}} &
  \multicolumn{1}{c|}{{\ul 1.76}} &
  37.59 \\ \midrule
\multicolumn{1}{l|}{\begin{tabular}[c]{@{}l@{}}FastMMoE$^\dagger$\\ ($l_v=3, K_v=4$)\end{tabular}} &
  60.78 &
  98.86 &
  86.17 &
  79.30 &
  50.49 &
  \multicolumn{1}{c|}{85.46} &
  \multicolumn{1}{c|}{76.84} &
  \multicolumn{1}{c|}{\textbf{2.31}} &
  44.65 \\ \midrule
\multicolumn{10}{c}{\cellcolor[HTML]{EFEFEF}\textit{\textbf{Retain 50\% vision tokens after pruning}}} \\
\multicolumn{1}{l|}{FastV} &
  60.00 &
  97.82 &
  84.28 &
  71.80 &
  49.68 &
  \multicolumn{1}{c|}{82.42} &
  \multicolumn{1}{c|}{74.33} &
  \multicolumn{1}{c|}{4.82} &
   \\ \cmidrule(r){1-9}
\multicolumn{1}{l|}{SparseVLM} &
  57.11 &
  89.94 &
  78.78 &
  28.20 &
  43.92 &
  \multicolumn{1}{c|}{77.36} &
  \multicolumn{1}{c|}{62.55} &
  \multicolumn{1}{c|}{16.60} &
   \\ \cmidrule(r){1-9}
\multicolumn{1}{l|}{FastMMoE$^*$} &
  \textbf{61.00} &
  {\ul 98.36} &
  \textbf{85.22} &
  \textbf{76.50} &
  \textbf{52.54} &
  \multicolumn{1}{c|}{\textbf{84.49}} &
  \multicolumn{1}{c|}{\textbf{76.35}} &
  \multicolumn{1}{c|}{\textbf{2.80}} &
  \multirow{-3}{*}{44.23} \\ \midrule
\multicolumn{1}{l|}{\begin{tabular}[c]{@{}l@{}}FastMMoE$^\dagger$\\ ($l_v=6, K_v=5$)\end{tabular}} &
  {\ul 60.11} &
  \textbf{98.46} &
  {\ul 85.05} &
  {\ul 76.20} &
  {\ul 50.24} &
  \multicolumn{1}{c|}{{\ul 83.71}} &
  \multicolumn{1}{c|}{{\ul 75.63}} &
  \multicolumn{1}{c|}{{\ul 3.53}} &
  55.02 \\ \midrule
\multicolumn{1}{l|}{\begin{tabular}[c]{@{}l@{}}FastMMoE$^\dagger$\\ ($l_v=3, K_v=3$)\end{tabular}} &
  59.33 &
  98.12 &
  84.71 &
  73.00 &
  49.51 &
  \multicolumn{1}{c|}{83.48} &
  \multicolumn{1}{c|}{74.69} &
  \multicolumn{1}{c|}{4.47} &
  64.55 \\ \midrule
\multicolumn{10}{c}{\cellcolor[HTML]{EFEFEF}\textit{\textbf{Retain 25\% vision tokens after pruning}}} \\
\multicolumn{1}{l|}{FastV} &
  57.89 &
  95.29 &
  81.01 &
  57.40 &
  46.51 &
  \multicolumn{1}{c|}{77.14} &
  \multicolumn{1}{c|}{69.21} &
  \multicolumn{1}{c|}{9.95} &
   \\ \cmidrule(r){1-9}
\multicolumn{1}{l|}{SparseVLM} &
  53.89 &
  81.76 &
  67.44 &
  13.60 &
  39.57 &
  \multicolumn{1}{c|}{72.12} &
  \multicolumn{1}{c|}{54.73} &
  \multicolumn{1}{c|}{24.43} &
   \\ \cmidrule(r){1-9}
\multicolumn{1}{l|}{FastMMoE$^*$} &
  \textbf{60.78} &
  \textbf{96.98} &
  \textbf{83.93} &
  \textbf{66.80} &
  \textbf{50.52} &
  \multicolumn{1}{c|}{\textbf{82.12}} &
  \multicolumn{1}{c|}{\textbf{73.52}} &
  \multicolumn{1}{c|}{\textbf{5.63}} &
  \multirow{-3}{*}{64.30} \\ \midrule
\multicolumn{1}{l|}{\begin{tabular}[c]{@{}l@{}}FastMMoE$^\dagger$\\ ($l_v=6, K_v=5$)\end{tabular}} &
  {\ul 59.00} &
  {\ul 96.43} &
  {\ul 83.25} &
  {\ul 63.20} &
  {\ul 48.06} &
  \multicolumn{1}{c|}{{\ul 81.38}} &
  \multicolumn{1}{c|}{{\ul 71.89}} &
  \multicolumn{1}{c|}{{\ul 7.27}} &
  70.47 \\ \bottomrule
\end{tabular}%
}
\end{table*}

The positions of the merged tokens remain unchanged. The remaining number of tokens to drop is $u_l = \epsilon_l - m_v^l > 0$. For the remaining windows, we choose the $\lfloor \frac{u_l}{W} \rfloor$ windows with the smallest $\bar{A}_v$ and discard all their tokens, as they are both less important and lack sufficient routing similarity.

\section{Experiments}
\label{sec:Experiments}
\subsection{Evaluation Details}

We evaluate our model across six public multimodal benchmarks: \textbf{MMMU}~\cite{yue2024mmmu} and \textbf{ScienceQA}~\cite{lu2022learnexplainmultimodalreasoning} assess cross-domain reasoning and multimodal understanding, while \textbf{MMBench}~\cite{liu2024mmbench} focuses on visual grounding and multi-round reasoning. \textbf{OCRBench}~\cite{liu2024ocrbench} measures text recognition ability on real-world visual-text inputs.

Additionally, \textbf{HallusionBench}~\cite{guan2024hallusionbench} and \textbf{AI2D}~\cite{kembhavi2016diagram} evaluate logical consistency and complex visual reasoning—testing robustness against hallucination and diagram understanding. Together, these benchmarks comprehensively reflect the general performance and applicability of \textbf{FastMMoE} across multimodal perception and reasoning tasks. 
All evaluations are performed using the \textbf{VLMEvalKit}~\cite{duan2025vlmevalkitopensourcetoolkitevaluating}.

\subsection{Implementation Details}

We conduct experiments on the MoE-based version of \textbf{InternVL3.5} and the \textbf{DeepSeek-VL2} models. For a fair comparison, we re-implemented the common visual token pruning baselines, including \textbf{FastV} and \textbf{SparseVLM}, under the same conditions. 
The hyperparameters of \textbf{FastMMoE} and other important details will be provided in the Appendix~\ref{sec:imple_ap}. 

% Inference is run on multiple NVIDIA A800 GPUs, where only the expert activation strategy and visual token processing modules are modified while keeping other components unchanged.

\subsection{Computing Cost Estimation}
% The FLOPs saving ratio of pruning tokens alone can be computed simply by:
% \begin{multline}
%     \mathcal{R} = 1 - [\frac{l(4N_v^2H+8L_vH^2+2N_vHE+6N_vHS_mK)}{N_L(4N_v^2H+8L_vH^2+2N_vHE+6N_vHS_mK)}+\\
% \frac{(N_L-l)(4\eta_l^2H+8\eta_lH^2+2\eta_lHE+6\eta_lHS_mK)}{N_L(4N_v^2H+8L_vH^2+2N_vHE+6N_vHS_mK)}]
% \end{multline}
% where $S_m$ is the experts intermediate size.

We estimate the theoretical FLOPs saving ratio for FastMMoE by analytically modeling the cost of token pruning and expert activation reduction. Due to the length and complexity of the formulation, we provide the detailed derivation and the final closed-form expressions in the Appendix~\ref{sec:flops}. This analysis confirms the computational efficiency trends observed in our experiments.

\subsection{Results}

\paragraph{Results on InternVL3.5.} 
\cref{tab:internvl_main_table} summarizes the performance of all methods evaluated on the MoE-based \textbf{InternVL3.5}. 
Across all token retention levels, \textbf{FastMMoE} consistently outperforms dense-model-oriented baselines such as \textbf{FastV} and \textbf{SparseVLM}. 
In particular, under the 75\% vision token retention setting with expert activation reduction ($l_v=3, K_v=4$), our method achieves a \textbf{44.65\% reduction in FLOPs}—comparable to directly pruning 50\% of  visual tokens—yet reaches an average score of \textbf{76.84}, which surpasses all 50\% retention methods (the best baseline achieves 76.35). 
This observation demonstrates that by appropriately reducing the number of activated experts, FastMMoE can achieve substantial inference acceleration even at higher token retention levels, while maintaining superior accuracy.

At more aggressive pruning configurations (50\% and 25\% retention), 
\textbf{FastMMoE} maintains leading performance with only minor degradation (average drops of 2.80 and 5.63, respectively), 
while effectively reducing computational costs by \textbf{44.23\%} and \textbf{64.30\%}. 
In contrast, \textbf{SparseVLM} severely deteriorates under low retention, especially on OCRBench where text information is easily lost due to na\"ive token merging. 
These results validate the robustness and efficiency of our routing-aware pruning framework.

\paragraph{Results on DeepSeek-VL2.}
\cref{tab:deepseek_vl2_performance} reports the averaged performance comparison on \textbf{DeepSeek-VL2}. 
The same pattern can be observed—\textbf{FastMMoE} consistently achieves smaller performance degradation while providing significant computational savings. 
When retaining 75\%, 50\%, and 25\% of visual tokens, FastMMoE attains average score drops of only \textbf{0.19}, \textbf{0.74}, and \textbf{2.36}, respectively, with corresponding FLOPs reductions of \textbf{21.09\%}, \textbf{43.07\%}, and \textbf{63.66\%}. 
Even at the most aggressive pruning ratio, our method remains clearly ahead of both FastV and SparseVLM, confirming its generalizability across different scales and architectures of MoE-MLLMs. 

\paragraph{Summary.}
Overall, the experimental results demonstrate that \textbf{FastMMoE} effectively exploits MoE-specific properties to achieve high inference efficiency while preserving accuracy. 
By jointly combining expert activation reduction and routing-aware token pruning, it attains a better trade-off between computation and performance across both InternVL3.5 and DeepSeek-VL2, consistently outperforming previous dense-model pruning baselines.

% \usepackage{multirow}
% \usepackage{graphicx}
% \usepackage[normalem]{ulem}
\begin{table}[!t]
\centering
\caption{\textbf{Peformance of all methods for DeepSeek-VL2.} More details are provided in the Appendix~\ref{sec:test_results}.}
\label{tab:deepseek_vl2_performance}
\resizebox{\columnwidth}{!}{%
\begin{tabular}{@{}lccc@{}}
\toprule
\textbf{Method}                        & \textbf{Avg.$\uparrow$}             & \textbf{Drop.$\downarrow$}         & \textbf{Saving FLOPs. (\%)} \\ \midrule
\multicolumn{1}{l|}{\textbf{Baseline}} & \multicolumn{1}{c|}{\textbf{72.68}} & \multicolumn{1}{c|}{\textbf{0}}    & \textbf{0}                  \\ \midrule
\multicolumn{4}{c}{\cellcolor[HTML]{EFEFEF}\textit{\textbf{Retain 75\% vision tokens after pruning}}}                                           \\
\multicolumn{1}{l|}{FastV}             & \multicolumn{1}{c|}{{\ul 72.18}}    & \multicolumn{1}{c|}{{\ul 0.5}}     &                             \\ \cmidrule(r){1-3}
\multicolumn{1}{l|}{SparseVLM}         & \multicolumn{1}{c|}{71.72}          & \multicolumn{1}{c|}{0.96}          &                             \\ \cmidrule(r){1-3}
\multicolumn{1}{l|}{FastMMoE$^*$}      & \multicolumn{1}{c|}{\textbf{72.49}} & \multicolumn{1}{c|}{\textbf{0.19}} & \multirow{-3}{*}{21.09}     \\ \midrule
\multicolumn{4}{c}{\cellcolor[HTML]{EFEFEF}\textit{\textbf{Retain 50\% vision tokens after pruning}}}                                           \\
\multicolumn{1}{l|}{FastV}             & \multicolumn{1}{c|}{{\ul 71.91}}    & \multicolumn{1}{c|}{{\ul 0.77}}    &                             \\ \cmidrule(r){1-3}
\multicolumn{1}{l|}{SparseVLM}         & \multicolumn{1}{c|}{70.31}          & \multicolumn{1}{c|}{2.37}          &                             \\ \cmidrule(r){1-3}
\multicolumn{1}{l|}{FastMMoE$^*$}      & \multicolumn{1}{c|}{\textbf{71.94}} & \multicolumn{1}{c|}{\textbf{0.74}} & \multirow{-3}{*}{43.07}     \\ \midrule
\multicolumn{4}{c}{\cellcolor[HTML]{EFEFEF}\textit{\textbf{Retain 25\% vision tokens after pruning}}}                                           \\
\multicolumn{1}{l|}{FastV}             & \multicolumn{1}{c|}{{\ul 70.02}}    & \multicolumn{1}{c|}{{\ul 2.66}}    &                             \\ \cmidrule(r){1-3}
\multicolumn{1}{l|}{SparseVLM}         & \multicolumn{1}{c|}{66.61}          & \multicolumn{1}{c|}{6.07}          &                             \\ \cmidrule(r){1-3}
\multicolumn{1}{l|}{FastMMoE$^*$}      & \multicolumn{1}{c|}{\textbf{70.32}} & \multicolumn{1}{c|}{\textbf{2.36}} & \multirow{-3}{*}{63.66}     \\ \bottomrule
\end{tabular}%
}
\end{table}
\section{Ablation Study \& Analysis}
\label{sec:ablation}
\subsection{Strategy of Reducing Activated Experts}

Although the sparse expert activation strategy for vision tokens is conceptually simple, it is necessary to systematically investigate how the starting layer for activation reduction and the number of retained experts affect model performance. We conduct extensive experiments on both InternVL3.5 and DeepSeek-VL2, showing a consistent trend: the later the reduction begins and the more experts are retained, the better the model performance is preserved.  

\begin{figure}[!h]
\centering
  \includegraphics[width=\columnwidth,keepaspectratio]{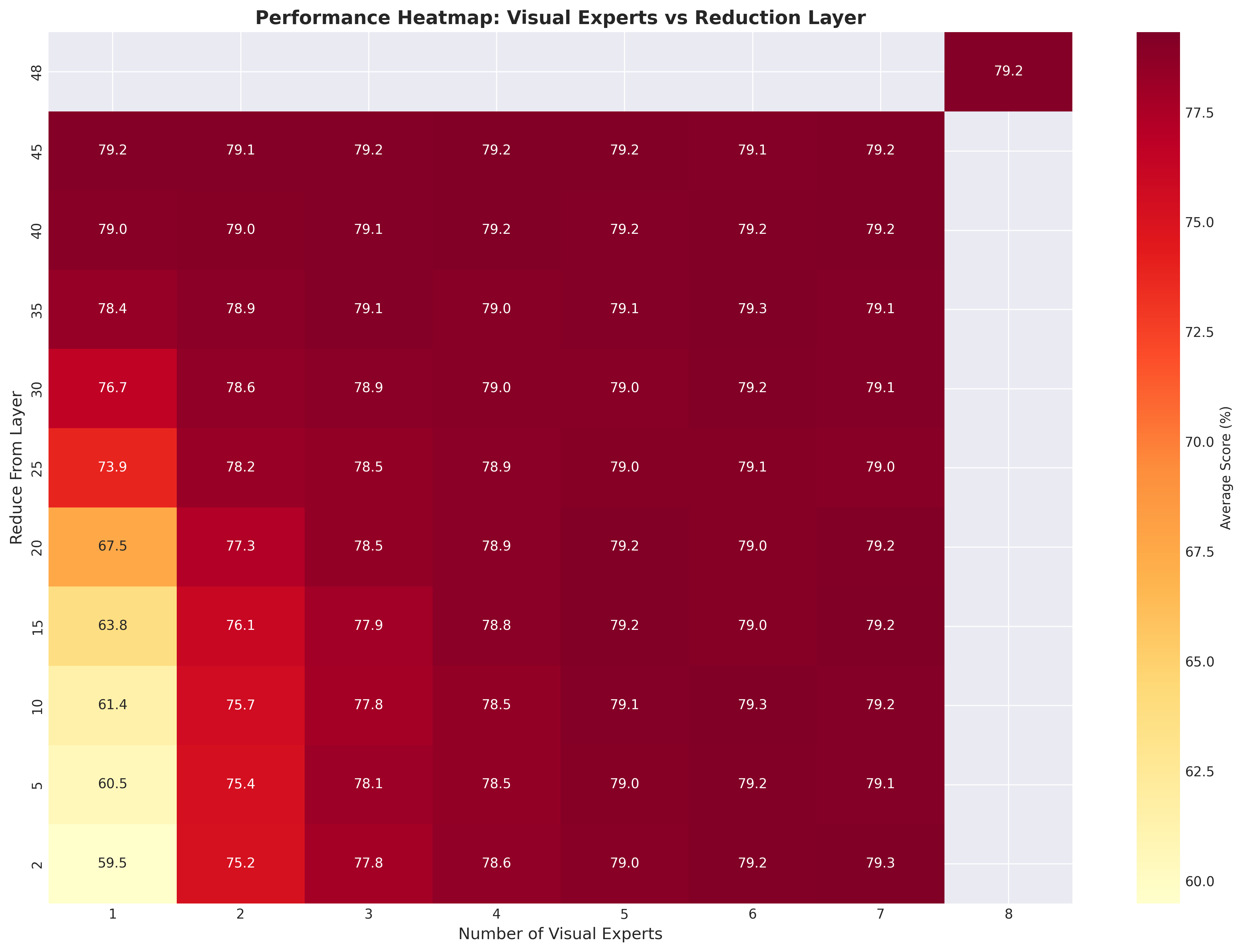}
  \caption{\textbf{Average Performance Heatmap for InternVL3.5.} We test the different choices of $l_v,K_v$ to reduce activated experts for vision tokens. More detailes are provided in the Appendix~\ref{sec:test_results}.}
  \label{fig:act_perf_internvl}
\end{figure}

\begin{figure}[!h]
\centering
  \includegraphics[width=\columnwidth,keepaspectratio]{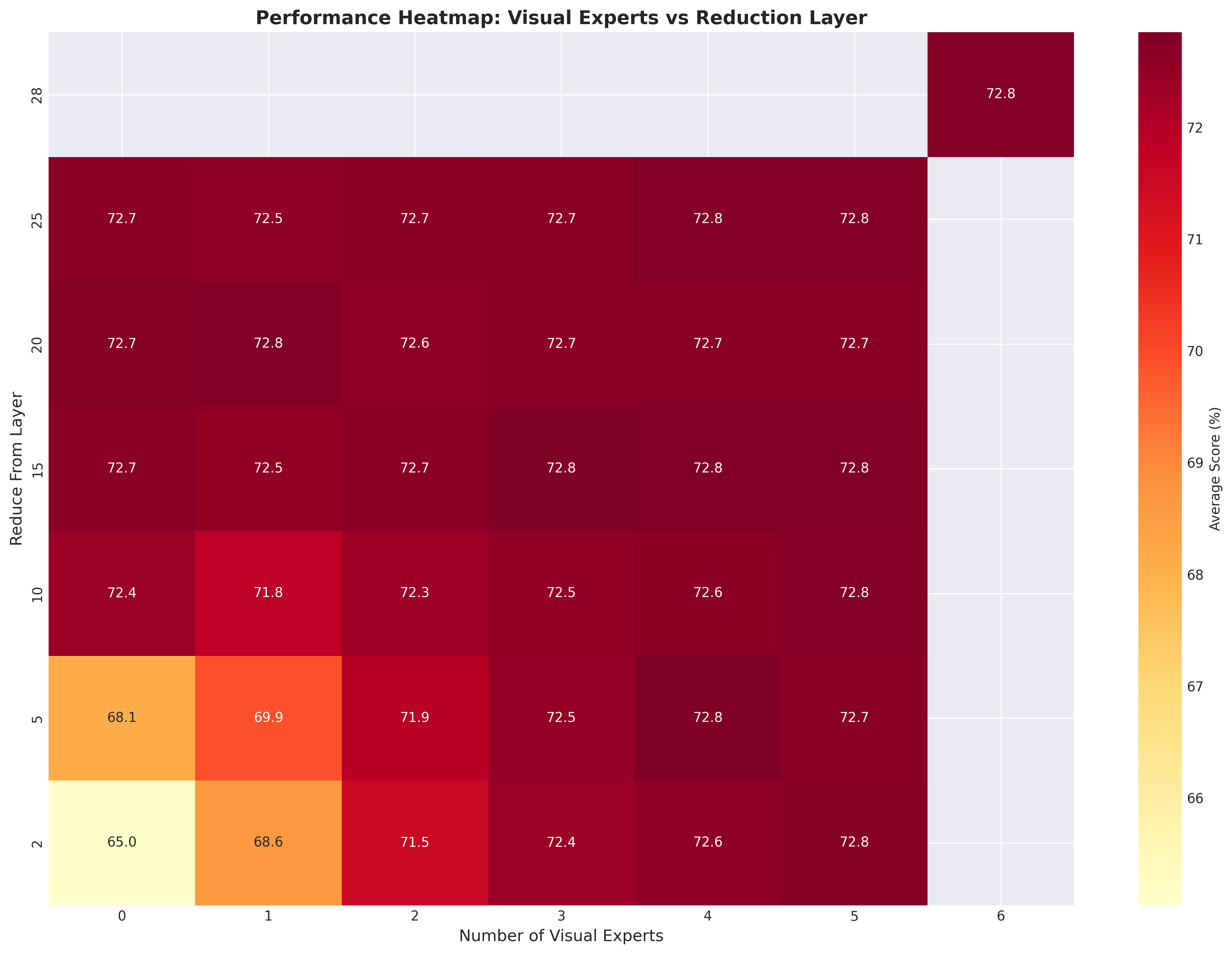}
  \caption{\textbf{Average Performance Heatmap for DeepSeek-VL2.} More detailes are provided in the Appendix~\ref{sec:test_results}.}
  \label{fig:act_perf_deepseek}
\end{figure}

Specifically, for \textbf{InternVL3.5}, even when the number of activated experts for vision tokens is halved starting from the second layer, the model still retains about \textbf{99.2\%} of its original performance, as shown in \cref{fig:act_perf_internvl}. Similarly, for \textbf{DeepSeek-VL2}, applying expert reduction starting from the tenth layer and keeping only the shared experts allows the model to maintain up to \textbf{99.4\%} of its initial performance, as shown in \cref{fig:act_perf_deepseek}. Remarkably, these results are achieved without any additional training, indicating that current vision tokens may not fully exploit the available experts in MoE-based MLLMs. This observation suggests that the existing training paradigm of MoE-MLLMs leaves ample room for refinement in modality-specific expert utilization.

\subsection{Reduce Activation across Different Modalities}

% \usepackage{booktabs}
% \usepackage{graphicx}
\begin{table}[!h]
\centering
\caption{Performance of reducing activation across different modalities on InternVL3.5.}
\label{tab:reduce_modal_ablation}
\resizebox{\columnwidth}{!}{%
\begin{tabular}{@{}r|cccc|c@{}}
\toprule
\multicolumn{1}{c|}{\textbf{Target Modality}} & \textbf{MMMU} & \textbf{MMBench} & \textbf{OCRBench} & \textbf{AI2D} & \textbf{Avg.$\uparrow$} \\ \midrule
Vision($l_v=3, K_v=4$) & 60.11 & 86.34 & 87.20 & 86.46 & 80.03 \\
Text($l_v=3, K_v=4$)   & 57.00 & 83.51 & 86.30 & 85.10 & 77.98 \\
All($l_v=3, K_v=4$)    & 56.89 & 83.59 & 84.10 & 84.23 & 77.20 \\ \bottomrule
\end{tabular}%
}
\end{table}

We further investigate the effect of applying expert activation reduction across different modalities. Specifically, we apply the same reduction configuration ($l_v=3, K_v=4$) to three settings: vision tokens only, text tokens only, and all tokens simultaneously.  
As shown in \cref{tab:reduce_modal_ablation}, reducing activation for text tokens leads to a larger performance drop compared to applying it only to vision tokens, while applying the strategy to all tokens results in the worst performance. This indicates that within MoE-MLLM architectures, semantic alignment and reasoning are primarily guided by text tokens, which exhibit higher expert utilization rates than vision tokens.

\subsection{Selection Strategy for Expert Activation Reduction}

% \begin{table}[!h]
% \centering
% \caption{\textbf{Ablation on expert selection strategies for activation reduction on InternVL3.5.} 
% We compare three strategies for selecting $K_v$ activated experts: 
% \textbf{TopK} (selecting experts with the highest routing weights), 
% \textbf{RandomK} (randomly selecting experts), 
% and \textbf{MinK} (selecting experts with the lowest routing weights). 
% TopK achieves the best performance, indicating that highly weighted experts contribute the most, while low-weight experts introduce redundancy.}
% \label{tab:reduce_stategy}
% \resizebox{\columnwidth}{!}{%
% \begin{tabular}{@{}c|c|c|cccc|c@{}}
% \toprule
% \textbf{$K_v$} & \textbf{$l_v$} & \textbf{Strategy} & \textbf{MMMU} & \textbf{MMBench} & \textbf{OCRBench} & \textbf{AI2D} & \textbf{Avg.$\uparrow$} \\ \midrule
% 4 & 2  & TopK    & 60.11 & 86.34 & 87.20 & 86.46 & 80.03 \\
% 4 & 24 & TopK    & 61.00 & 87.03 & 87.50 & 86.56 & \textbf{80.52} \\
% 4 & 2  & RandomK & 60.11 & 84.28 & 85.40 & 86.31 & 79.02 \\
% 4 & 24 & RandomK & 60.00 & 85.74 & 86.10 & 86.88 & 79.68 \\
% 4 & 2  & MinK    & 57.11 & 81.87 & 74.80 & 82.87 & 74.16 \\
% 4 & 24 & MinK    & 60.44 & 85.74 & 83.80 & 86.59 & 79.14 \\ \bottomrule
% \end{tabular}%
% }
% \end{table}

\begin{table}[!h]
\centering
\caption{\textbf{Ablation on expert selection strategies for activation reduction on InternVL3.5.} 
We compare three strategies for selecting $K_v$ activated experts: 
% \textbf{TopK} (selecting experts with the highest routing weights), 
% \textbf{RandomK} (randomly selecting experts), 
% and \textbf{MinK} (selecting experts with the lowest routing weights). 
TopK achieves the best performance, indicating that highly weighted experts contribute the most, while low-weight experts introduce redundancy.}
\label{tab:reduce_stategy}
\resizebox{\columnwidth}{!}{%
\begin{tabular}{@{}c|c|c|cccc|c@{}}
\toprule
\textbf{$K_v$} & \textbf{Strategy} & \textbf{$l_v$} & \textbf{MMMU} & \textbf{MMBench} & \textbf{OCRBench} & \textbf{AI2D} & \textbf{Avg.$\uparrow$} \\ \midrule
\multirow{6}{*}{4} & \multirow{2}{*}{TopK}    & 2  & 60.11 & 86.34 & 87.20 & 86.46 & 80.03 \\ \cmidrule(l){3-8} 
                   &                          & 24 & 61.00 & 87.03 & 87.50 & 86.56 & 80.52 \\ \cmidrule(l){2-8} 
                   & \multirow{2}{*}{RandomK} & 2  & 60.11 & 84.28 & 85.40 & 86.31 & 79.02 \\ \cmidrule(l){3-8} 
                   &                          & 24 & 60.00 & 85.74 & 86.10 & 86.88 & 79.68 \\ \cmidrule(l){2-8} 
                   & \multirow{2}{*}{MinK}    & 2  & 57.11 & 81.87 & 74.80 & 82.87 & 74.16 \\ \cmidrule(l){3-8} 
                   &                          & 24 & 60.44 & 85.74 & 83.80 & 86.59 & 79.14 \\ \bottomrule
\end{tabular}%
}
\end{table}

To verify whether all activated experts contribute equally, we conduct an ablation study on different expert selection strategies using InternVL3.5. 
\textbf{TopK} selects the $K_v$ experts with the highest routing weights, 
\textbf{RandomK} randomly selects $K_v$ experts, 
and \textbf{MinK} uses the $K_v$ experts with the smallest weights. 
As shown in \cref{tab:reduce_stategy}, the TopK strategy consistently achieves the highest scores across benchmarks, followed by RandomK, while MinK performs the worst. 
This indicates that experts with smaller routing weights contribute less to overall performance, confirming that expert contributions are roughly proportional to their routing weights.

\subsection{Vision Tokens Exhibit Magnitude Concentration}

We find that, across most decoder layers, vision tokens exhibit significantly higher magnitude stability than text tokens, as measured by the stability score derived from expert output norms. This indicates that expert outputs for vision tokens have more concentrated magnitudes, meaning that activation reduction primarily affects vector directions rather than norms. Such norm consistency explains why reducing over half of the activated experts for vision tokens has a minimal impact on performance (see the Appendix~\ref{sec:magnitude_concentration} for a detailed derivation and formulas).

\subsection{Merge Methods for Redundant Tokens}

% \usepackage{booktabs}
% \usepackage{multirow}
% \usepackage{graphicx}
\begin{table}[!h]
\centering
\caption{Comparison among different token merge methods on InternVL3.5.}
\label{tab:merge_method}
\resizebox{\columnwidth}{!}{%
\begin{tabular}{@{}c|c|cccc|c@{}}
\toprule
\textbf{Retain Ratio} & \textbf{Method} & \textbf{MMMU} & \textbf{MMBench} & \textbf{OCRBench} & \textbf{AI2D} & \textbf{Avg.$\uparrow$} \\ \midrule
\multirow{2}{*}{75\%} & mean  & 60.44 & 86.34 & 80.40 & 86.40 & 78.40 \\ \cmidrule(l){2-7} 
                      & mlerp & 60.67 & 86.60 & 80.70 & 86.27 & 78.56 \\ \midrule
\multirow{2}{*}{50\%} & mean  & 60.33 & 85.31 & 76.80 & 84.87 & 76.83 \\ \cmidrule(l){2-7} 
                      & mlerp & 61.00 & 85.22 & 76.50 & 84.49 & 76.80 \\ \midrule
\multirow{2}{*}{25\%} & mean  & 60.89 & 83.76 & 67.00 & 81.87 & 73.38 \\ \cmidrule(l){2-7} 
                      & mlerp & 60.78 & 83.93 & 66.80 & 82.12 & 73.41 \\ \bottomrule
\end{tabular}%
}
\end{table}

To evaluate whether the \textbf{MLERP}~\cite{kim2024token} token merging method outperforms simple arithmetic averaging (\textbf{mean}), we compare both approaches under different visual token retention rates, as shown in \cref{tab:merge_method}.  
At the 75\% retention level, MLERP delivers a clear advantage over mean averaging. At 50\%, the difference between the two methods is negligible, whereas at 25\%, MLERP shows a slight improvement. Overall, MLERP consistently outperforms mean averaging at 75\% and 25\% retention, which motivates our choice of MLERP as the default token merging scheme in \textbf{FastMMoE}.

\subsection{Balance Between Routing Similarity and Attention Weights}

\begin{table}[!h]
\centering
\caption{Camparison among different $\alpha$ value on InternVL3.5.}
\label{tab:alpha_ablation}
\resizebox{\columnwidth}{!}{%
\begin{tabular}{@{}c|c|cccc|c@{}}
\toprule
\textbf{Retain Ratio} & \textbf{$\alpha$} & \textbf{MMMU}  & \textbf{MMBench} & \textbf{OCRBench} & \textbf{AI2D}  & \textbf{Avg.$\uparrow$} \\ \midrule
\multirow{3}{*}{75\%} & 0.6 & 60.33          & 86.17          & 79.70          & 86.20          & 78.10          \\ \cmidrule(l){2-7} 
                      & 0.5 & \textbf{60.67} & \textbf{86.60} & \textbf{80.70} & \textbf{86.27} & \textbf{78.56} \\ \cmidrule(l){2-7} 
                      & 0.4 & 60.78          & 86.25          & 80.50          & 86.20          & 78.43          \\ \midrule
\multirow{3}{*}{50\%} & 0.7 & 60.00          & 85.31          & 75.30          & 84.49          & 76.27          \\ \cmidrule(l){2-7} 
                      & 0.6 & \textbf{61.00} & \textbf{85.22} & \textbf{76.50} & \textbf{84.49} & \textbf{76.80} \\ \cmidrule(l){2-7} 
                      & 0.5 & 59.56          & 85.65          & 76.00          & 84.59          & 76.45          \\ \midrule
\multirow{3}{*}{25\%} & 1                 & \textbf{60.78} & \textbf{83.93}   & \textbf{66.80}    & \textbf{82.12} & \textbf{73.41} \\ \cmidrule(l){2-7} 
                      & 0.9 & 60.00          & 83.33          & 64.20          & 81.02          & 72.14          \\ \cmidrule(l){2-7} 
                      & 0.8 & 59.44          & 82.99          & 63.80          & 81.09          & 71.83          \\ \bottomrule
\end{tabular}%
}
\end{table}

\cref{tab:alpha_ablation} illustrates the effect of the weighting parameter $\alpha$, which controls the balance between routing-probability similarity and cross-modal attention weights in the redundancy computation. Across all retention levels, a clear pattern emerges: overly small or large $\alpha$ values degrade performance, while moderate settings yield the best results. 
For instance, at a 75\% retention rate, the optimal $\alpha$ is 0.5 (average score 78.56), while at 50\% retention the best performance occurs at $\alpha=0.6$ (76.80). Under a more aggressive 25\% retention, routing similarity becomes more dominant, and $\alpha=1.0$ achieves the highest score (73.41). 
These results indicate that routing similarity and attention guidance play complementary roles—balanced weighting benefits mild pruning, whereas stronger reliance on routing similarity favors heavy compression. More details are provided in the Appendix~\ref{sec:test_results}.

\subsection{Ablation Study on Merge Rate}

\begin{table}[!h]
\centering
\caption{\textbf{Performance under different merge rates $\gamma$} on InternVL3.5.}
%Values are reported for multiple retain ratios (75\%, 50\%, 25\%) with MMMU, MMBench, OCRBench, AI2D as evaluation tasks. Avg.$\uparrow$ denotes the average score; best results per setting are in \textbf{bold}.
\label{tab:merge_ratio_ablation}
\resizebox{\columnwidth}{!}{%
\begin{tabular}{@{}c|c|cccc|c@{}}
\toprule
\textbf{Retain Ratio} & \textbf{$\gamma$} & \textbf{MMMU}  & \textbf{MMBench} & \textbf{OCRBench} & \textbf{AI2D}  & \textbf{Avg.$\uparrow$} \\ \midrule
75\%                  & 0.025 & \textbf{60.67} & \textbf{86.60} & \textbf{80.70} & \textbf{86.27} & \textbf{78.56} \\ \midrule
\multirow{2}{*}{50\%} & 0.05              & \textbf{61.00} & \textbf{85.22}   & \textbf{76.50}    & \textbf{84.49} & \textbf{76.80}          \\ \cmidrule(l){2-7} 
                      & 0.025 & 60.44          & 85.05          & 74.10          & 84.26          & 75.96          \\ \midrule
\multirow{4}{*}{25\%} & 0.15  & 59.33          & 83.76          & 69.20          & 80.99          & 73.32          \\ \cmidrule(l){2-7} 
                      & 0.1   & \textbf{60.78} & \textbf{83.93} & \textbf{66.80} & \textbf{82.12} & \textbf{73.41} \\ \cmidrule(l){2-7} 
                      & 0.05  & 59.00          & 82.73          & 62.30          & 80.44          & 71.12          \\ \cmidrule(l){2-7} 
                      & 0.025 & 59.44          & 82.56          & 60.00          & 78.72          & 70.18          \\ \bottomrule
\end{tabular}%
}
\end{table}

We investigate the influence of the merge rate $\gamma$, which controls the proportion of retained vision tokens produced by merging high-redundancy windows.  
Details on the theoretical upper bounds of $\gamma$ under different token retention ratios, as well as the multi-stage pruning formulation, are provided in Appendix~\ref{sec:gamma_theory}.  

In our ablation (\cref{tab:merge_ratio_ablation}), $\gamma$ is selected according to these bounds, with 75\% retention testing only its maximal feasible value, 50\% retention evaluating two settings, and 25\% retention covering a wider range to reflect heavier pruning.  
Results show that very small $\gamma$ tends to underutilize merging opportunities, whereas excessively large values may over-compress and lose fine-grained details.  
Moderate merge rates close to the theoretical upper bound generally achieve the best trade-off between token compression and semantic fidelity, which motivates our adoption of lightweight merging configurations as the default in \textbf{FastMMoE}.

\section{Conclusion}

We propose \textbf{FastMMoE}, a training-free acceleration framework for MoE-based multimodal large language models. 
FastMMoE jointly integrates two complementary strategies: (1) a vision-token-oriented expert activation reduction mechanism that selectively decreases activated experts without retraining, and (2) a routing-aware token pruning method that evaluates visual token redundancy through routing-probability similarity and attention importance. 
Comprehensive experiments on InternVL3.5 and DeepSeek-VL2 demonstrate that FastMMoE achieves substantial FLOPs reduction while preserving nearly all original performance, consistently outperforming dense-model pruning baselines such as FastV and SparseVLM. 
Further analyses reveal that vision tokens exhibit higher magnitude stability across experts, explaining why expert activation reduction exerts limited influence on performance. 
Our findings highlight that current MoE-MLLMs often underutilize visual experts, suggesting new opportunities for adaptive expert routing and more efficient multimodal interaction in future research.
{
    \small
    \bibliographystyle{ieeenat_fullname}
    \bibliography{main}
}

\clearpage
\setcounter{page}{1}
\maketitlesupplementary

\appendix

% \section{Implementation Details and Hyperparameters Settings}
% \label{sec:imple_ap}

% For the experiments reported in Table~\ref{tab:internvl_main_table}, the hyperparameters of \textbf{FastMMoE} are set as follows:  

% \vspace{0.5em}
% \noindent\textbf{InternVL3.5 (MoE-based)}  
% \begin{itemize}
%     \item \textbf{75\% token retention}: $\alpha = 0.5$, $\gamma = 0.025$, $W=5$
%     \item \textbf{50\% token retention}: $\alpha = 0.6$, $\gamma = 0.05$, $W=5$
%     \item \textbf{25\% token retention}: $\alpha = 1.0$, $\gamma = 0.1$, $W=5$
%     \item \textbf{token pruning layer id}: 5,8,12
% \end{itemize}

% \noindent\textbf{DeepSeek-VL2 (MoE-based)}
% \begin{itemize}
%     \item \textbf{75\% token retention}: $\alpha = 0.3$, $\gamma = 0.05$, $W=3$
%     \item \textbf{50\% token retention}: $\alpha = 0.9$, $\gamma = 0.05$, $W=3$
%     \item \textbf{25\% token retention}: $\alpha = 0.7$, $\gamma = 0.2$, $W=3$
%     \item \textbf{token pruning layer id}: 2,5,8
% \end{itemize}

% These settings correspond to the best-performing configurations observed in the ablation studies (\S\ref{sec:ablation}).  

% All evaluations are conducted using \texttt{VLMEvalKit} on \textbf{NVIDIA A800 GPUs}, with the sampling temperature fixed at 0 and sampling disabled to ensure reproducible and stable results across all benchmarks.

\section{Implementation Details and Hyperparameters Settings}
\label{sec:imple_ap}

For the experiments reported in the main text, the hyperparameters of \textbf{FastMMoE} are set as follows:  

\vspace{0.5em}
\noindent\textbf{InternVL3.5 (MoE-based)}  
\begin{itemize}
    \item \textbf{75\% token retention}: $\alpha = 0.5$, $\gamma = 0.025$, $W=5$
    \item \textbf{50\% token retention}: $\alpha = 0.6$, $\gamma = 0.05$, $W=5$
    \item \textbf{25\% token retention}: $\alpha = 1.0$, $\gamma = 0.1$, $W=5$
    \item \textbf{token pruning layer id}: 5,8,12
\end{itemize}

\noindent\textbf{DeepSeek-VL2 (MoE-based)}
\begin{itemize}
    \item \textbf{75\% token retention}: $\alpha = 0.3$, $\gamma = 0.05$, $W=3$
    \item \textbf{50\% token retention}: $\alpha = 0.9$, $\gamma = 0.05$, $W=3$
    \item \textbf{25\% token retention}: $\alpha = 0.7$, $\gamma = 0.2$, $W=3$
    \item \textbf{token pruning layer id}: 2,5,8
\end{itemize}
\paragraph{Hyperparameter Tuning Strategy.}
These settings correspond to the best-performing configurations observed in the ablation studies. Across our detailed experiments, we observed a consistent pattern:
\begin{itemize}
    \item \textbf{High Retention (e.g., 75\%):} A lower $\alpha$ ($\le 0.5$) is generally used, as attention weights constitute the primary component of the redundancy score. The merge rate $\gamma$ typically adopts the theoretical upper bound (detailed in Appendix~\ref{sec:gamma_theory}). 
    \item \textbf{Lower Retention (e.g., 50\%, 25\%):} As the retention ratio decreases, appropriately increasing both $\alpha$ and $\gamma$ yields better performance. This validates our insight that routing similarity effectively measures token similarity, becoming increasingly critical for preserving information as pruning intensifies.
    \item \textbf{Model-Specific Behavior (DeepSeek-VL2):} Interestingly, for DeepSeek-VL2 at 50\% and 25\% retention ratios, attention weights still require a high degree of participation in the redundancy score to maintain optimal performance. We attribute this to two primary factors: (1) The presence of shared experts makes DeepSeek-VL2 structurally more similar to dense models, and attention-based visual token pruning has already been proven to be a highly effective approach for dense architectures. (2) The relatively small total number of experts in DeepSeek-VL2 (i.e., 72) may not provide sufficiently precise and effective visual token similarity information when relying heavily on routing distributions alone.
\end{itemize}
\paragraph{Baseline Implementation and Fair Comparison.}
To ensure a fair comparison of FLOPs overhead, we implemented all baseline methods (including FastV and SparseVLM) using a \textbf{multi-stage progressive pruning design}, strictly aligning the pruning layers across all methods. This multi-stage setup aligns with the core methodology and insights of PDrop~\citep{xing2024pyramiddrop}. FastMMoE consistently outperforms these strong multi-stage baselines.
\paragraph{Evaluation Setup.}
All evaluations are conducted using \texttt{VLMEvalKit} on \textbf{NVIDIA A800 GPUs}, with the sampling temperature fixed at 0 and sampling disabled to ensure reproducible and stable results across all benchmarks.

\section{Details of Test Results}
\label{sec:test_results}

To supplement the quantitative results summarized in the main paper, 
this section presents the \emph{complete benchmark results and per-configuration analyses} 
for both \textbf{DeepSeek-VL2} and \textbf{InternVL3.5}.  
These tables are provided for reproducibility and to facilitate more fine-grained comparison.

\subsection{Comprehensive results on DeepSeek-VL2.}
\cref{tab:deepseek_main_results} reports the full benchmark performance 
of all pruning and acceleration methods on the \textbf{MoE-based DeepSeek-VL2}. 
It extends \cref{tab:deepseek_vl2_performance} in the main text by including results 
across six multimodal benchmarks (MMMU, SQA$^\text{I}$, MMBench, OCRBench, HallusionBench, AI2D).  
The table also differentiates between \textbf{FastMMoE$^*$} 
(i.e., token pruning only without reducing activated experts) 
and the complete \textbf{FastMMoE} method.  

From the full results we observe that, on DeepSeek-VL2, the performance gap among the three pruning methods 
(\textbf{FastMMoE}, \textbf{FastV}, and \textbf{SparseVLM}) is smaller than that observed on InternVL3.5.  
A plausible explanation is that DeepSeek-VL2’s architecture includes \textbf{two shared experts} 
that are always activated for every token.  
This design is structurally closer to a dense Transformer, meaning that methods originally developed for dense models (e.g., FastV, SparseVLM) retain relatively high performance.  
Consequently, while \textbf{FastMMoE} still achieves the best balance between accuracy and FLOPs reduction, 
the relative advantage over the dense-model-oriented baselines is not as pronounced as in the InternVL3.5 experiments.

\subsection{Full ablation of expert activation reduction on InternVL3.5.}
\cref{tab:internvl_reduce_act_all} provides the \emph{complete results} 
for the expert activation reduction strategy on \textbf{InternVL3.5-30B-A3B}, 
complementing the activation reduction heatmaps shown in the main text (Section~\ref{sec:ablation}).  
Each row in \cref{tab:internvl_reduce_act_all} corresponds to a specific $(K_v, l_v)$ configuration, 
where $K_v$ is the number of experts activated per vision token and $l_v$ is the starting layer index 
from which activation reduction is applied.  
The \textbf{Avg.$\uparrow$} column reports the average score across all six benchmarks, 
corresponding to the same metric definition used in the main paper.  
This table enables precise reproduction of the InternVL3.5 activation-reduction experiments, 
and it verifies that the general trends described in the main text hold consistently 
across all benchmarks.

\subsection{Full ablation of expert activation reduction on DeepSeek-VL2.}
The final long table in this section lists the complete activation reduction ablation results 
for \textbf{DeepSeek-VL2}, following the same structure and column definitions as 
\cref{tab:deepseek_reduce_act_all}.  
This facilitates direct cross-model comparison, allowing readers to observe whether 
the activation-reduction trends evident in InternVL3.5 experiments also manifest 
in DeepSeek-VL2.

All results here are obtained under the same evaluation protocol using \texttt{VLMEvalKit}, 
ensuring complete consistency with the benchmarks and metrics used in the main paper.

\subsection{Reduce Share Experts for DeepSeek-VL2}

DeepSeek-VL2 adopts a MoE architecture that contains two \emph{shared experts} which are
always activated for every token, in addition to the routed experts.  
These shared experts are designed to handle general-purpose features across modalities,
e.g., global semantic composition and common reasoning patterns, which complement the 
specialized experts selected via routing.
Given their architectural role, we hypothesized that reducing the number of shared experts 
during vision-token activation reduction could negatively impact overall performance, 
especially on tasks requiring fine-grained multimodal reasoning.

\cref{tab:reduce_share_experts} verifies this hypothesis:  
when halving the number of routed experts for vision tokens ($K_v$ from 6 to 3, $l_v=15$) but \textbf{keeping} 
both shared experts active (\texttt{No} in column ``Reduce ShareExperts''), average performance across 
benchmarks remains essentially unchanged compared to the full-expert baseline (72.83 vs.\ 72.77).
However, if we also reduce the number of \emph{shared experts} by half (\texttt{Yes} in ``Reduce ShareExperts''),
OCRBench accuracy drops sharply from 81.10 to 70.80 --- a degradation of more than 10 points --- while the overall
average drops to 70.95.  
This large decline indicates that shared experts carry crucial modality-agnostic knowledge
that is especially important for OCR and similar text-rich visual tasks.
Given the critical nature of such tasks in real-world multimodal applications,  
we adopt the conservative strategy of \textbf{never reducing shared experts} in our activation-reduction pipeline
for DeepSeek-VL2.

% ------------- Table -------------
\begin{table*}[!t]
\centering
\caption{\textbf{Effect of reducing shared experts in DeepSeek-VL2 during expert activation reduction.}
``Reduce ShareExperts''=Yes indicates halving the shared-expert count alongside routed experts;
``No'' keeps all shared experts active.  
Reducing shared experts causes a severe drop in OCRBench performance and a notable decrease
in the overall average score, highlighting their importance for modality-general knowledge retention.}
\label{tab:reduce_share_experts}
\resizebox{\textwidth}{!}{%
\begin{tabular}{@{}c|c|c|cccccc|c@{}}
\toprule
\multicolumn{1}{l|}{\textbf{Reduce ShareExperts}} &
  \multicolumn{1}{l|}{\textbf{$K_v$}} &
  \multicolumn{1}{l|}{\textbf{$l_v$}} &
  \textbf{MMMU} &
  \textbf{SQA$^\text{I}$} &
  \textbf{MMBench} &
  \textbf{OCRBench} &
  \textbf{HallusionBench} &
  \textbf{AI2D} &
  \textbf{Avg.$\uparrow$} \\ \midrule
No  & 6 & 29 & 51.89 & 96.88 & 83.25 & \textbf{81.40} & 40.83 & 82.38 & 72.77 \\
No  & 3 & 15 & \textbf{52.11} & \textbf{96.88} & 83.16 & \textbf{81.10} & \textbf{41.44} & \textbf{82.29} & \textbf{72.83} \\
Yes & 3 & 15 & 51.89 & 96.58 & 82.90 & \underline{70.80} & 41.22 & 82.32 & 70.95 \\ \bottomrule
\end{tabular}%
}
\end{table*}

\subsection{Full ablation of $\alpha$ and $\gamma$ configurations}

In the main text, \cref{tab:alpha_ablation} and \cref{tab:merge_ratio_ablation} 
summarize the key results for the routing-similarity weighting parameter $\alpha$ 
and merge rate $\gamma$ under several vision-token retention ratios on InternVL3.5.  
To provide a complete view of the parameter search space, 
Appendix~\ref{sec:test_results} includes two extended tables 
(\cref{tab:internvl_pruning_ablation} for InternVL3.5 and 
\cref{tab:deepseek_pruning_ablation} for DeepSeek-VL2) that enumerate the full 
$(\alpha,\gamma)$ combinations tested and their corresponding scores on all six benchmarks.

These extended tables complement the condensed results in the main text by showing:
\begin{itemize}
    \item The full performance landscape across a wide range of $\alpha$ and $\gamma$ values.
    \item How the optimal $\alpha$ and $\gamma$ vary with pruning intensity (different retention ratios).
    \item That, consistent with the conclusions in the main paper, moderate $\alpha$ values and merge rates close to the theoretical upper bound 
          provide the best trade-off between accuracy and compression.
\end{itemize}

The DeepSeek-VL2 results further confirm the general trends, while showing smaller performance fluctuations across $(\alpha,\gamma)$—consistent with its more stable architecture that includes two permanently activated shared experts.

\begin{table*}[!t]
\centering
\caption{\textbf{Peformance of all methods across different benchmarks for DeepSeek-VL2.} FastMMoE$^*$ denotes that we only apply token pruning method without reducing activated experts. The best results and second best results are indicated by \textbf{boldface} and {\ul underline}, respectively. FastMMoE$^\dagger$ denotes that we apply token pruning with reducing activated experts.}
\label{tab:deepseek_main_results}
\resizebox{\textwidth}{!}{%
\begin{tabular}{@{}lccccccccc@{}}
\toprule
\multicolumn{1}{l|}{\textbf{Method}} &
  \textbf{MMMU} &
  \textbf{SQA$^\text{I}$} &
  \textbf{MMBench} &
  \textbf{OCRBench} &
  \textbf{HallusionBench} &
  \textbf{AI2D} &
  \textbf{Avg.$\uparrow$} &
  \multicolumn{1}{c|}{\textbf{Drop.$\downarrow$}} &
  \textbf{Saving FLOPs. (\%)} \\ \midrule
\multicolumn{1}{l|}{\textbf{Baseline}} &
  \textbf{51.33} &
  \textbf{96.88} &
  \textbf{83.25} &
  \textbf{81.40} &
  \textbf{40.83} &
  \multicolumn{1}{c|}{\textbf{82.38}} &
  \multicolumn{1}{c|}{\textbf{72.68}} &
  \multicolumn{1}{c|}{\textbf{0}} &
  \textbf{0} \\ \midrule
\multicolumn{10}{c}{\cellcolor[HTML]{EFEFEF}\textit{\textbf{Retain 75\% vision tokens after pruning}}} \\
\multicolumn{1}{l|}{FastV} &
  50.67 &
  96.93 &
  82.56 &
  {\ul 80.30} &
  {\ul 40.66} &
  \multicolumn{1}{c|}{{\ul 81.99}} &
  \multicolumn{1}{c|}{{\ul 72.18}} &
  \multicolumn{1}{c|}{0.50} &
   \\ \cmidrule(r){1-9}
\multicolumn{1}{l|}{SparseVLM} &
  51.11 &
  96.53 &
  \textbf{82.65} &
  78.00 &
  40.45 &
  \multicolumn{1}{c|}{81.57} &
  \multicolumn{1}{c|}{71.72} &
  \multicolumn{1}{c|}{0.96} &
   \\ \cmidrule(r){1-9}
\multicolumn{1}{l|}{FastMMoE$^*$} &
  \textbf{51.56} &
  \textbf{96.98} &
  {\ul 83.25} &
  \textbf{80.80} &
  40.30 &
  \multicolumn{1}{c|}{\textbf{82.06}} &
  \multicolumn{1}{c|}{\textbf{72.49}} &
  \multicolumn{1}{c|}{\textbf{0.19}} &
  \multirow{-3}{*}{21.09} \\ \midrule
\multicolumn{1}{l|}{\begin{tabular}[c]{@{}l@{}}FastMMoE$^\dagger$\\ ($l_v=10, K_v=2$)\end{tabular}} &
  {\ul 51.44} &
  96.63 &
  82.82 &
  78.90 &
  \textbf{40.68} &
  \multicolumn{1}{c|}{81.54} &
  \multicolumn{1}{c|}{72.00} &
  \multicolumn{1}{c|}{0.68} &
  39.15 \\ \midrule
\multicolumn{10}{c}{\cellcolor[HTML]{EFEFEF}\textit{\textbf{Retain 50\% vision tokens after pruning}}} \\
\multicolumn{1}{l|}{FastV} &
  51.00 &
  \textbf{96.48} &
  {\ul 82.90} &
  {\ul 79.20} &
  \textbf{40.71} &
  \multicolumn{1}{c|}{{\ul 81.19}} &
  \multicolumn{1}{c|}{{\ul 71.91}} &
  \multicolumn{1}{c|}{{\ul 0.77}} &
   \\ \cmidrule(r){1-9}
\multicolumn{1}{l|}{SparseVLM} &
  49.33 &
  95.88 &
  82.47 &
  72.80 &
  40.83 &
  \multicolumn{1}{c|}{80.51} &
  \multicolumn{1}{c|}{70.31} &
  \multicolumn{1}{c|}{2.37} &
   \\ \cmidrule(r){1-9}
\multicolumn{1}{l|}{FastMMoE$^*$} &
  {\ul 51.11} &
  {\ul 96.33} &
  \textbf{82.99} &
  \textbf{79.30} &
  {\ul 40.55} &
  \multicolumn{1}{c|}{\textbf{81.35}} &
  \multicolumn{1}{c|}{\textbf{71.94}} &
  \multicolumn{1}{c|}{\textbf{0.74}} &
  \multirow{-3}{*}{43.07} \\ \midrule
\multicolumn{1}{l|}{\begin{tabular}[c]{@{}l@{}}FastMMoE$^\dagger$\\ ($l_v=2, K_v=2$)\end{tabular}} &
  \textbf{51.56} &
  95.69 &
  82.65 &
  75.40 &
  37.75 &
  \multicolumn{1}{c|}{80.38} &
  \multicolumn{1}{c|}{70.57} &
  \multicolumn{1}{c|}{2.11} &
  61.76 \\ \midrule
\multicolumn{10}{c}{\cellcolor[HTML]{EFEFEF}\textit{\textbf{Retain 25\% vision tokens after pruning}}} \\
\multicolumn{1}{l|}{FastV} &
  \textbf{50.56} &
  \textbf{95.44} &
  \textbf{82.13} &
  {\ul 74.30} &
  {\ul 38.90} &
  \multicolumn{1}{c|}{78.82} &
  \multicolumn{1}{c|}{{\ul 70.02}} &
  \multicolumn{1}{c|}{{\ul 2.65}} &
   \\ \cmidrule(r){1-9}
\multicolumn{1}{l|}{SparseVLM} &
  48.11 &
  94.74 &
  81.44 &
  59.00 &
  37.33 &
  \multicolumn{1}{c|}{{\ul 79.02}} &
  \multicolumn{1}{c|}{66.61} &
  \multicolumn{1}{c|}{6.07} &
   \\ \cmidrule(r){1-9}
\multicolumn{1}{l|}{FastMMoE$^*$} &
  \textbf{50.56} &
  {\ul 95.34} &
  {\ul 82.04} &
  \textbf{74.70} &
  \textbf{40.11} &
  \multicolumn{1}{c|}{\textbf{79.15}} &
  \multicolumn{1}{c|}{\textbf{70.32}} &
  \multicolumn{1}{c|}{\textbf{2.36}} &
  \multirow{-3}{*}{63.66} \\ \bottomrule
\end{tabular}%
}
\end{table*}

\section{Theoretical Analysis on Expert Activation Reduction}
\label{sec:magnitude_concentration}

We provide the detailed mathematical derivation to explain why reducing the number of activated experts for vision tokens causes only minor performance degradation.

\subsection{Details of Vision Tokens Exhibit Magnitude Concentration}

\begin{figure}[!h]
\centering
  \includegraphics[width=\columnwidth,keepaspectratio]{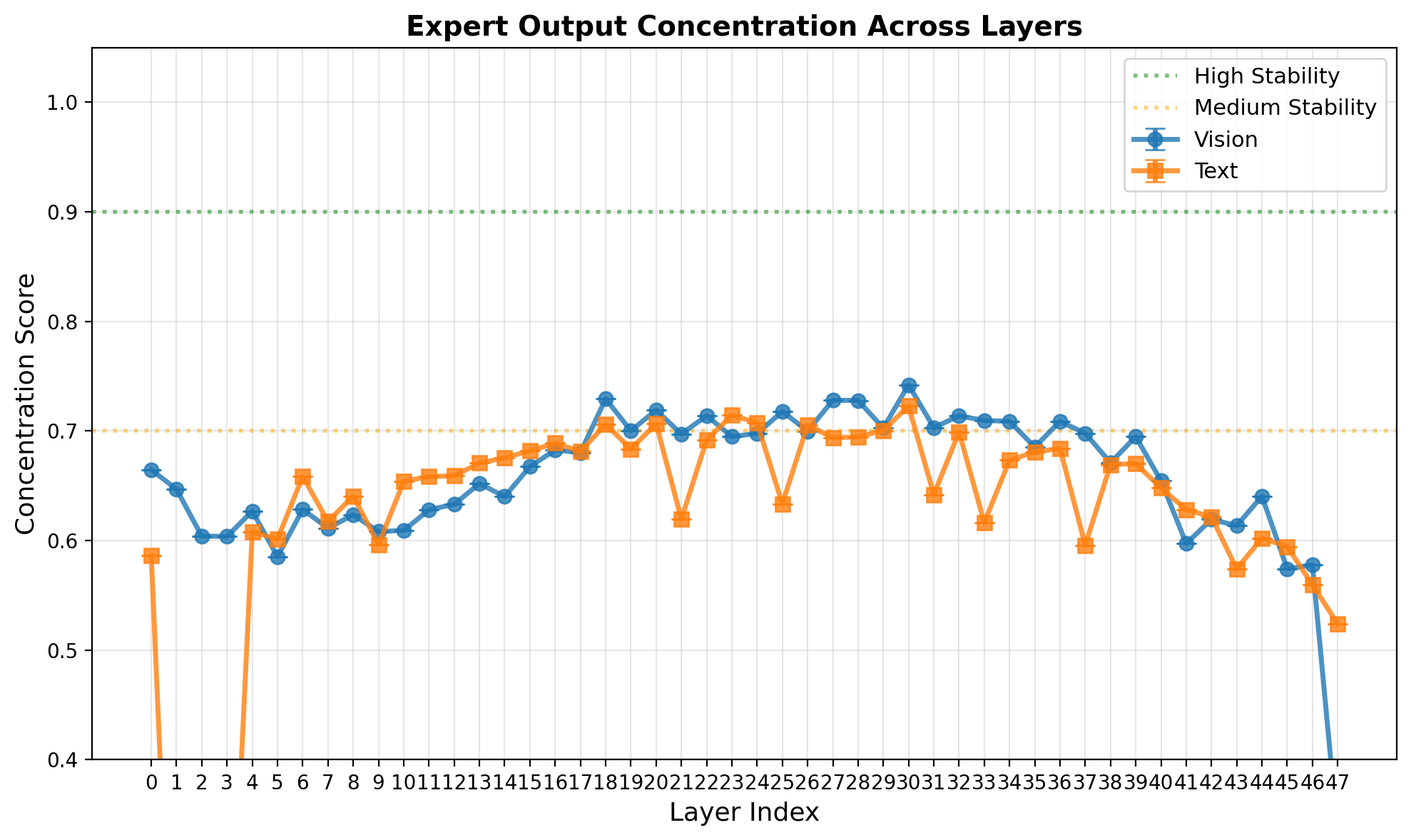}
  \caption{\textbf{Magnitude stability score $V_m$ for InternVL3.5 across layers.} Higher $V_m$ means tighter magnitude concentration among expert outputs.}
  \label{fig:stab_main_internvl}
\end{figure}

\begin{figure}[!h]
\centering
  \includegraphics[width=\columnwidth,keepaspectratio]{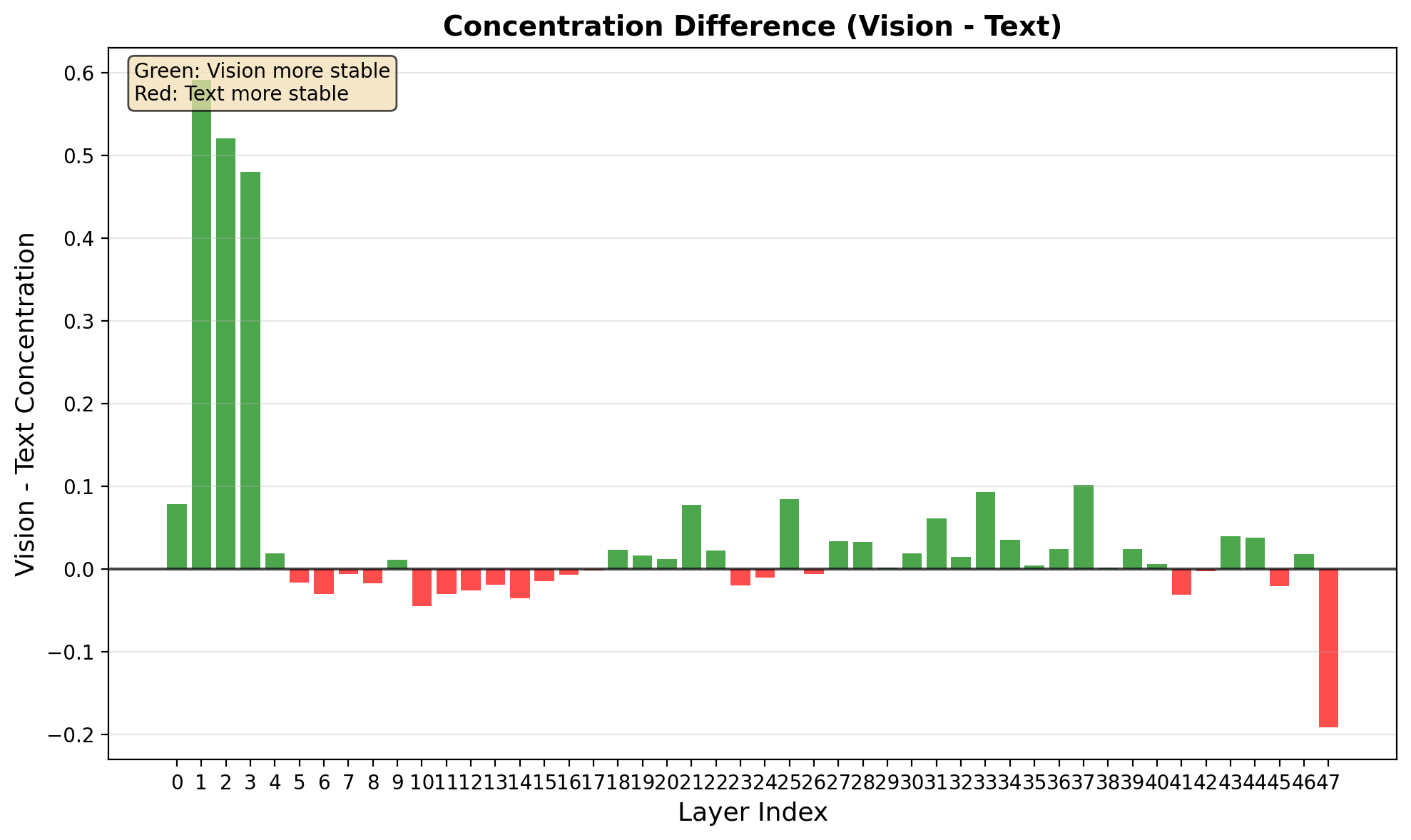}
  \caption{\textbf{Modal stability difference $V_{\text{vision}}-V_{\text{text}}$ in InternVL3.5.} Green bars: vision more stable; red bars: text more stable.}
  \label{fig:stab_diff_internvl}
\end{figure}

\begin{figure}[!h]
\centering
  \includegraphics[width=\columnwidth,keepaspectratio]{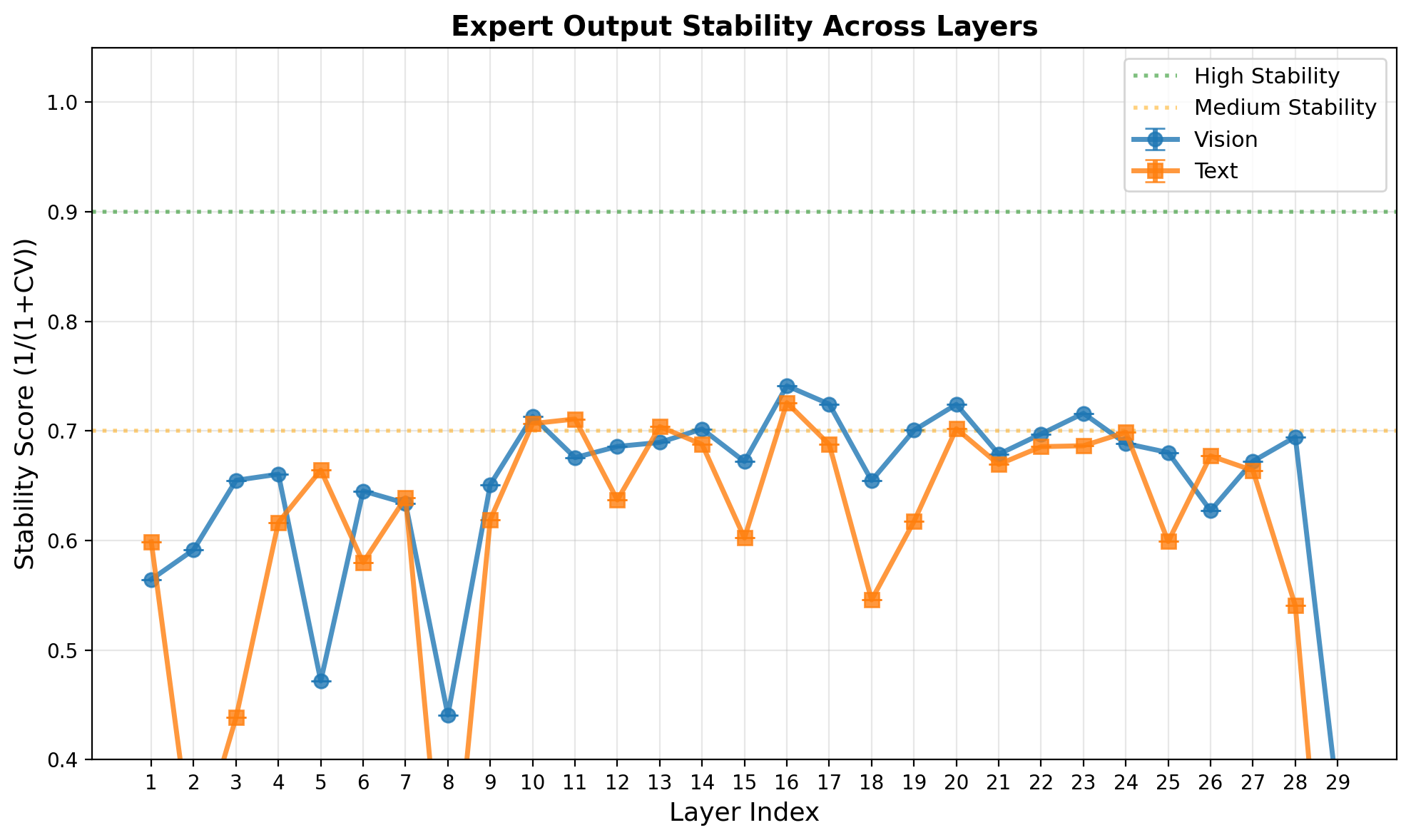}
  \caption{\textbf{Magnitude stability score $V_m$ for DeepSeek-VL2 across layers.}}
  \label{fig:stab_main_deepseek}
\end{figure}

\begin{figure}[!h]
\centering
  \includegraphics[width=\columnwidth,keepaspectratio]{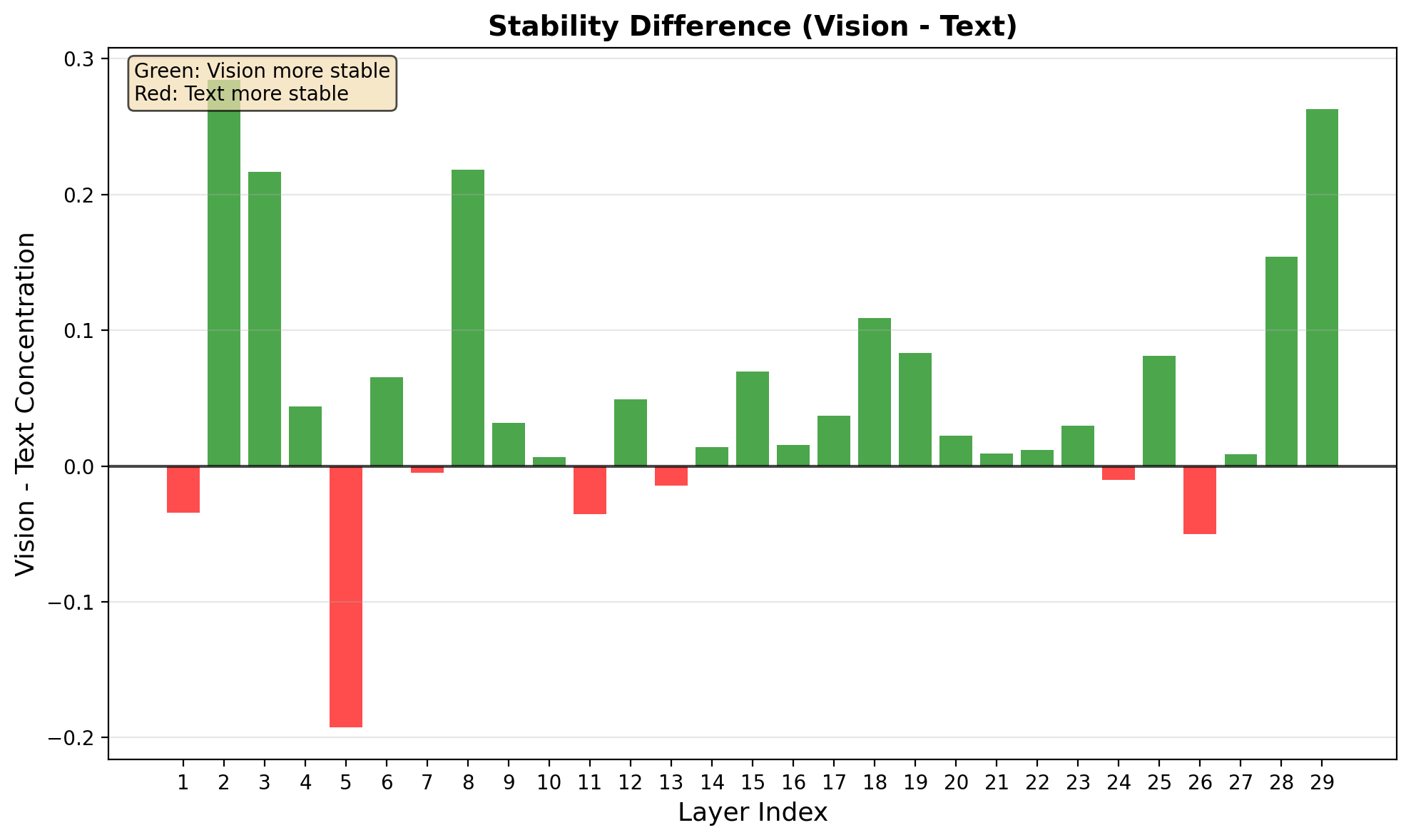}
  \caption{\textbf{Modal stability difference $V_{\text{vision}}-V_{\text{text}}$ in DeepSeek-VL2.}}
  \label{fig:stab_diff_deepseek}
\end{figure}

One important empirical observation motivating our method is that the output vector magnitudes of \emph{vision} tokens tend to be more concentrated across experts compared to \emph{text} tokens. This magnitude concentration means that, after activation reduction, the fused output norm changes very little; any difference between the reduced and original outputs is mainly in vector direction rather than magnitude, thereby minimizing semantic distortion while lowering computational cost.

To validate this hypothesis, we evaluate the magnitude statistics of expert outputs layer-by-layer for two representative MoE-based MLLMs: \textbf{InternVL3.5} and \textbf{DeepSeek-VL2}. For a multimodal input sequence, and for each decoder layer, we collect the intermediate outputs from all selected experts \emph{before} the weighted fusion operation in Eq.~\ref{con:moe_cal}. Formally, for token $i$ of modality $m \in \{\mathrm{Vision},\mathrm{Text}\}$ processed by expert $e$, we denote its raw output vector by
\[
\mathbf{f}_{m,i}^{(e)} \in \mathbb{R}^d,
\]
and compute its Euclidean norm
\[
\ell_{m,i}^{(e)} = \|\mathbf{f}_{m,i}^{(e)}\|_2.
\]

For each modality $m$ in a given layer, we aggregate all such norms and compute the \emph{coefficient of variation} (CV):
\[
\mathrm{CV}_m = \frac{\sigma_m}{\mu_m},
\]
where $\mu_m$ and $\sigma_m$ are the mean and standard deviation of the expert output norms for modality $m$. A smaller $\mathrm{CV}_m$ indicates that magnitudes are more tightly clustered around the mean, and hence the fused representation is more stable to changes in expert selection.

To present the results in a normalized and more interpretable way, we define the \textbf{Magnitude Stability Score}:
\[
V_m = \frac{1}{1 + \mathrm{CV}_m},
\]
which maps CV values into $(0,1]$, with higher $V_m$ implying greater magnitude concentration. This metric is computed separately for vision and text tokens across all decoder layers.

\cref{fig:stab_main_internvl,fig:stab_diff_internvl} present the magnitude stability score $V_m$ for InternVL3.5, while \cref{fig:stab_main_deepseek,fig:stab_diff_deepseek} give the same analysis for DeepSeek-VL2. 
In the first figure of each pair, $V_m$ is plotted per decoder layer for both modalities, together with horizontal reference lines marking ``High Stability'' ($V_m \approx 0.9$) and ``Medium Stability'' ($V_m \approx 0.7$). 
In the second figure, the per-layer difference $\Delta V = V_{\text{vision}} - V_{\text{text}}$ is shown, with green bars where vision tokens are more stable and red bars where text tokens are more stable. 

Across most layers in both models, vision tokens exhibit higher stability scores than text tokens, indicating more concentrated expert-output magnitudes. 
For InternVL3.5, large positive gaps (up to $+0.5$) appear in the early layers (Layer~2–3), while medium positive differences are common in later middle layers; a few layers towards the end show negative gaps, with text stability slightly higher. 
For DeepSeek-VL2, positive differences are distributed more evenly across layers, occasionally exceeding $+0.20$ in both early and late stages, but some mid-layer and output-stage layers display small negative differences ($\approx -0.1$ to $-0.2$). 
In both models, many layers maintain $V_m$ values around the Medium Stability regime, supporting the observation that magnitude concentration is generally high—especially for vision tokens. 

This empirical evidence verifies that vision-token expert outputs tend to have lower variance in magnitude.  
As a result, reweighting after pruning lower-weight experts alters the fused vector norm only marginally; the main effect of activation reduction is a slight change in output vector \emph{direction} relative to the original, a phenomenon quantified in the subsequent theoretical analysis subsection.

\begin{figure*}[!t]
  \centering % 整体居中figure*的内容
  \begin{subfigure}{\textwidth} % 子图宽度设置为文本宽度
    \centering % 子图内部内容居中
    \includegraphics[width=0.7\textwidth]{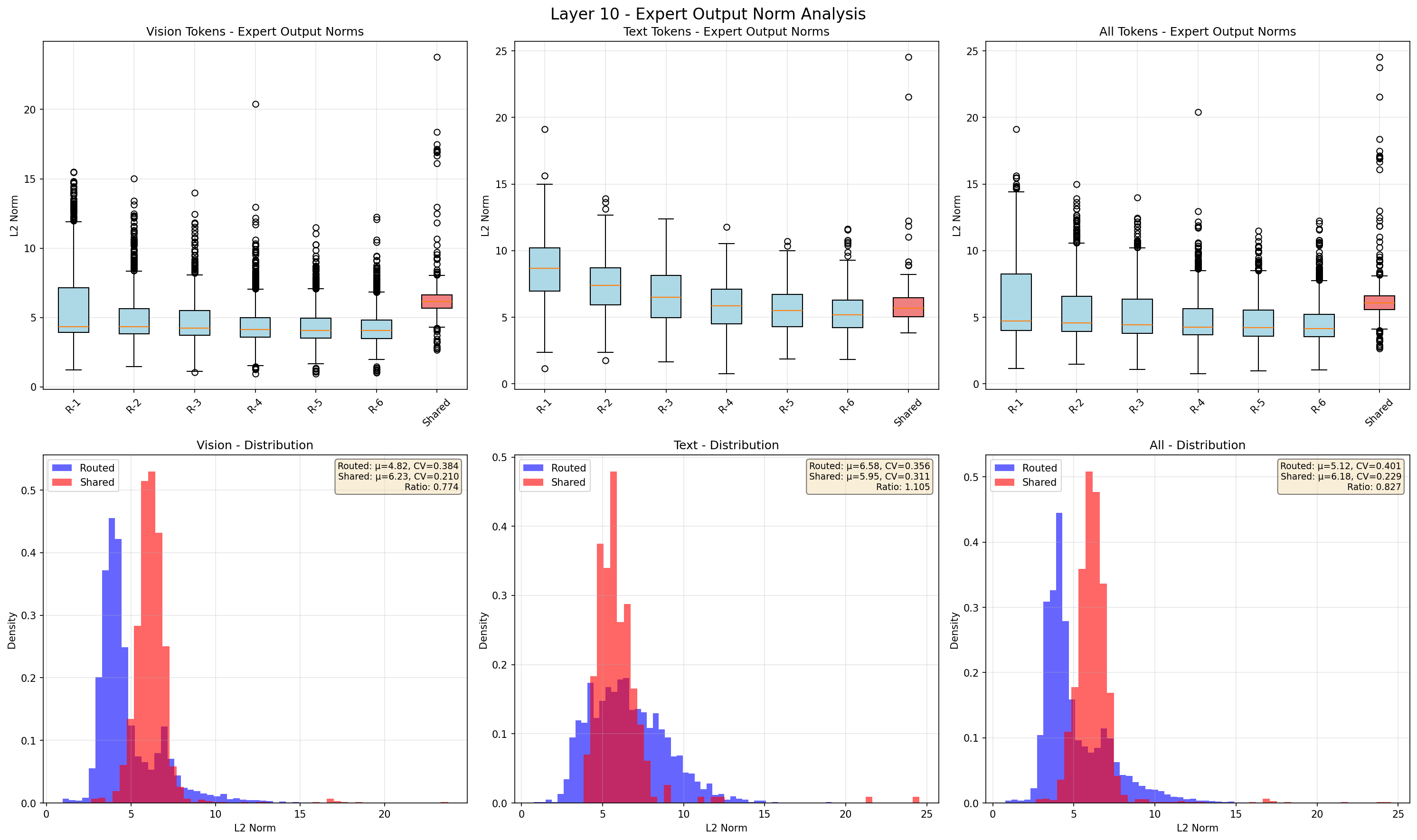} % 调整子图内图片的宽度
  \end{subfigure}

\begin{subfigure}{\textwidth} % 子图宽度设置为文本宽度
    \centering % 子图内部内容居中
    \includegraphics[width=0.7\textwidth]{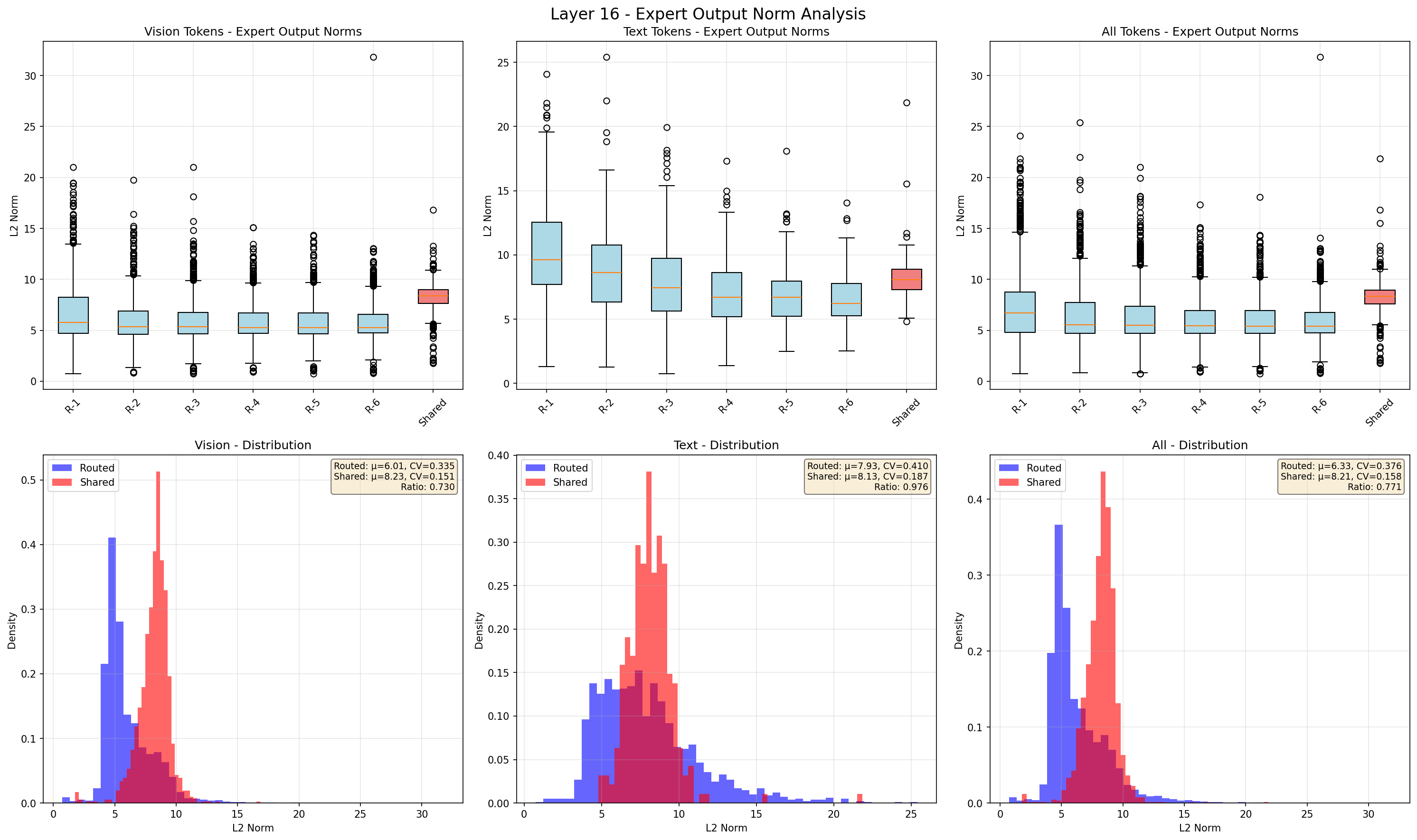} % 调整子图内图片的宽度
\end{subfigure}

\begin{subfigure}{\textwidth} % 子图宽度设置为文本宽度
    \centering % 子图内部内容居中
    \includegraphics[width=0.7\textwidth]{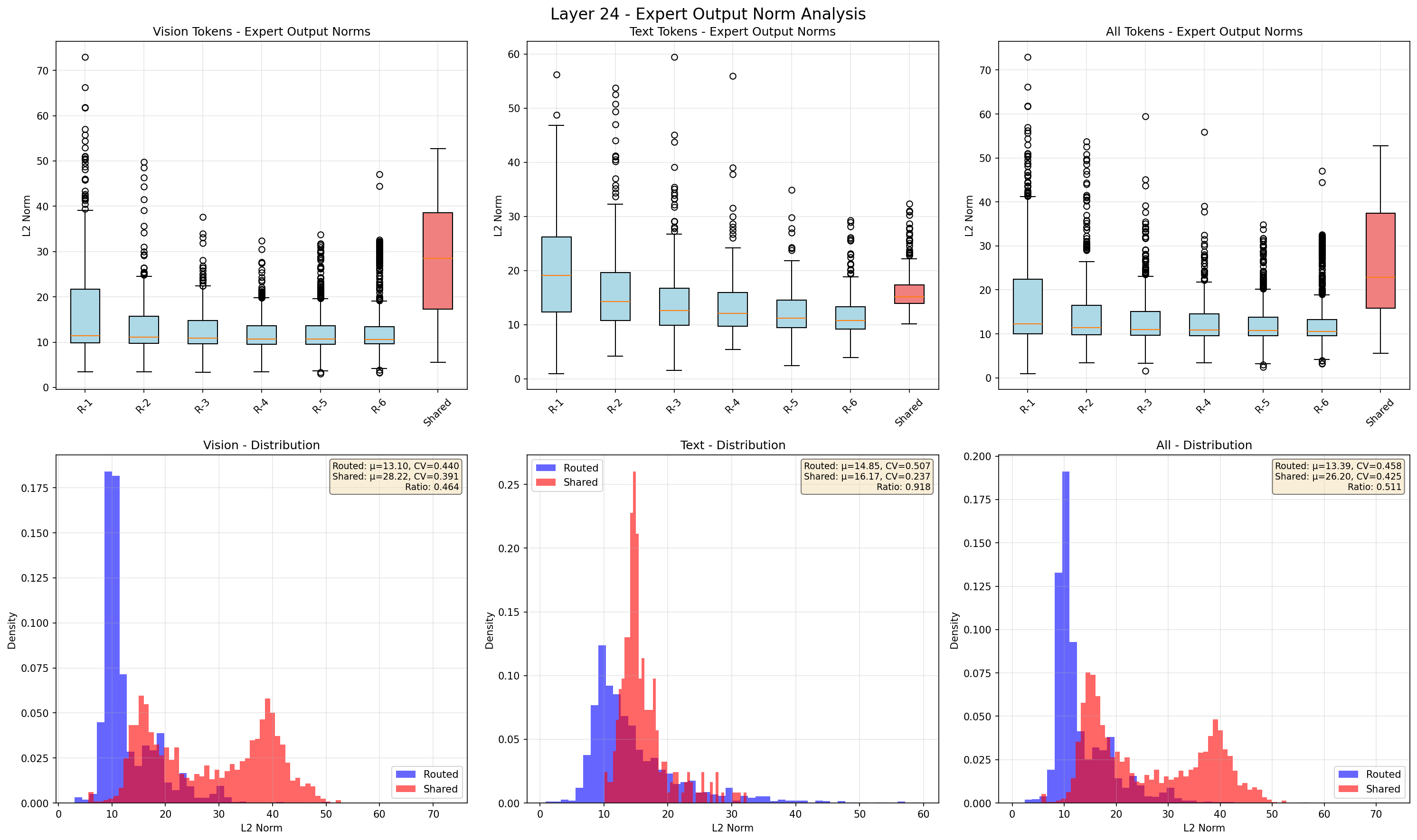} % 调整子图内图片的宽度
\end{subfigure}

  \caption{The experts outputs norm distribution of different modalities in DeepSeek-VL2.} % 整个figure的总标题
  \label{fig:ds_experts_outputs_norm} % 整个figure的标签 (建议修改以区别于之前的标签)
\end{figure*}

\begin{figure*}[!t]
  \centering % 整体居中figure*的内容
  \begin{subfigure}{\textwidth} % 子图宽度设置为文本宽度
    \centering % 子图内部内容居中
    \includegraphics[width=0.7\textwidth]{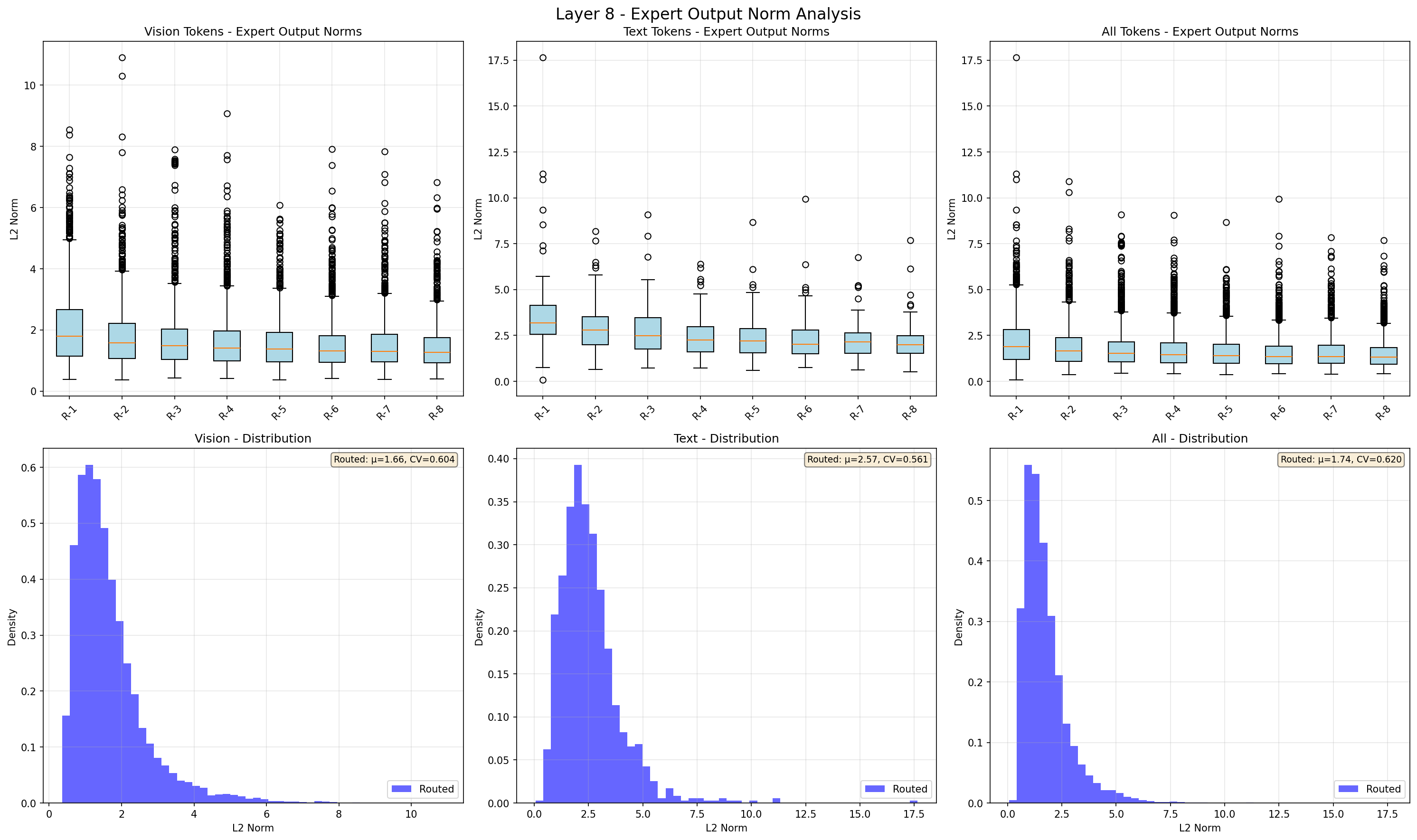} % 调整子图内图片的宽度
  \end{subfigure}

  \begin{subfigure}{\textwidth} % 子图宽度设置为文本宽度
    \centering % 子图内部内容居中
    \includegraphics[width=0.7\textwidth]{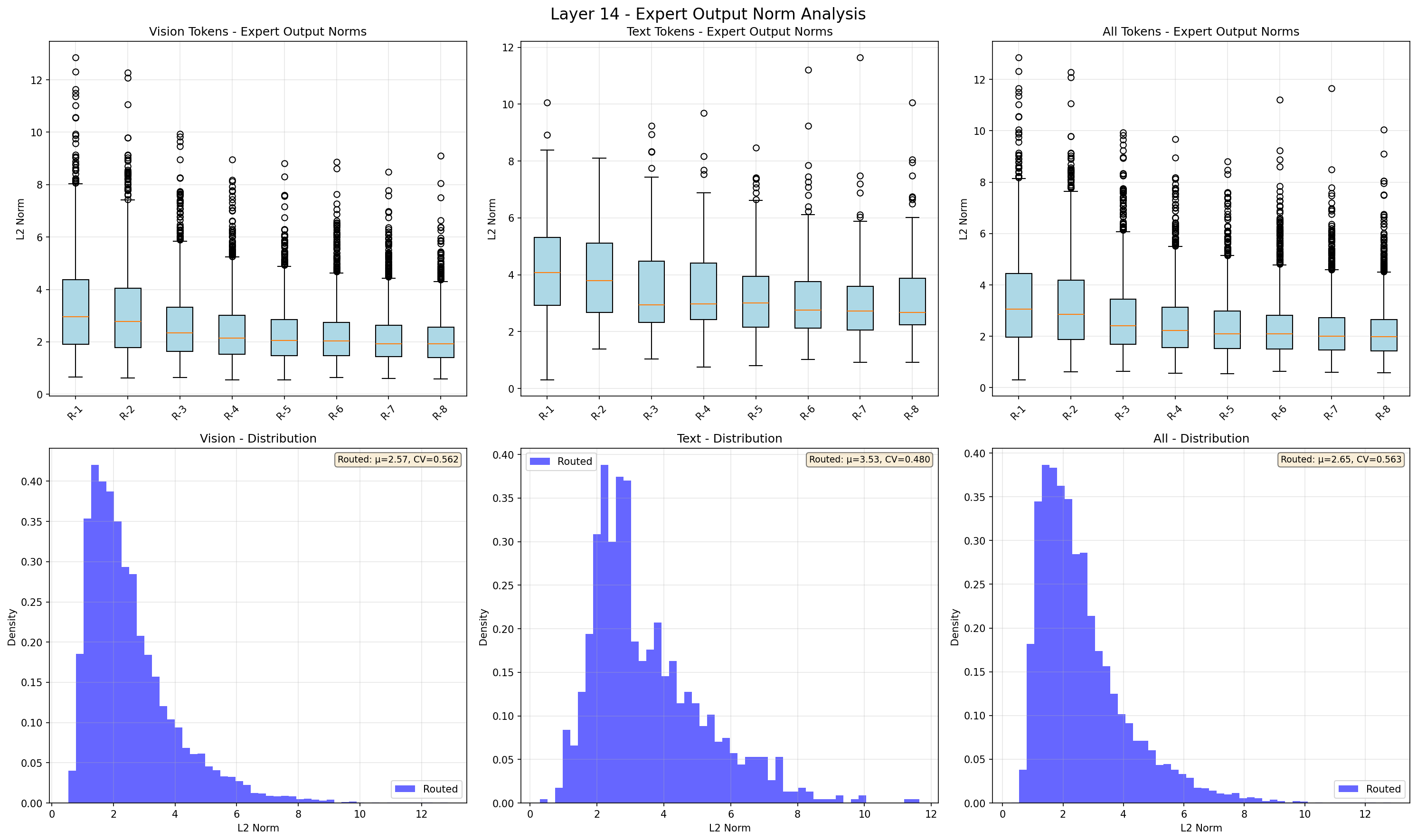} % 调整子图内图片的宽度
  \end{subfigure}

  \begin{subfigure}{\textwidth} % 子图宽度设置为文本宽度
    \centering % 子图内部内容居中
    \includegraphics[width=0.7\textwidth]{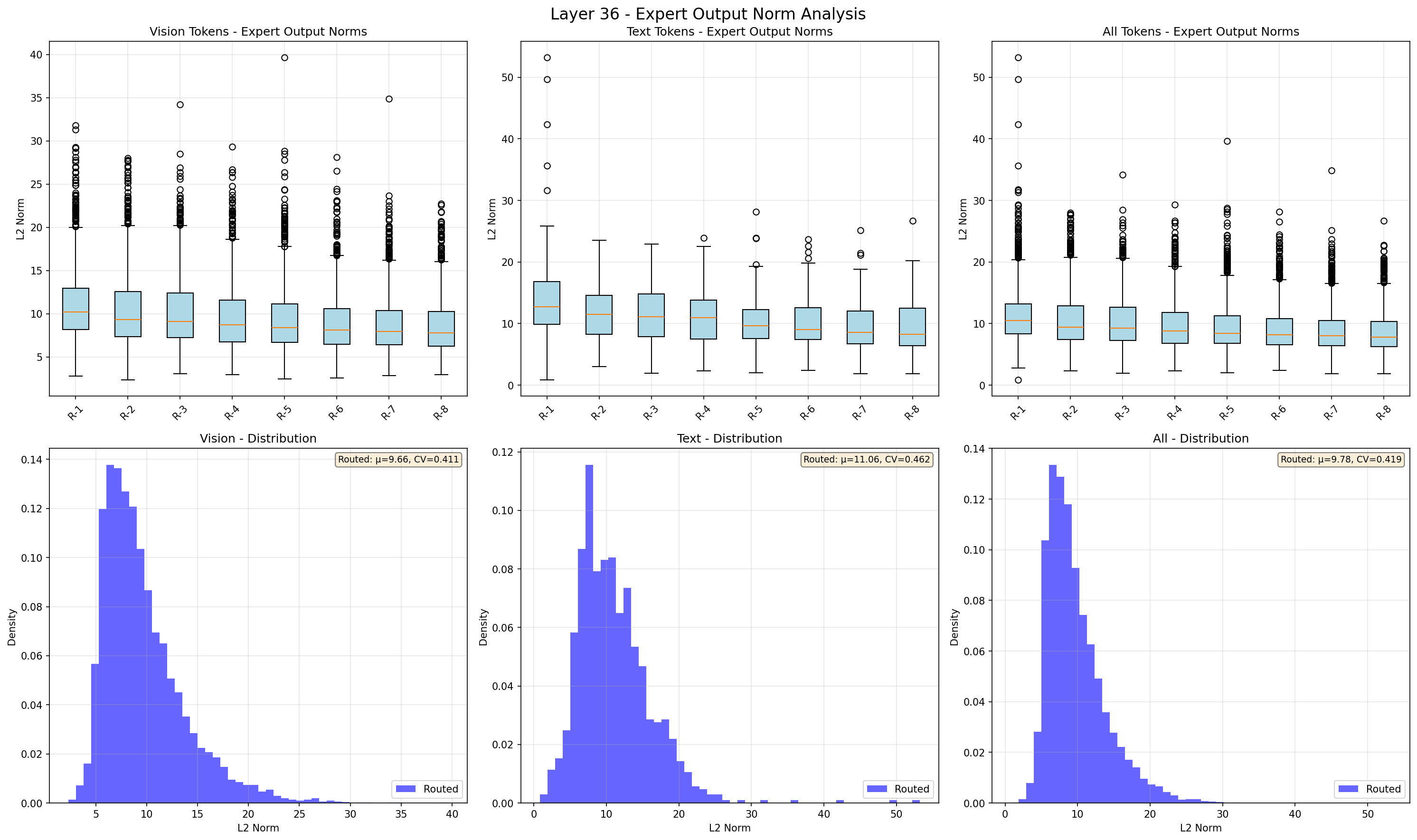} % 调整子图内图片的宽度
  \end{subfigure}

  \caption{The experts outputs norm distribution of different modalities in InternVL3.5.} % 整个figure的总标题
  \label{fig:internvl_experts_outputs_norm}
\end{figure*}

\cref{fig:ds_experts_outputs_norm,fig:internvl_experts_outputs_norm} illustrate the distribution of expert output L2 norms across different layers for DeepSeek-VL2 and InternVL3.5, respectively. R-1 denotes the expert with the highest routing weight in the current layer, with subsequent ranks indicating decreasing weights. We observe a clear positive correlation between routing weight rank and output norm magnitude: higher-weighted experts tend to produce larger and more stable (lower CV) outputs. 

Under our expert activation reduction strategy, retaining experts with the highest routing weights minimizes the loss in output norm concentration. Since these top-ranked experts dominate both the magnitude and stability of the fused representation, pruning lower-weight experts leads to only minor changes in norm distribution and, consequently, minimal performance degradation.

\subsection{Setup}

Let \( e_i(\cdot) \) denote the transformation function of the \(i\)-th expert, and let 
\( x \in \mathbb{R}^d \) be the hidden input vector.
The routing network outputs the routing weight vector
\( \hat{\mathbf{a}} = G(x) \in \mathbb{R}^E \),
where \(E\) is the number of non-shared experts.

After Top-\(K\) routing selection, we choose the set \(\mathcal{R}=\{r_1,r_2,\ldots,r_K\}\) of experts with the largest routing weights:
\[
\hat{a}_{r_1} \ge \hat{a}_{r_2} \ge \dots \ge \hat{a}_{r_K} > 0.
\]
The weights are normalized as:
\[
a_i = \frac{\hat{a}_{r_i}}{\sum_{j=1}^K \hat{a}_{r_j}}, \quad i=1,\dots,K,
\]
so that \(a_i > 0\) and \(\sum_{i=1}^K a_i = 1\).

Let \(p_i = e_{r_i}(x) \in \mathbb{R}^d\) denote the expert output vectors before fusion.  
The original MoE output is:
\begin{equation}
\label{eq:y_original}
y = \sum_{i=1}^K a_i\, p_i.
\end{equation}

\subsection{Reduction of Activated Experts}

We reduce the number of activated experts for a given token from \(K\) to \(m\) (\(m < K\)), keeping only the top-\(m\) weights and setting others to zero:
\[
a_i' =
\begin{cases}
\frac{a_i}{\sum_{j=1}^m a_j}, & i \le m, \\[4pt]
0, & i > m.
\end{cases}
\]
The reduced output is:
\begin{equation}
\label{eq:y_reduced}
y' = \sum_{i=1}^m a_i'\, p_i
    = \sum_{i=1}^m \frac{a_i}{\sum_{j=1}^m a_j} \, p_i.
\end{equation}

\subsection{Angle Between Original and Reduced Outputs}

Under the strong assumption that expert outputs are mutually orthogonal:
\[
p_i^\top p_j = 0, \quad \forall i \neq j,
\]
and have equal norms \(\|p_i\| = 1\), the cosine of the angle \(\theta\) between \(y\) and \(y'\) simplifies to:
\begin{equation}
\label{eq:cos_frac}
\cos\theta = \frac{\sqrt{\sum_{i=1}^m a_i^2}}{\sqrt{\sum_{i=1}^K a_i^2}},
\end{equation}
where \(a_1 \ge a_2 \ge \dots \ge a_K\).

\subsection{Lower Bound Proof}

Because \(a_i^2\) is non-negative and sorted in non-increasing order, the average of the first \(m\) terms is at least the average of all \(K\) terms:
\[
\frac{1}{m} \sum_{i=1}^m a_i^2 \;\ge\; \frac{1}{K} \sum_{i=1}^K a_i^2.
\]
Multiplying through by \(m\) gives:
\[
\frac{\sum_{i=1}^m a_i^2}{\sum_{i=1}^K a_i^2} \;\ge\; \frac{m}{K}.
\]
Taking square roots in \eqref{eq:cos_frac} yields the tight lower bound:
\begin{equation}
\label{eq:cos_lower_bound}
\cos\theta \ge \sqrt{\frac{m}{K}}.
\end{equation}
Equality occurs when all \(a_i\) are equal (\(a_i = 1/K\)), i.e., in the equal-weight case.

This bound is valid under the orthogonality and equal-norm assumption, and the equal-weight case corresponds to the worst possible \(\cos\theta\) (largest angle). For any non-uniform distribution of \(a_i\), the numerator \(\sum_{i=1}^m a_i^2\) increases relative to the denominator \(\sum_{i=1}^K a_i^2\), making the ratio larger and hence \(\theta\) smaller.

\subsection{Interpretation}

Eq.~\eqref{eq:cos_lower_bound} shows that, even in the worst case, reducing expert activation from \(K\) to \(m\) changes the output direction by at most
\(\theta_{\max} = \arccos\left( \sqrt{\frac{m}{K}} \right)\).
When \(m/K\) is reasonably large, this angular deviation is small.
If expert output norms are also similar, the change in output magnitude \(\|y'\|\) relative to \(\|y\|\) is negligible, so the primary effect of activation reduction is a slight change in direction, explaining the minimal performance drop empirically observed.

\subsection{The Similarity of Experts Outputs}

\begin{figure*}[!t]
  \centering
  \begin{subfigure}{\textwidth}
    \centering
    \includegraphics[width=0.95\textwidth]{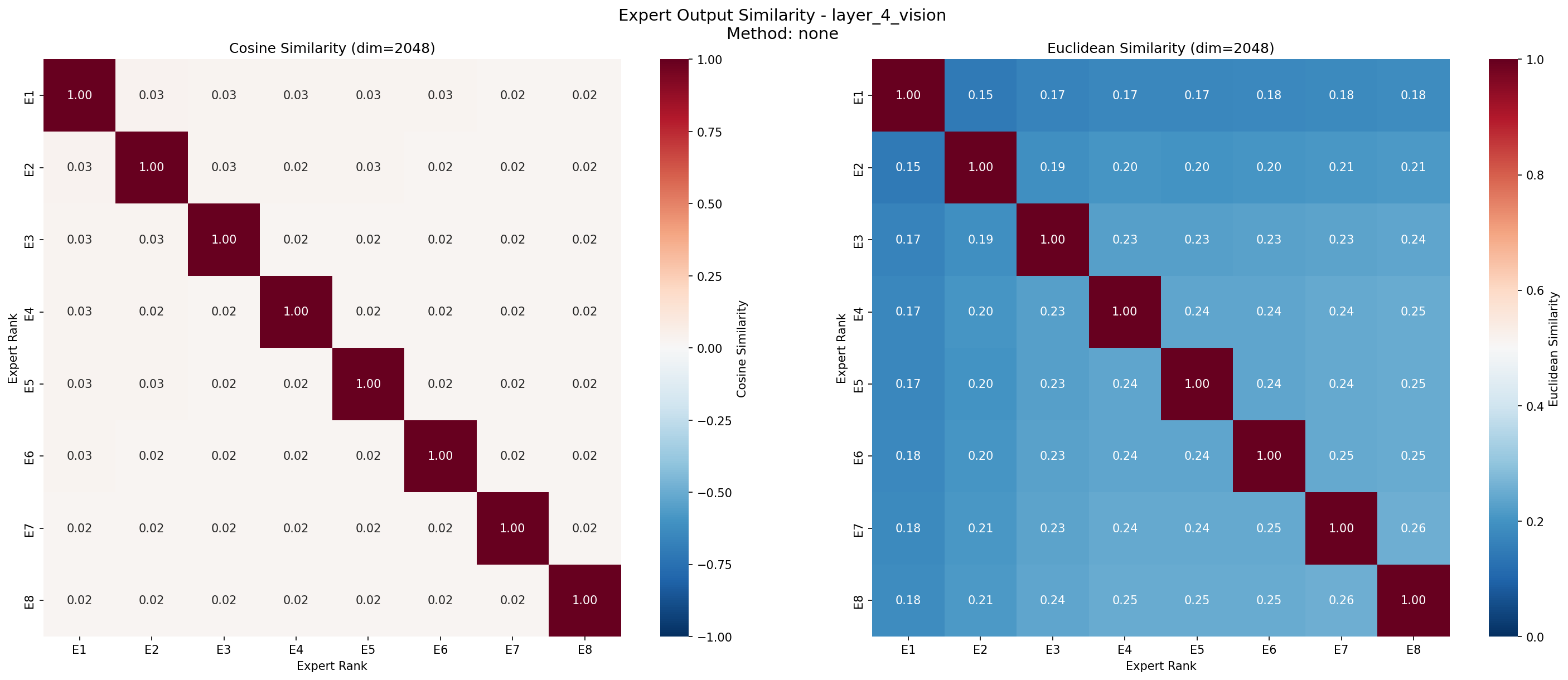}
  \end{subfigure}

\begin{subfigure}{\textwidth}
    \centering
    \includegraphics[width=0.95\textwidth]{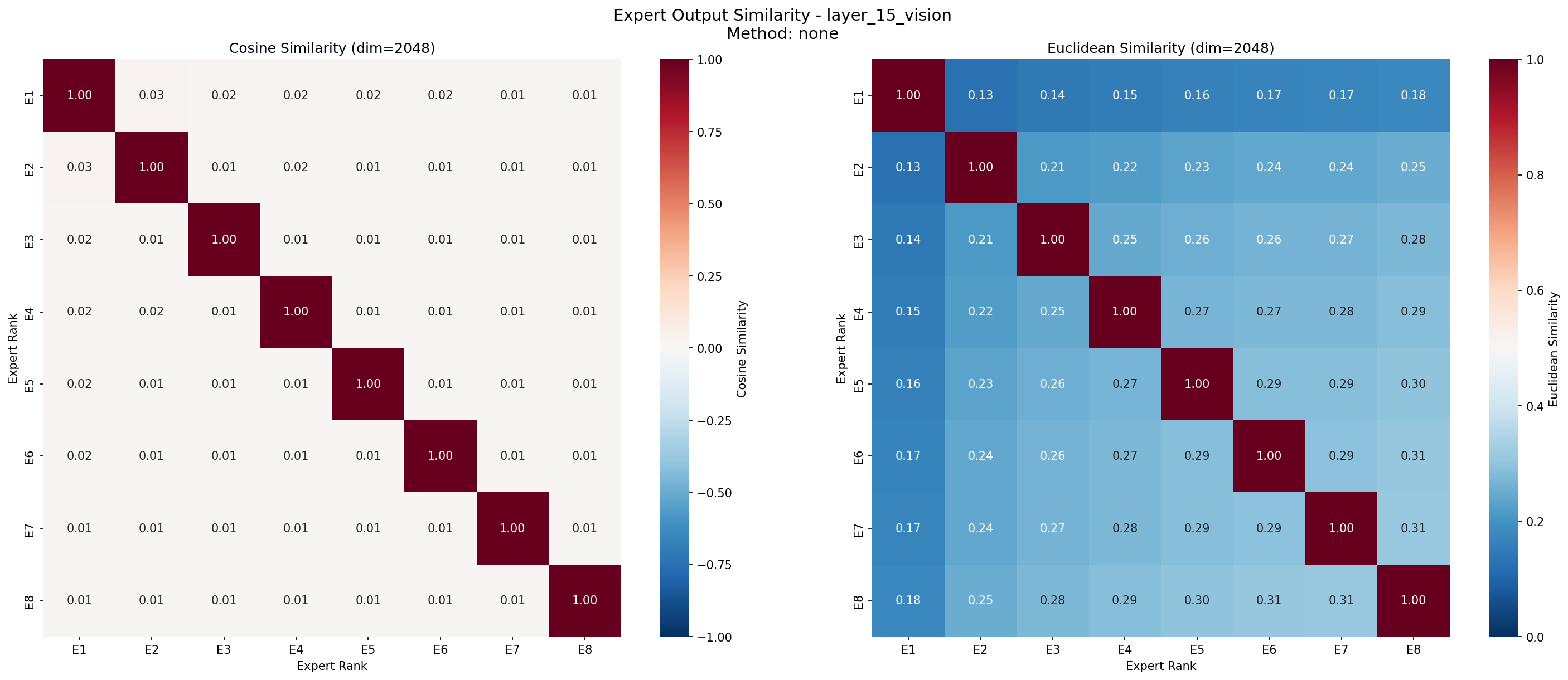}
\end{subfigure}

\begin{subfigure}{\textwidth}
    \centering
    \includegraphics[width=0.95\textwidth]{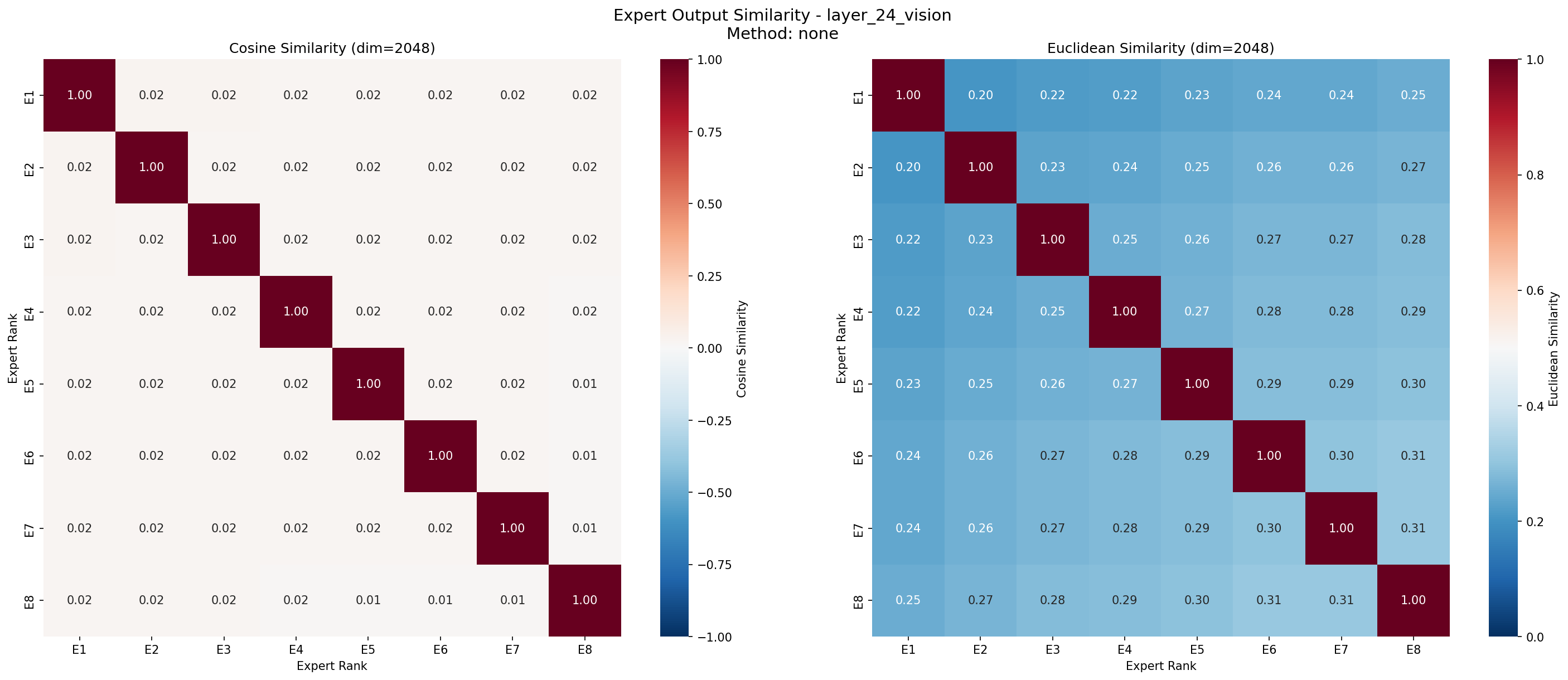}
\end{subfigure}

  \caption{The cosine similarity of experts outputs across layers without PCA.} % 整个figure的总标题
  \label{fig:experts_similarity_nonpca} % 整个figure的标签 (建议修改以区别于之前的标签)
\end{figure*}

We measure the pairwise similarity between expert outputs for the vision modality across multiple layers in InternVL3.5. 
Following the definition $\mathrm{EuclideanSim} = \frac{1}{1+\mathrm{Dist}}$, \cref{fig:experts_similarity_nonpca} shows cosine and Euclidean similarities computed directly in the high-dimensional (2048-d) output space. 
In most layers, the cosine similarity between different experts is close to zero, indicating near-orthogonality and supporting the orthogonality assumption adopted in our theoretical analysis (\S\ref{sec:magnitude_concentration}). 

\begin{figure*}[!t]
  \centering
  \begin{subfigure}{\textwidth}
    \centering
    \includegraphics[width=0.95\textwidth]{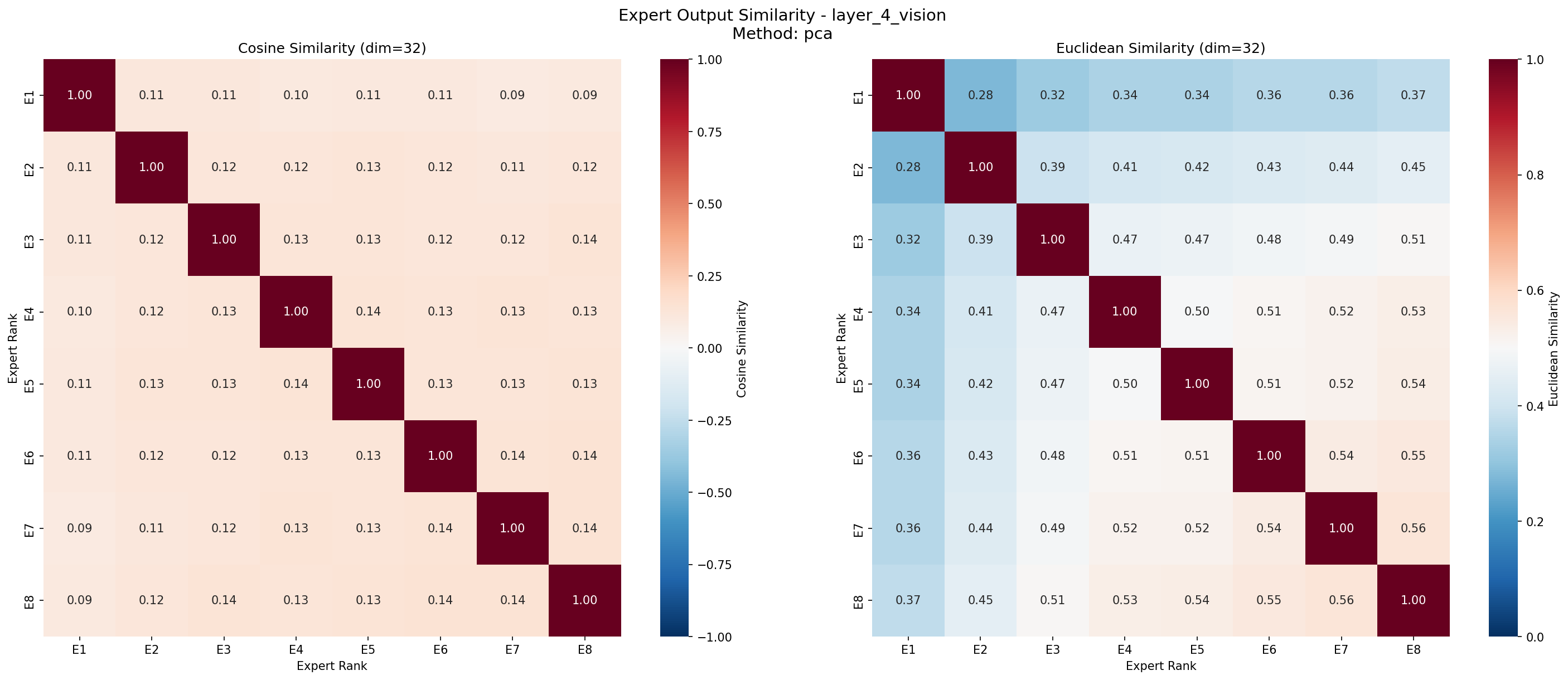}
  \end{subfigure}

\begin{subfigure}{\textwidth}
    \centering
    \includegraphics[width=0.95\textwidth]{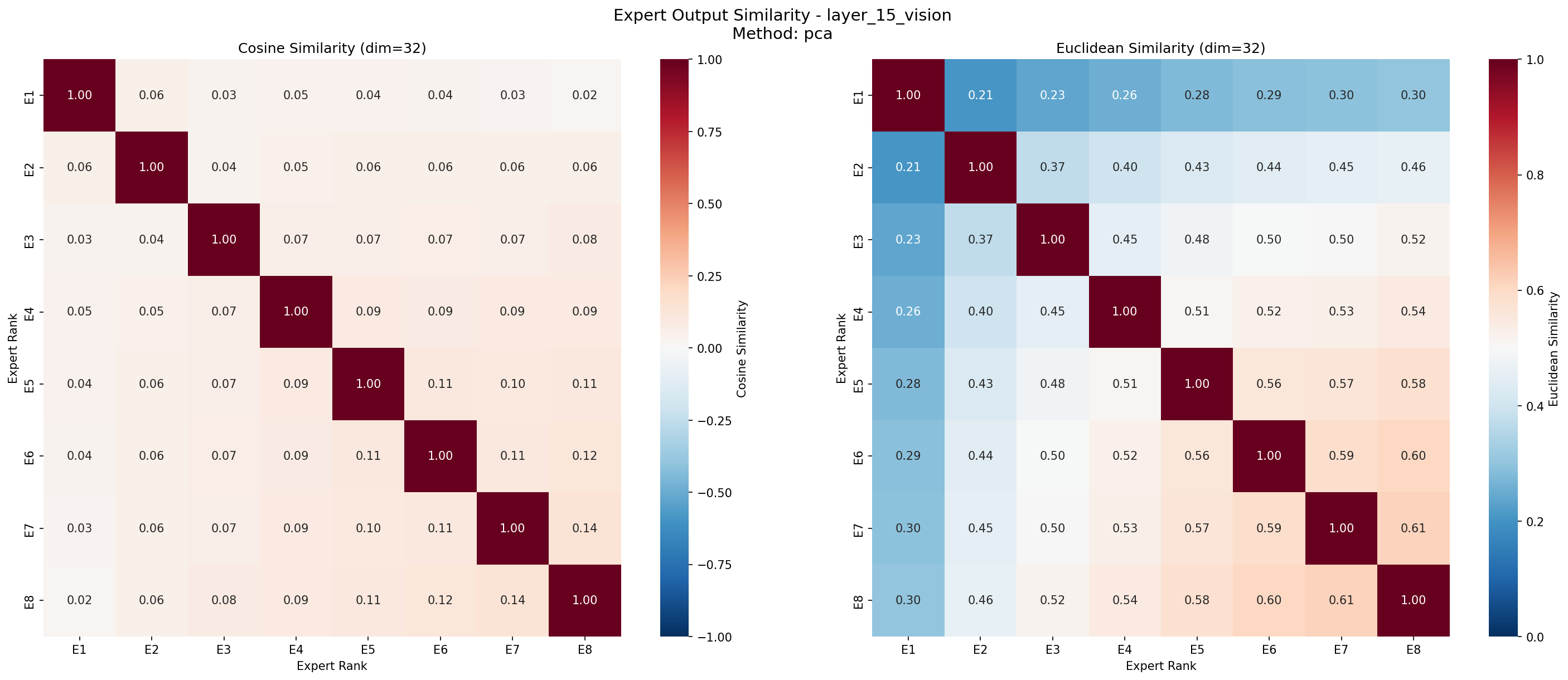}
\end{subfigure}

\begin{subfigure}{\textwidth}
    \centering
    \includegraphics[width=0.95\textwidth]{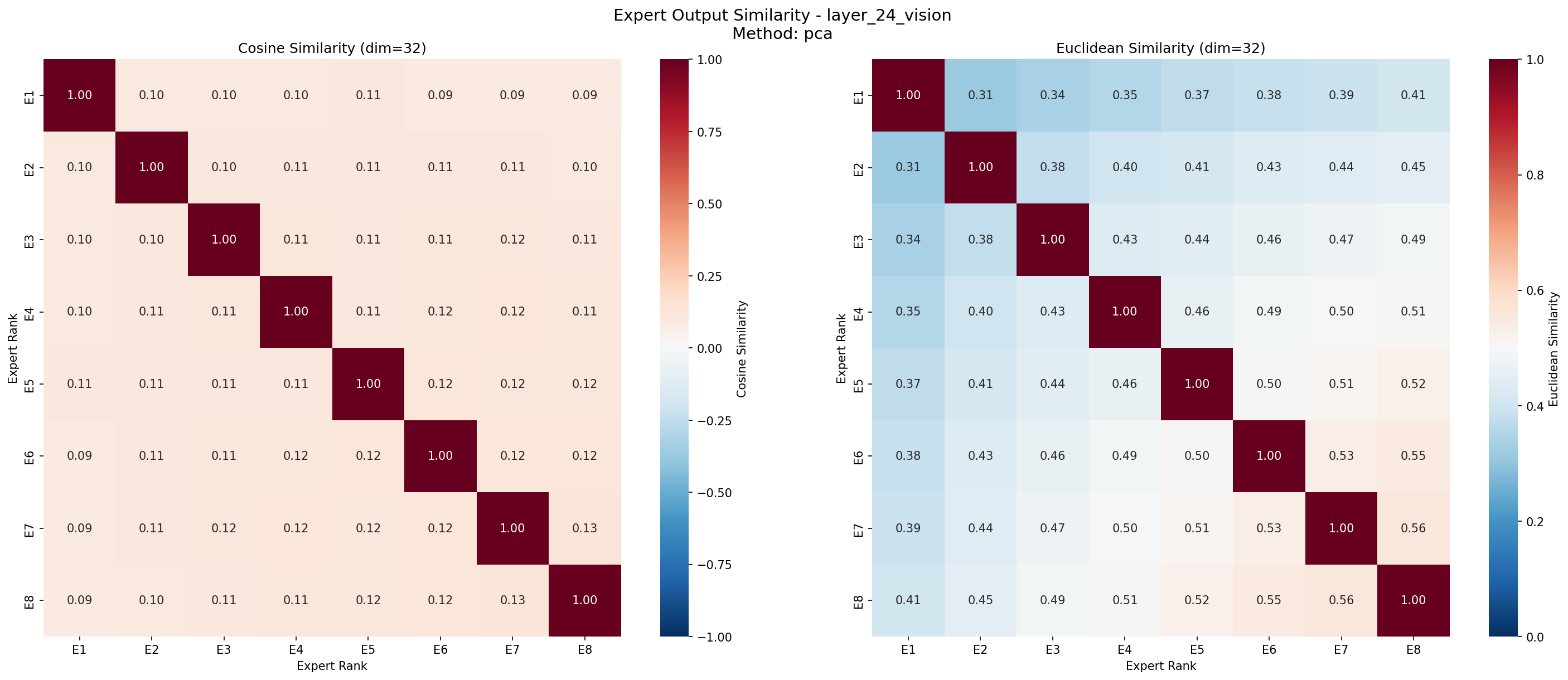}
\end{subfigure}

  \caption{The cosine similarity of experts outputs across layers with PCA.} % 整个figure的总标题
  \label{fig:experts_similarity_pca} % 整个figure的标签 (建议修改以区别于之前的标签)
\end{figure*}

However, the observation that expert outputs are entirely unrelated in semantics would be counter-intuitive. 
It is well known in high-dimensional statistics that sparse vectors tend to appear orthogonal by chance. 
To better reveal potential semantic correlations, we perform PCA to project expert outputs to a 32-dimensional subspace before computing similarities (\cref{fig:experts_similarity_pca}). 
After dimensionality reduction, the pairwise cosine similarities increase noticeably (0.05--0.14), and Euclidean similarities also rise (up to $\approx0.6$ in some cases), suggesting that experts may share certain semantic components in lower-dimensional spaces.

This result implies that while experts appear nearly orthogonal in the original output space, their features are not completely independent in the task-relevant subspace. 
Consequently, when reducing the number of activated experts, the remaining top-weight experts could partially compensate for the removed ones, resulting in an even smaller deviation angle between $y$ and $y'$ than predicted by the worst-case bound in Eq.~\eqref{eq:cos_lower_bound}.

\subsection{Case Study on Different Models}

\begin{figure}[H]
\centering
  \includegraphics[width=\columnwidth,keepaspectratio]{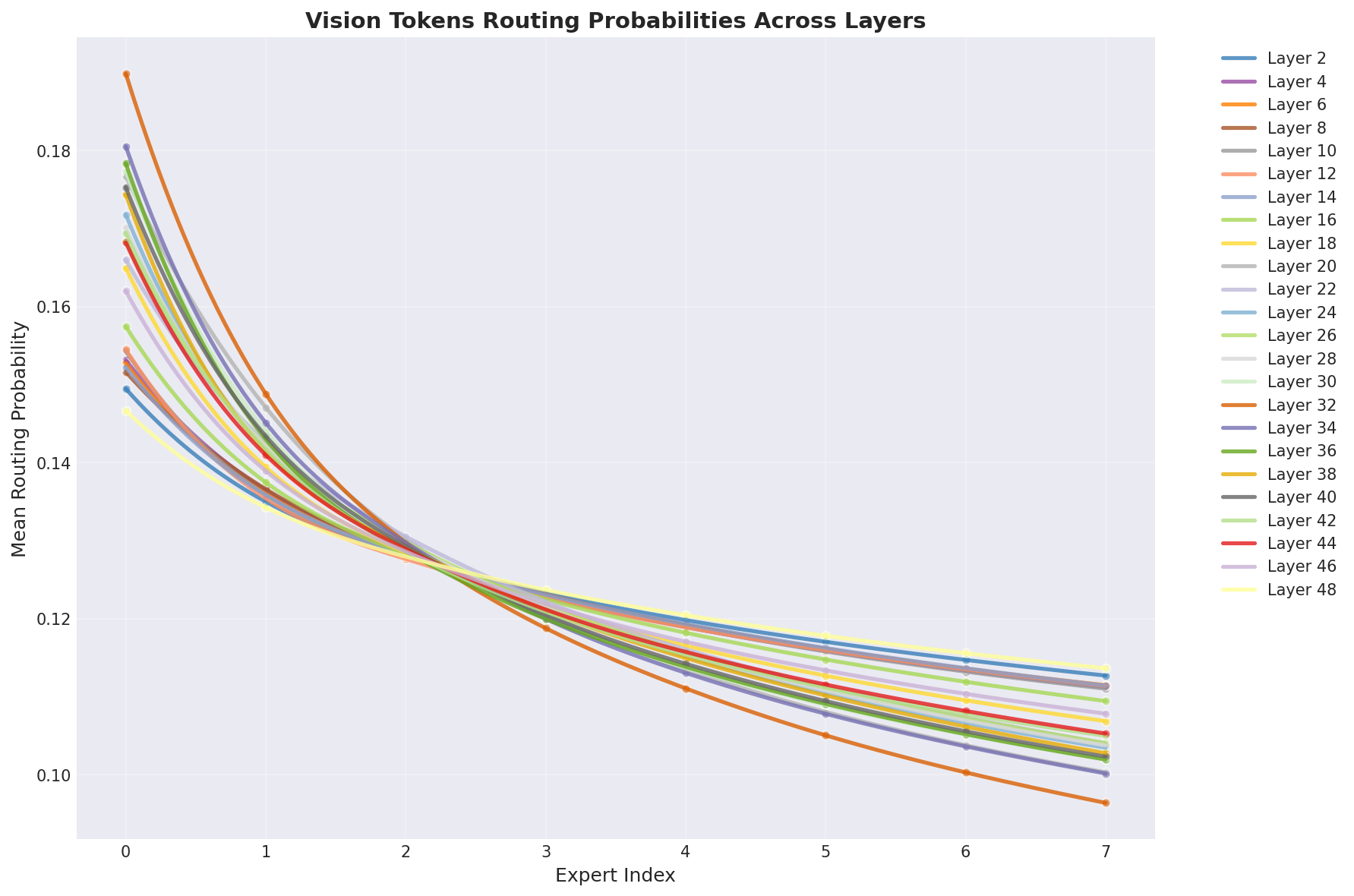}
  \caption{\textbf{InternVL3.5 Vision Tokens – Top-$K$ Routing Probabilities Across Layers.} 
  Mean normalized routing probabilities for the Top-$K$ experts ($K=8$) assigned to vision tokens, 
  sorted in descending order per layer. The first index represents the highest-probability expert, 
  revealing the distribution concentration towards high-weight experts across layers.}
  \label{fig:internvl_topk_logits_vision}
\end{figure}

\begin{figure}[H]
\centering
  \includegraphics[width=\columnwidth,keepaspectratio]{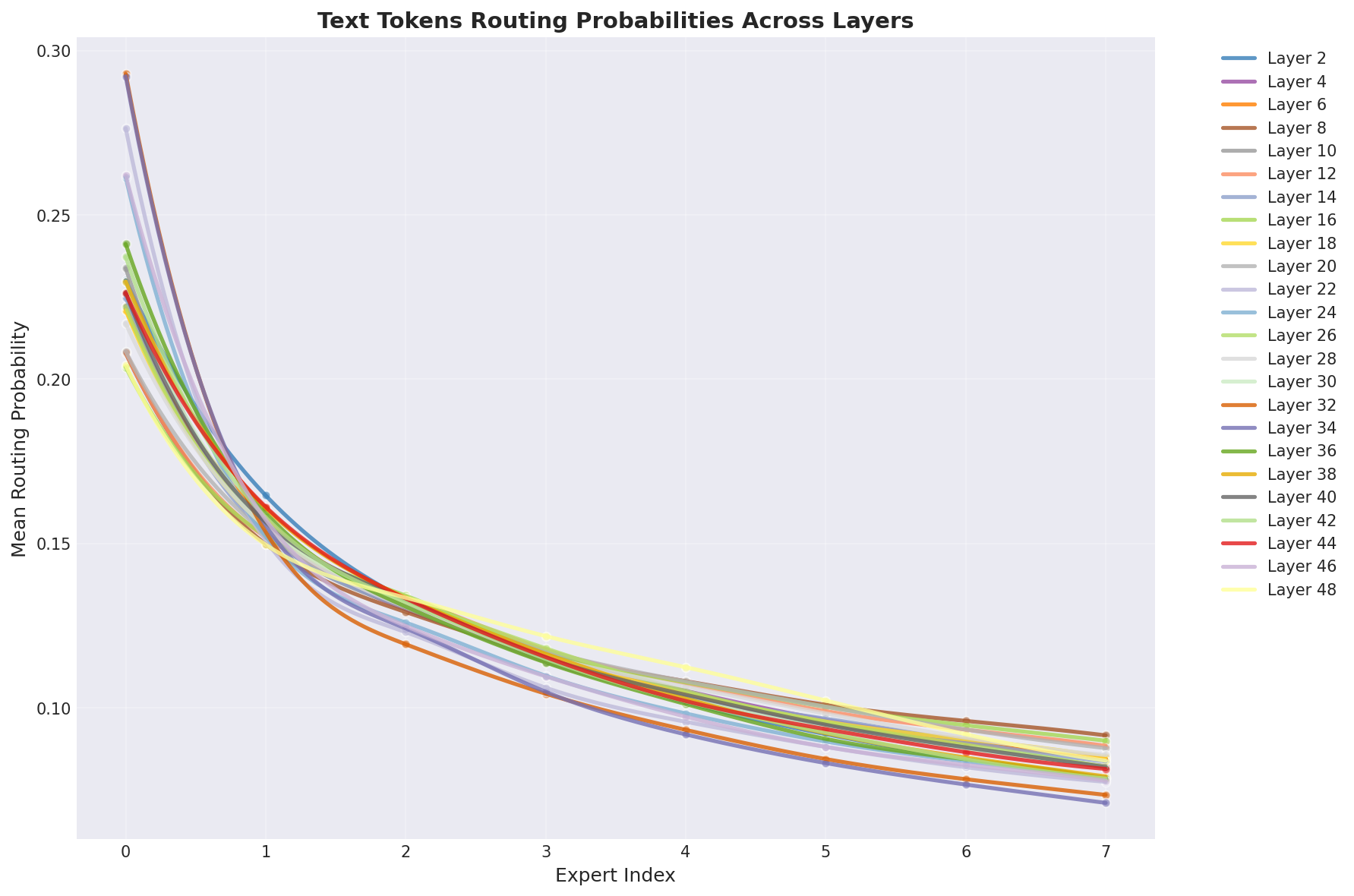}
  \caption{\textbf{InternVL3.5 Text Tokens – Top-$K$ Routing Probabilities Across Layers.} 
  Similar to Figure~\ref{fig:internvl_topk_logits_vision}, but for text tokens. 
  Text tokens display consistently higher Top-1 and Top-2 routing probabilities, 
  indicating stronger expert activation concentration compared to vision tokens.}
  \label{fig:internvl_topk_logits_text}
\end{figure}

\begin{figure}[!h]
\centering
  \includegraphics[width=\columnwidth,keepaspectratio]{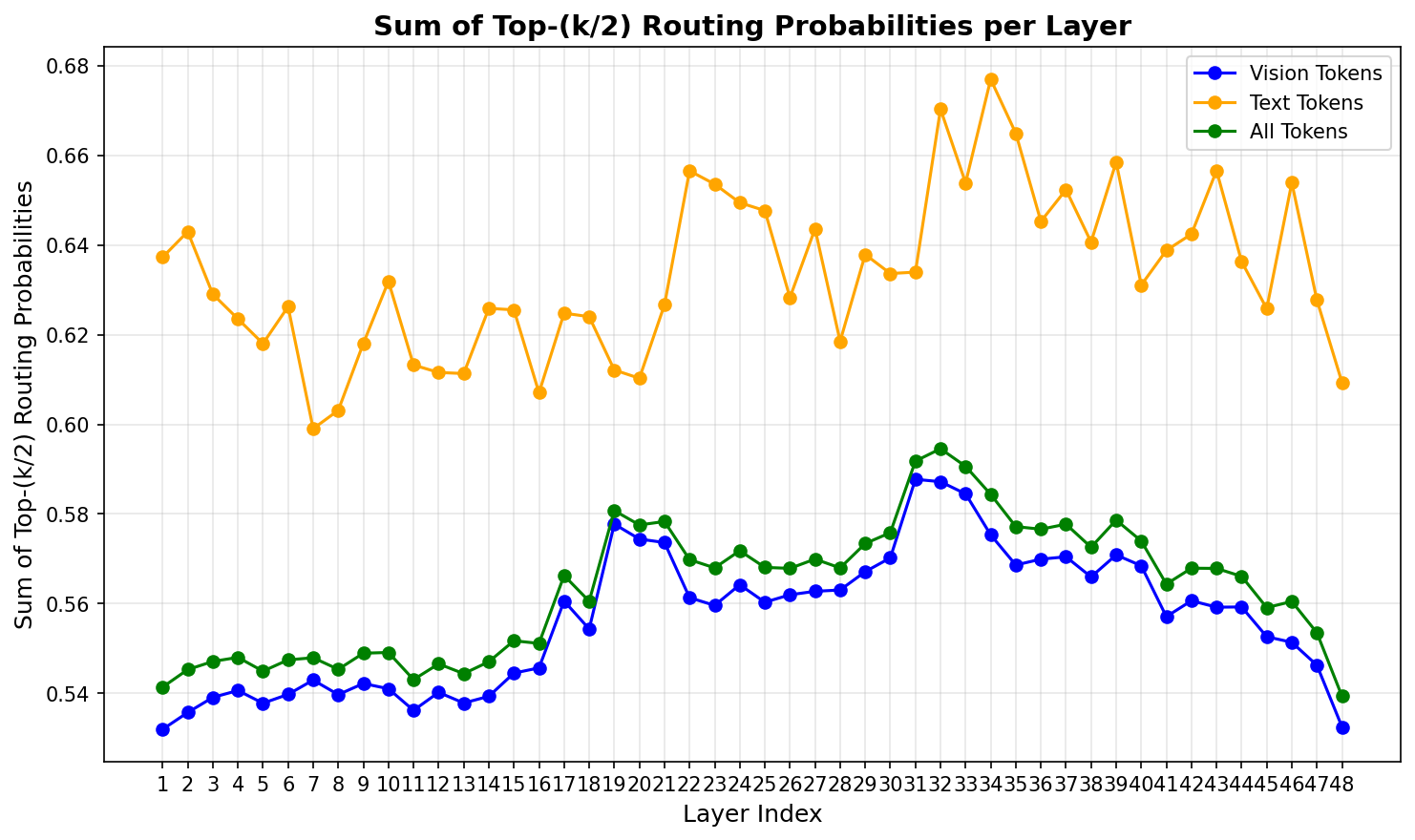}
  \caption{\textbf{InternVL3.5 – Sum of Top-$(K/2)$ Routing Probabilities Per Layer.} 
  Layer-wise average sum of the Top-$(K/2)$=$4$ normalized routing probabilities for vision tokens (blue), 
  text tokens (orange), and all tokens (green). Text tokens maintain higher concentration ($>0.60$) 
  than vision tokens ($\approx0.54$–$0.58$) in most layers.}
  \label{fig:internvl_topk_sum}
\end{figure}

\begin{figure}[!h]
\centering
  \includegraphics[width=\columnwidth,keepaspectratio]{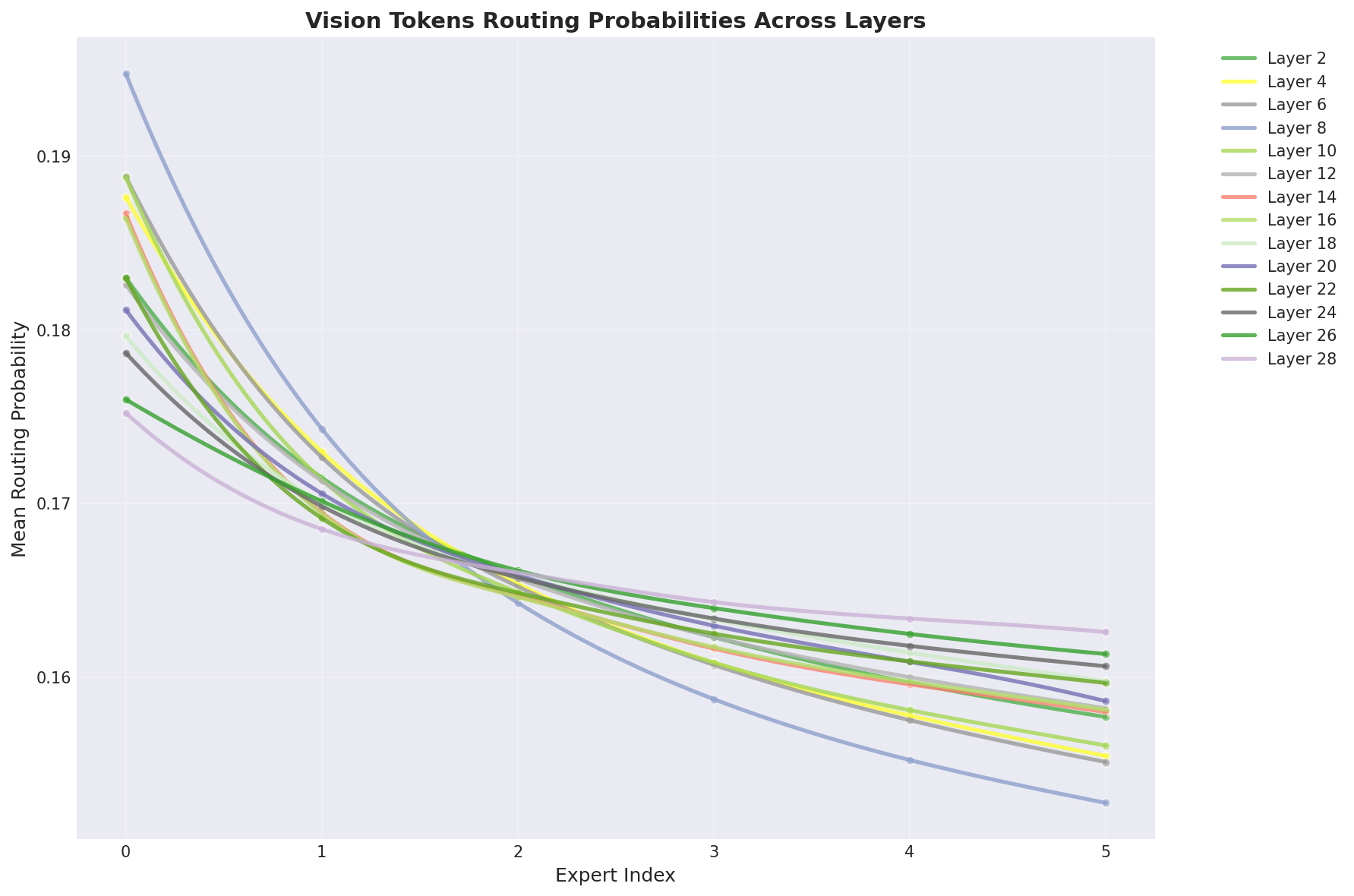}
  \caption{\textbf{DeepSeek-VL2 Vision Tokens – Top-$K$ Routing Probabilities Across Layers.} 
  Mean normalized routing probabilities for the Top-$K$=$6$ experts activated by vision tokens, 
  sorted per layer. Probability drops from the highest to lowest experts are milder than in text tokens, 
  reflecting more uniform expert usage.}
  \label{fig:deepseek_topk_logits_vision}
\end{figure}

\begin{figure}[!h]
\centering
  \includegraphics[width=\columnwidth,keepaspectratio]{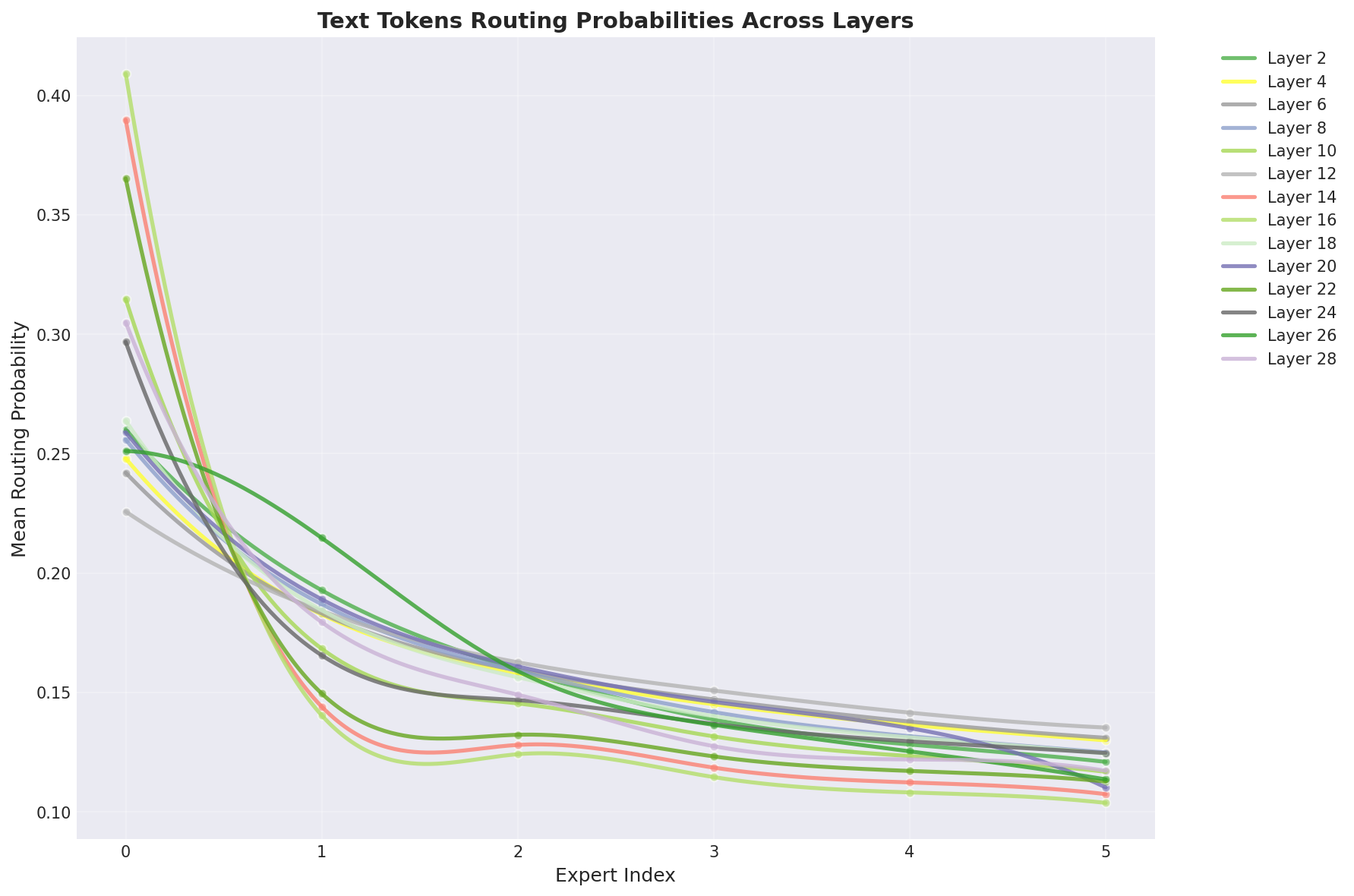}
  \caption{\textbf{DeepSeek-VL2 Text Tokens – Top-$K$ Routing Probabilities Across Layers.} 
  Similar to Figure~\ref{fig:deepseek_topk_logits_vision}, but for text tokens. 
  Text tokens reach Top-1 probabilities of $0.35$–$0.41$ in certain layers (e.g., Layer~16, Layer~18), 
  indicating strong expert activation concentration.}
  \label{fig:deepseek_topk_logits_text}
\end{figure}

\begin{figure}[!h]
\centering
  \includegraphics[width=\columnwidth,keepaspectratio]{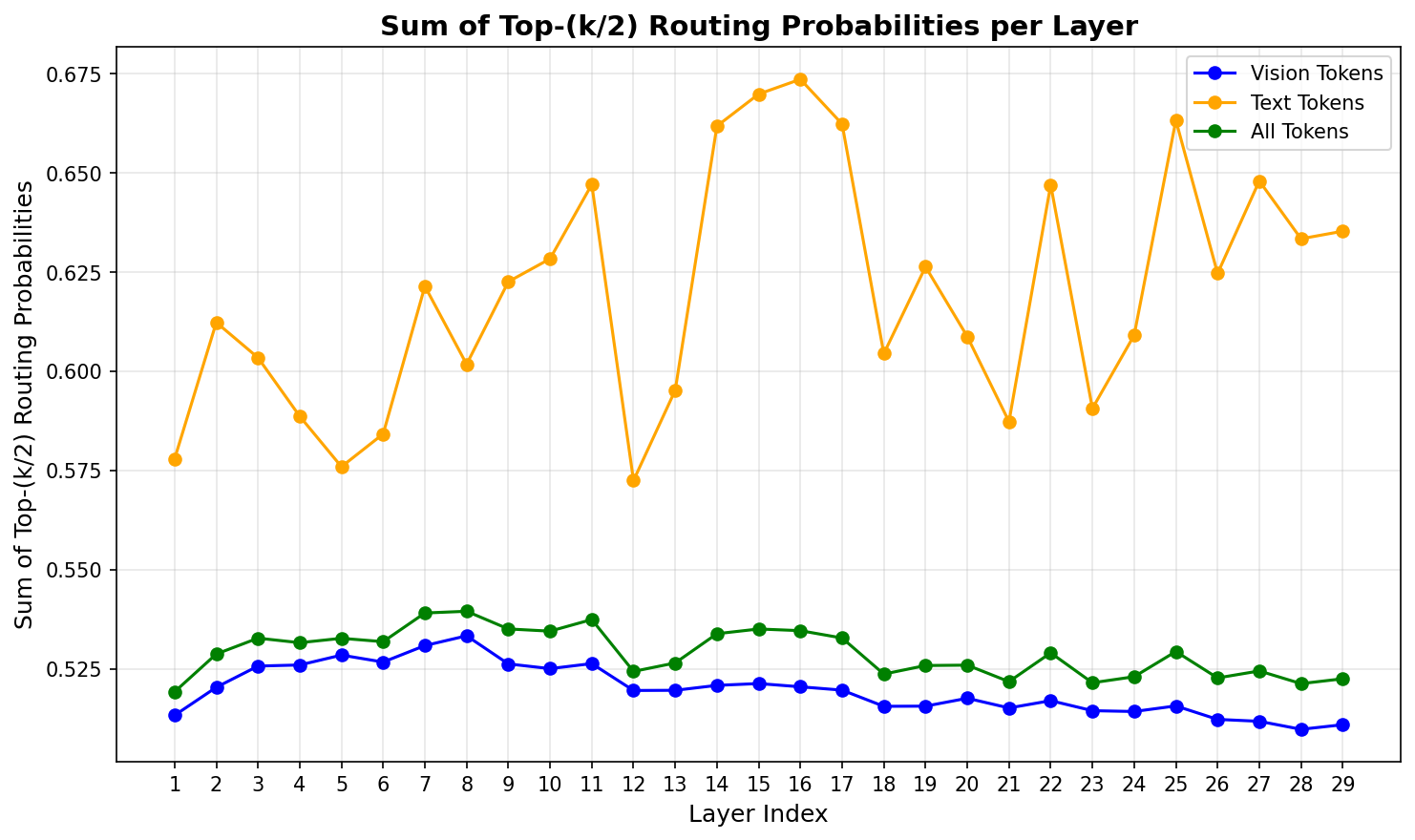}
  \caption{\textbf{DeepSeek-VL2 – Sum of Top-$(K/2)$ Routing Probabilities Per Layer.} 
  Layer-wise average sum of the Top-$(K/2)$=$3$ normalized routing probabilities for vision tokens (blue), 
  text tokens (orange), and all tokens (green). Text tokens maintain $0.58$–$0.67$ range, 
  whereas vision tokens remain stable around $\approx0.52$–$0.53$.}
  \label{fig:deepseek_topk_sum}
\end{figure}

Figures~\ref{fig:internvl_topk_logits_vision}–\ref{fig:internvl_topk_sum} present the routing-probability distributions for \textbf{InternVL3.5}, 
while Figures~\ref{fig:deepseek_topk_logits_vision}–\ref{fig:deepseek_topk_sum} show the corresponding statistics for \textbf{DeepSeek-VL2}. 
For each model, we visualize the mean normalized routing probabilities of Top-$K$ experts (sorted in descending order) 
for both \emph{vision} and \emph{text} tokens across multiple layers, and plot the layer-wise sum of Top-$(K/2)$ routing probabilities.

From the InternVL3.5 results (\cref{fig:internvl_topk_logits_vision,fig:internvl_topk_logits_text}), we observe that:
\begin{itemize}
    \item Text tokens have significantly higher Top-1 and Top-2 routing probabilities than vision tokens across nearly all layers, 
    with Top-1 probabilities often exceeding $0.27$–$0.30$, versus $0.16$–$0.19$ for vision tokens.
    \item Vision-token curves are more uniform across experts (smaller drop from Top-1 to later experts), 
    implying weaker concentration in a few high-weight experts and a more distributed expert utilization.
    \item This difference is quantitatively confirmed in \cref{fig:internvl_topk_sum}, 
    where text tokens maintain a Top-$(K/2)$ sum well above $0.60$, 
    substantially higher than the $\approx 0.54$–$0.58$ range for vision tokens.
\end{itemize}
According to our theoretical framework in \S\ref{sec:magnitude_concentration}, such weaker concentration for vision tokens suggests that reducing the number of activated experts (e.g., halving $K$) will induce even smaller angular deviation (Eq.~\ref{eq:cos_lower_bound}) and minimal norm change, thus preserving accuracy.

For DeepSeek-VL2 (\cref{fig:deepseek_topk_logits_vision,fig:deepseek_topk_logits_text}), the gap between text and vision tokens shows a similar but \emph{less pronounced} pattern:
\begin{itemize}
    \item Text tokens still show a steep drop from Top-1 to subsequent experts, with Top-1 probabilities up to $0.35$–$0.41$ in certain layers (e.g., Layer~16, Layer~18), indicating strong expert activation concentration.
    \item Vision tokens’ Top-1 probabilities are slightly higher relative to InternVL3.5 (around $0.17$–$0.19$ in many layers), possibly due to the presence of two permanently activated \textbf{shared experts}, which biases the routing distribution towards high-weight slots.
    \item In \cref{fig:deepseek_topk_sum}, text tokens retain a Top-$(K/2)$ sum in the $0.58$–$0.67$ range, while vision tokens are more stable around $\approx 0.52$–$0.53$, with very low variance across layers.
\end{itemize}
The presence of shared experts in DeepSeek-VL2 reduces the relative performance gap among pruning methods (as also noted in \S\ref{sec:test_results}), since dense-model-oriented baselines (FastV, SparseVLM) can still leverage these shared experts for stable performance. 
From the angle-bound perspective, the elevated and stable Top-$(K/2)$ sum for vision tokens means that activation reduction will induce bounded deviation even if $K_v \ll K$.

\paragraph{Conclusions.}
Across both models, text tokens consistently exhibit stronger expert-concentration patterns than vision tokens, reflected in higher Top-1 routing probabilities and larger Top-$(K/2)$ sums. 
This corroborates our \emph{magnitude concentration} and \emph{angular deviation bound} analyses: 
vision tokens’ more uniform routing distributions imply that pruning lower-weight experts yields minimal representational change, 
especially in InternVL3.5. 
In DeepSeek-VL2, shared experts structurally mitigate the impact of activation reduction across modalities, explaining the smaller performance disparity among different pruning baselines observed in experiments.

\subsection{Attention Stability and Representation Drift under Expert Reduction}
\label{sec:representation_drift}

To comprehensively address potential concerns regarding representation drift and cross-modal alignment disruption caused by reducing activated experts, we compared the internal states of the full-expert model ($K_v=8$) with those of the reduced-expert models on the same multimodal inputs using InternVL3.5.

\begin{figure}[!h]
  \centering
  \begin{subfigure}{0.49\linewidth}
    \centering
    \includegraphics[width=\linewidth]{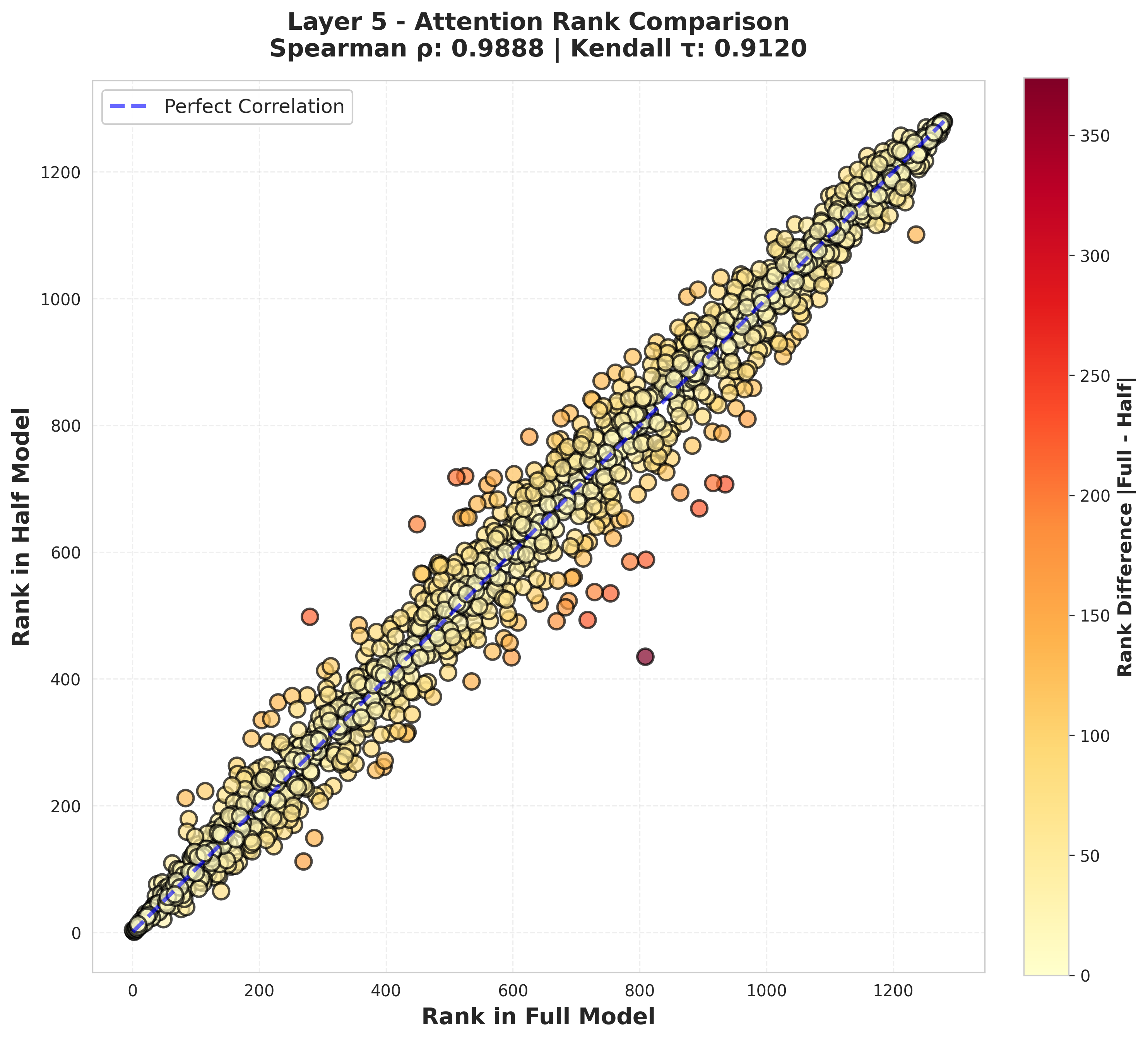}
    \caption{Layer 5}
  \end{subfigure}
  \hfill
  \begin{subfigure}{0.49\linewidth}
    \centering
    \includegraphics[width=\linewidth]{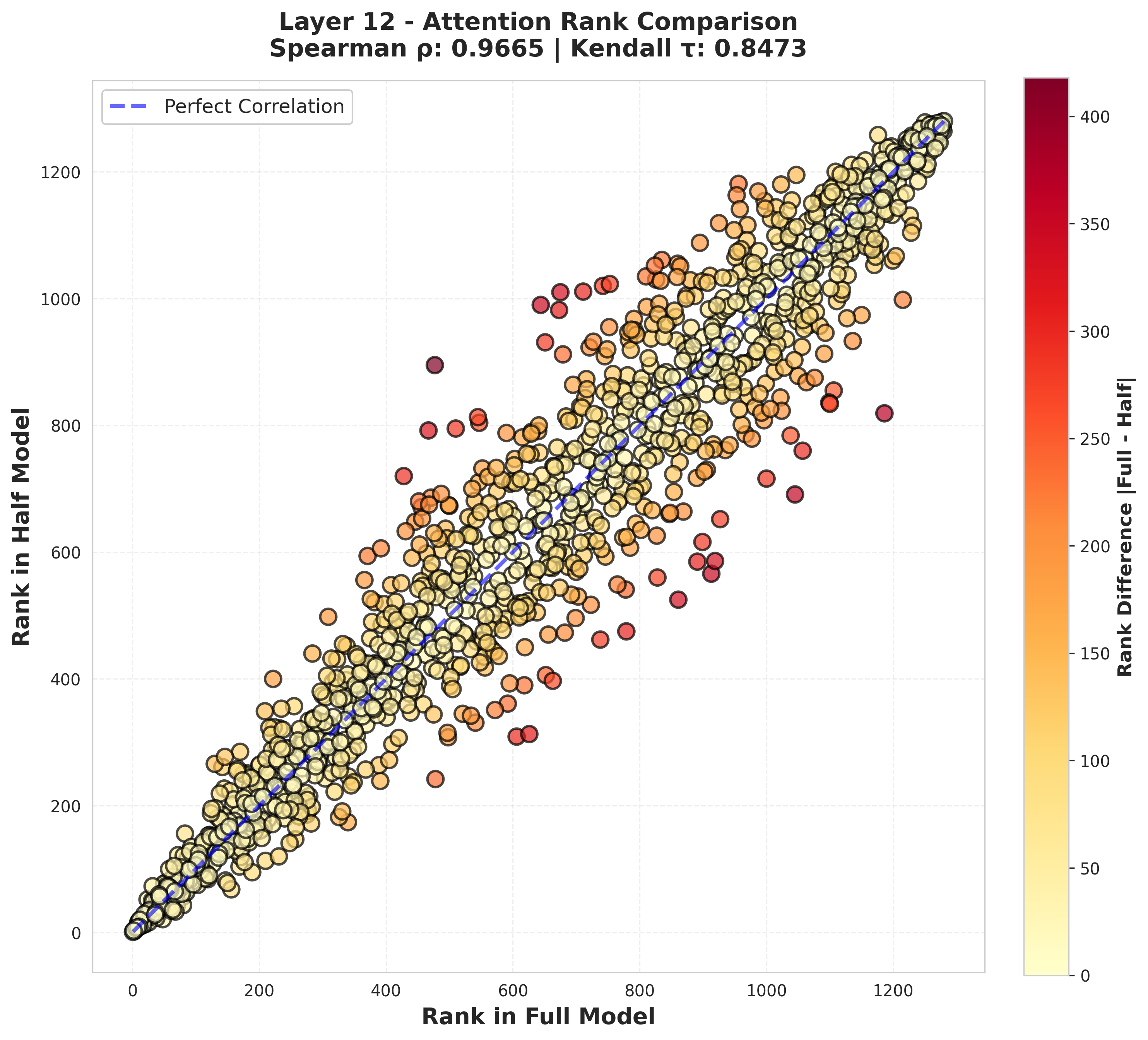}
    \caption{Layer 12}
  \end{subfigure}
  \vspace{-0.5em}
  \caption{\textbf{Attention Stability under Expert Reduction.} The Spearman rank correlation of text-to-image attention weights remains highly stable ($>0.96$) after reducing active experts, confirming that cross-modal alignment is preserved.}
  \label{fig:rank_correlation_appendix}
\end{figure}

\paragraph{Attention Stability.} 
We first evaluate whether expert activation reduction disrupts the cross-modal attention patterns, which are crucial for our routing-aware token pruning module. We compute the Spearman rank correlation of the text-to-image cross-attention weights between the baseline and the reduced models. As shown in \cref{fig:rank_correlation_appendix}, despite halving the number of active experts, the attention weights at different depths (e.g., Layer 5 and Layer 12) maintain an exceptionally high Spearman correlation ($>0.96$). This strong stability confirms that the original cross-modal semantic alignment is well-preserved, thereby validating the reliability of using attention weights as a metric for token redundancy even after expert reduction.

\begin{figure}[!h]
  \centering
  \includegraphics[width=0.7\columnwidth,keepaspectratio]{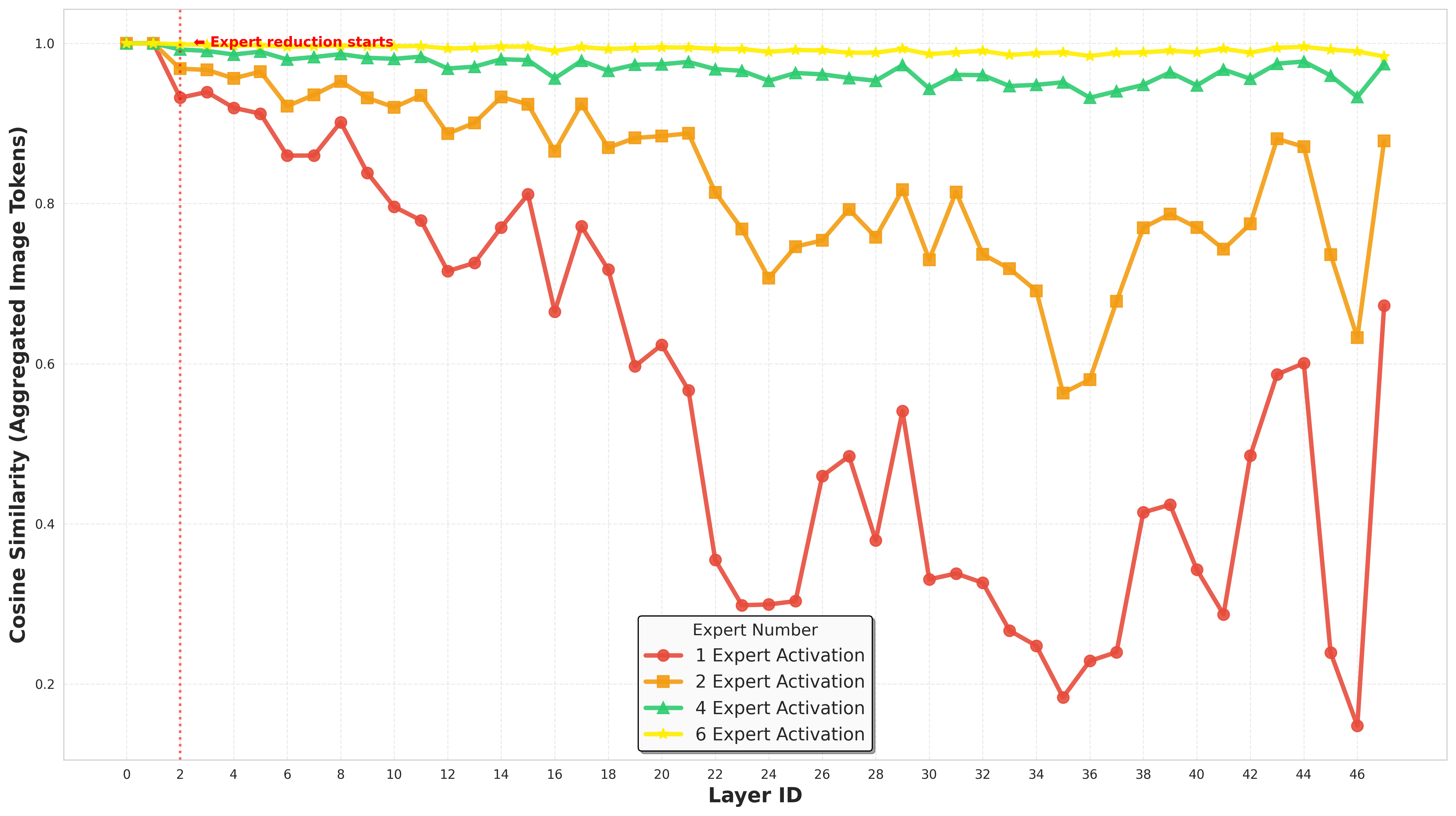}
  \caption{\textbf{Feature Cosine Similarity across Layers.} Aggregated visual hidden states maintain a high similarity ($>0.95$) with the full 8-expert baseline when reduced to 4 experts (green line), demonstrating minimal representation drift.}
  \label{fig:cosine_sim_compare_appendix}
\end{figure}

\paragraph{Feature Representation Drift.} 
Furthermore, we quantify the representation drift by measuring the cosine similarity of the aggregated visual hidden states between the full-expert baseline and reduced variants. Specifically, for a given layer $l$, the aggregated visual hidden state $\mathbf{\bar{h}}_{v}^{(l)}$ is defined as the mean over the hidden states of all $N_v$ visual tokens in the sequence:
\begin{equation}
    \mathbf{\bar{h}}_{v}^{(l)} = \frac{1}{N_v} \sum_{i=1}^{N_v} \mathbf{h}_{i,v}^{(l)},
\end{equation}
where $\mathbf{h}_{i,v}^{(l)}$ denotes the hidden state of the $i$-th visual token at layer $l$. We then compute the cosine similarity between $\mathbf{\bar{h}}_{v, \text{full}}^{(l)}$ and $\mathbf{\bar{h}}_{v, \text{reduced}}^{(l)}$ for each layer. As illustrated in \cref{fig:cosine_sim_compare_appendix}, when reducing the active experts to $K_v=4$ (the optimal trade-off point, indicated by the green line), the visual hidden states maintain a high cosine similarity ($>0.95$) with the full model across the vast majority of layers. This minimal representation drift provides a direct internal explanation for the negligible downstream performance degradation observed in \cref{tab:internvl_reduce_act_all}. Conversely, extreme reduction scenarios ($K_v=1$ or $2$) lead to a sharp decline in cosine similarity, indicating significant feature collapse, which aligns with the sharp drop in benchmark accuracy for those settings.

\section{Theoretical Analysis on Merge Rate \texorpdfstring{$\gamma$}{γ}}
\label{sec:gamma_theory}

This section provides the theoretical derivation of the merge rate $\gamma$ used in the routing-aware token pruning module of \textbf{FastMMoE}. 
While the main text (\S\ref{sec:ablation}) summarizes the empirical effects of different $\gamma$ values, 
here we formalize the computation of its theoretical upper bound under multi-stage pruning to support the experimental configurations.

\subsection{Upper Bound Derivation}

The merge rate $\gamma$ determines the proportion of retained vision tokens that are generated by merging high-redundancy windows, as defined in \cref{con:merge_rate}. 
In our framework, $\gamma$ is upper-bounded by the extreme case in which no tokens are directly dropped; instead, all windows exceeding the redundancy threshold are merged into compressed representations until the target number of vision tokens $\eta_l$ is reached. 
This upper bound can be derived as:
\begin{equation}
    \hat{\gamma} = \frac{\frac{N_v^l - \eta_l}{W-1}}{\eta_l} 
                 = \frac{\frac{N_v^l}{\eta_l} - 1}{W-1}
                 = \frac{\frac{1}{\beta} - 1}{W-1},
\end{equation}
where $N_v^l$ is the number of vision tokens before pruning at layer $l$, 
$W$ denotes the window size, and $\beta = \frac{\eta_l}{N_v^l}$ is the per-layer pruning retain ratio.

\subsection{Multi-Stage Pruning Formulation}

FastMMoE adopts a multi-stage pruning scheme similar to SparseVLM, applying token reduction at three designated layers. 
If each stage uses the same per-stage retain ratio $\beta$, the cumulative retain ratio $r$ after three stages can be expressed as:
\begin{equation}
    r = \beta^3, \quad \beta = r^{\frac{1}{3}}.
\end{equation}
For target overall retain ratios of 75\%, 50\%, and 25\%, the corresponding $\beta$ values are approximately $0.91$, $0.79$, and $0.63$. 
Assuming a window size $W = 5$, the maximum theoretical $\gamma$ values $\hat{\gamma}$ for each setting are therefore $0.025$, $0.05$, and $0.15$, respectively.

\begin{table}[!h]
\centering
\caption{\textbf{Theoretical upper bounds of the merge rate $\hat{\gamma}$}. 
Calculated for three target overall retain ratios (75\%, 50\%, 25\%) under a three-stage pruning setup with equal per-stage ratio $\beta$ and window size $W=5$.}
\label{tab:gamma_upper_bound_appendix}
\begin{tabular}{@{}c|c|c|c@{}}
\toprule
\textbf{$W$} & \textbf{Retain Ratio} & \textbf{$\beta$} & \textbf{$\hat{\gamma}$} \\ \midrule
\multirow{3}{*}{5} 
& 75\% & 0.91 & 0.025 \\ \cmidrule(l){2-4} 
& 50\% & 0.79 & 0.05  \\ \cmidrule(l){2-4} 
& 25\% & 0.63 & 0.15  \\ \bottomrule
\end{tabular}%
\end{table}

\subsection{Experimental Setup Based on Theoretical Bounds}

In the ablation study (Table~\ref{tab:merge_ratio_ablation} in the main paper), 
the $\gamma$ values are selected according to these theoretical bounds to ensure interpretability and consistency:  
for 75\% retention, only $\gamma=0.025$ is tested (its maximal feasible value);  
for 50\% retention, $\gamma \in \{0.025, 0.05\}$;  
and for 25\% retention, $\gamma \in \{0.025, 0.05, 0.1, 0.15\}$ to reflect the effect of stronger compression.  

This theoretical analysis provides quantitative support for the chosen grid of merge rates in our experiments and helps explain the empirical observation that 
moderate $\gamma$ values near the upper bound achieve a superior trade-off between token reduction and semantic fidelity.

\section{FLOPs Analysis}
\label{sec:flops}

In this section, we provide a complete theoretical derivation of the FLOPs savings
under three configurations:
(1) \emph{token pruning only},
(2) \emph{expert activation reduction only},
(3) \emph{combined token pruning and activation reduction}.

\subsection{Layer Index Ranges}
Transformer layer index ranges for the two target architectures are:
\begin{itemize}
    \item \textbf{DeepSeek-VL2}: $30$ layers, indexed from $0$ (first layer) to $29$ (last layer).
    \item \textbf{InternVL3.5-30B-A3B}: $48$ layers, indexed from $0$ to $47$.
\end{itemize}
All pruning layer indices in the formulas refer to these ranges.

\subsection{Notation}
\begin{itemize}
    \item $B$: batch size (set to $B=1$).
    \item $L$: total token length (vision + text).
    \item $L_v$: number of vision tokens before pruning.
    \item $H$: hidden size; $A$: attention head count; $d$: head dimension ($H=A\cdot d$).
    \item $S$: intermediate hidden size for dense MLP.
    \item $S_m$: intermediate hidden size for MoE experts.
    \item $E$: number of non-shared experts.
    \item $K$: baseline active experts per token.
    \item $K_v$: active experts per vision token post-reduction; $p=K_v/K$ activation ratio.
    \item $l_v$: first layer to apply activation reduction (inclusive).
    \item $\beta$: per-stage vision-token retention ratio after pruning.
    \item $C(x)$: per-layer FLOPs for a vision-token subsequence of length $x$ in an MoE layer.
    \item $C_0, C_r, C_{r^2}, C_{r^3}$: shorthand for $C(L_v)$, $C(\beta L_v)$, $C(\beta^2 L_v)$, $C(\beta^3 L_v)$.
    \item $D_0, D_r, D_{r^2}, D_{r^3}$: FLOPs adjustments due to reduced expert activation at different pruning stages.
    \item $m_0, m_r, m_{r^2}, m_{r^3}$: number of layers in each stage to which activation reduction is applied.
    \item $R$: ratio of total FLOPs after a given optimization (token pruning and/or activation reduction) to the total FLOPs of the baseline model without optimization.  
          $R$ satisfies $0 < R \le 1$.  
          The \textbf{actual FLOPs saving ratio} is $1-R$.
\end{itemize}

\subsection{Baseline FLOPs}
\paragraph{Attention:}
\begin{align}
\mathrm{FLOPs}_{\mathrm{attn}} &\approx 4BL^2H + 8BLH^2.
\end{align}
\paragraph{Dense MLP:}
\begin{align}
\mathrm{FLOPs}_{\mathrm{mlp}} &\approx 6BLHS.
\end{align}
\paragraph{MoE FFN:}
\begin{align}
\mathrm{FLOPs}_{\mathrm{moe}} &\approx 2BLHE + 6BLHS_mK.
\end{align}

\subsection{DeepSeek-VL2 Formulas}
Pruning at layers $2,5,8$ with per-stage retention $\beta$, $C'(x)$ replaces $K$ with $K_v$.

\paragraph{Token pruning only:}
$R_{\mathrm{prune}}$ is the proportion of FLOPs remaining after applying only vision-token pruning (no activation reduction); actual savings is $1-R_{\mathrm{prune}}$:
\begin{equation}
R_{\mathrm{prune}} =
\frac{6L_vHS + \sum_{t=0}^{3}C(\beta^t L_v)}{6L_vHS + 29C(L_v)}.
\end{equation}

\paragraph{Activation reduction only:}
$R_{\mathrm{act}}$ is the proportion of FLOPs remaining after activation reduction from layer $l_v$ onward, without pruning; savings is $1-R_{\mathrm{act}}$:
\begin{equation}
R_{\mathrm{act}} =
\frac{6L_vHS + l_v C(L_v) + (29-l_v) C'(L_v)}{6L_vHS + 29C(L_v)}.
\end{equation}

\paragraph{Combined:}
$R_{\mathrm{comb}}$ is the FLOPs proportion when both token pruning and activation reduction are applied; savings is $1-R_{\mathrm{comb}}$:
\begin{equation}
R_{\mathrm{comb}} =
\frac{\sum_{s\in\{0,r,r^2,r^3\}}\left[N_s C_s + m_s\cdot\Delta_s\right]}
     {6L_vHS + 29C(L_v)},
\end{equation}
where $N_s$ is layers in stage $s$, $\Delta_s = C'_s - C_s$, and $m_s$ is the number of layers in stage $s$ affected by activation reduction.

\subsection{InternVL3.5-30B-A3B Formulas}
Pruning layers: $5,8,12$ with per-stage retention $\beta$.

\paragraph{Token pruning only:}
$R_{\mathrm{prune}}$ is FLOPs proportion after pruning only; savings is $1-R_{\mathrm{prune}}$:
\begin{equation}
R_{\mathrm{prune}} =
\frac{6C_0 + 3C_r + 4C_{r^2} + 35C_{r^3}}{48C_0}.
\end{equation}

\paragraph{Activation reduction only:}
$R_{\mathrm{act}}$ is FLOPs proportion after activation reduction only; savings is $1-R_{\mathrm{act}}$:
\begin{equation}
R_{\mathrm{act}} =
\frac{l_v C_0 + (48-l_v) C'(L_v)}{48C_0}.
\end{equation}

\paragraph{Combined:}
$R_{\mathrm{comb}}$ is FLOPs proportion when both strategies are applied; savings is $1-R_{\mathrm{comb}}$:
\begin{equation}
R_{\mathrm{comb}} =
\frac{\sum_{s\in\{0,r,r^2,r^3\}}\left[N_s C_s + m_s\cdot\Delta_s\right]}{48C_0},
\end{equation}
where $N_s \in \{6,3,4,35\}$ for $\{C_0,C_r,C_{r^2},C_{r^3}\}$ stages, $m_s$ is computed as defined in Notation, and $\Delta_s = C'_s - C_s$.

\subsection*{Interpretation}
Across all configurations:
\begin{itemize}
    \item $R$ measures the fraction of FLOPs used by the optimized model compared to the unoptimized baseline.
    \item $1-R$ is the actual FLOPs saving ratio.
\end{itemize}
For example, $R=0.55$ means the optimized model consumes $55\%$ of the original FLOPs, corresponding to a $1-R = 0.45$ (i.e., $45\%$ reduction).

\begin{figure}[H]
\centering
  \includegraphics[width=\columnwidth,keepaspectratio]{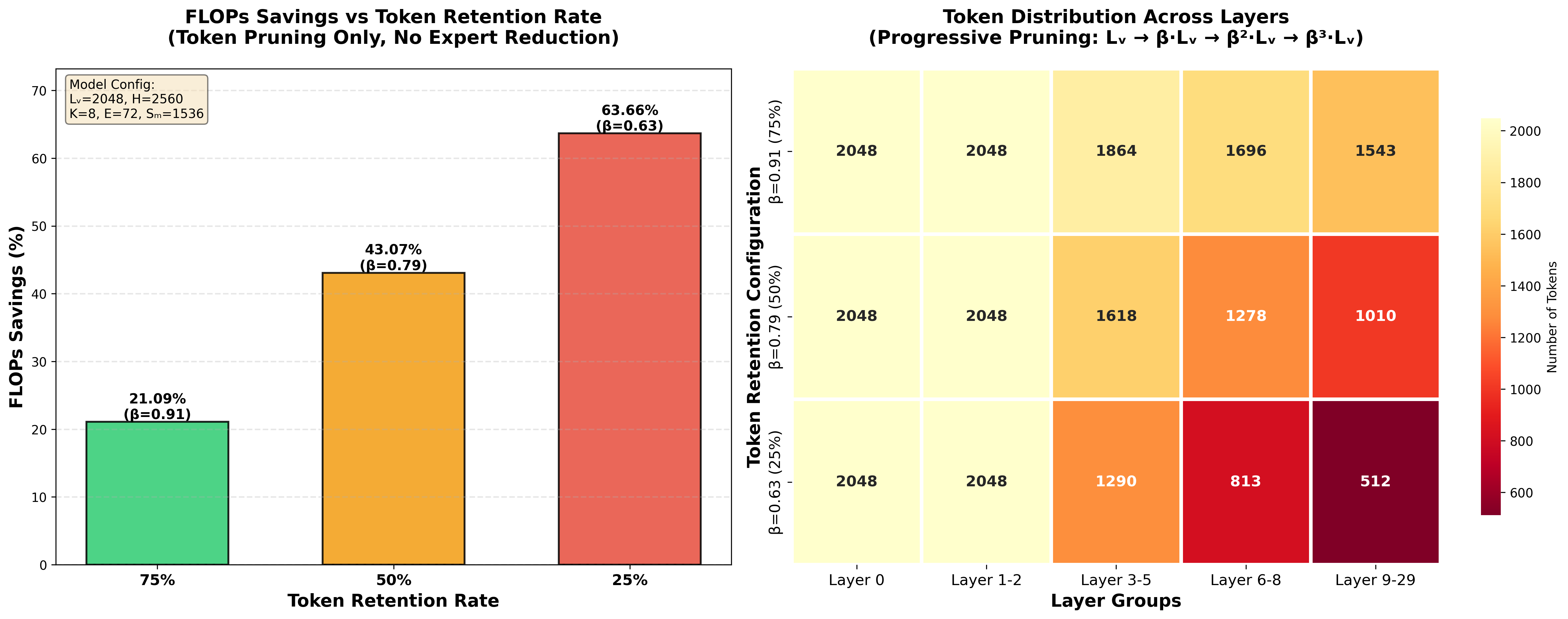}
  \caption{\textbf{FLOPs savings for DeepSeek-VL2 with token pruning only.} 
  (\emph{Left}) FLOPs savings at three token retention rates: 
  $\beta=0.91$ (75\%), $\beta=0.79$ (50\%), and $\beta=0.63$ (25\%). 
  (\emph{Right}) Token distribution across layer groups under progressive pruning strategy 
  ($L_v \to \beta L_v \to \beta^2 L_v \to \beta^3 L_v$). 
  Lower retention rates yield higher FLOPs savings, 
  with 25\% retention achieving 43.74\% FLOPs reduction.}
  \label{fig:deepseek_flops_formula1}
\end{figure}

\begin{figure}[H]
\centering
  \includegraphics[width=\columnwidth,keepaspectratio]{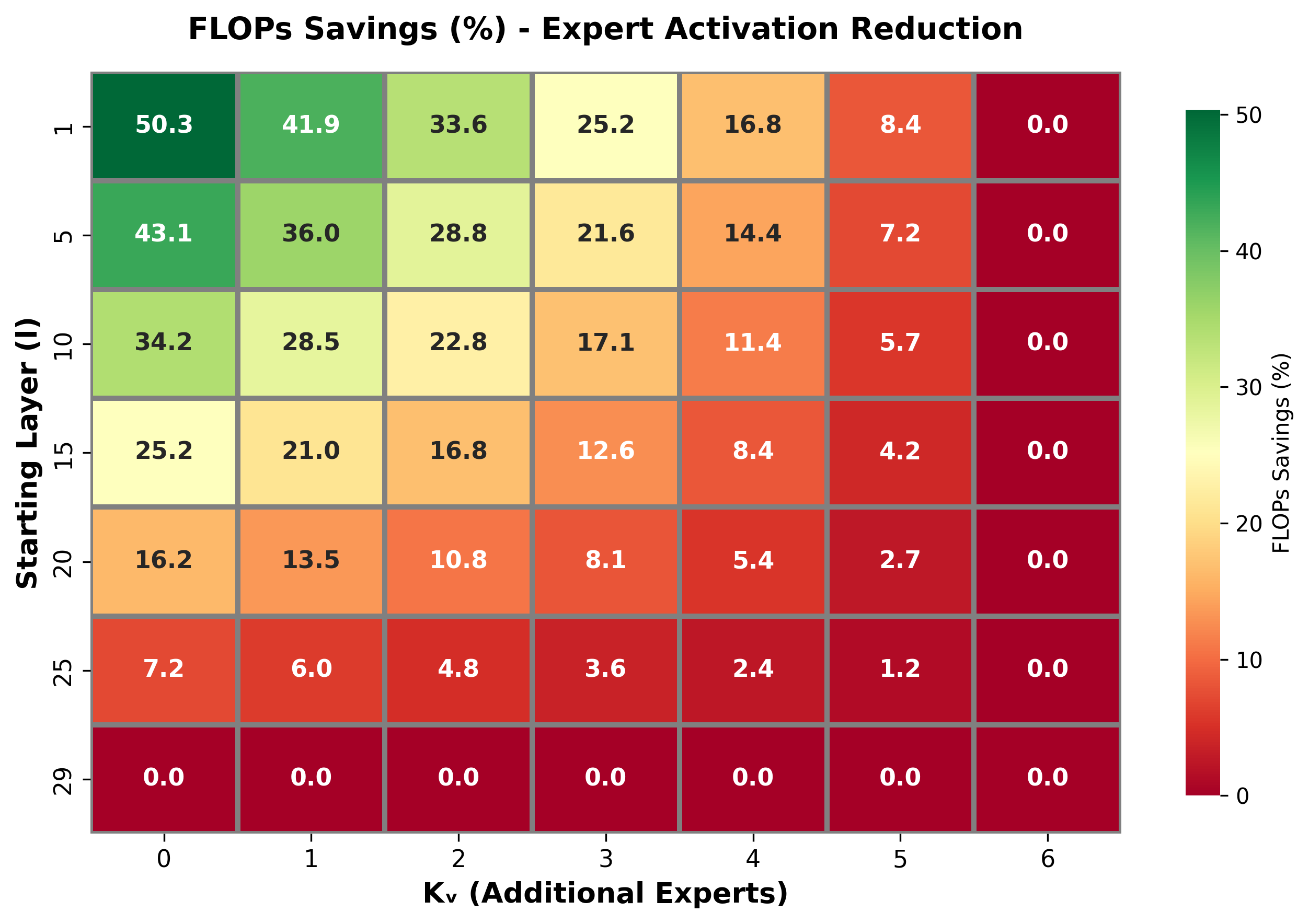}
  \caption{\textbf{FLOPs savings for DeepSeek-VL2 with expert activation reduction only.} 
  Heatmap shows FLOPs savings (\%) for different combinations of starting layer $l$ 
  and number of additional activated experts $K_v$ (beyond 2 shared experts). 
  Earlier activation reduction (smaller $l$) and fewer activated experts (smaller $K_v$) 
  yield higher savings. 
  Best case ($l=1, K_v=0$): 43.58\% FLOPs reduction; 
  worst case ($l=29, K_v=6$): 0\% reduction.}
  \label{fig:deepseek_flops_formula2}
\end{figure}

\begin{figure}[H]
\centering
  \includegraphics[width=\columnwidth,keepaspectratio]{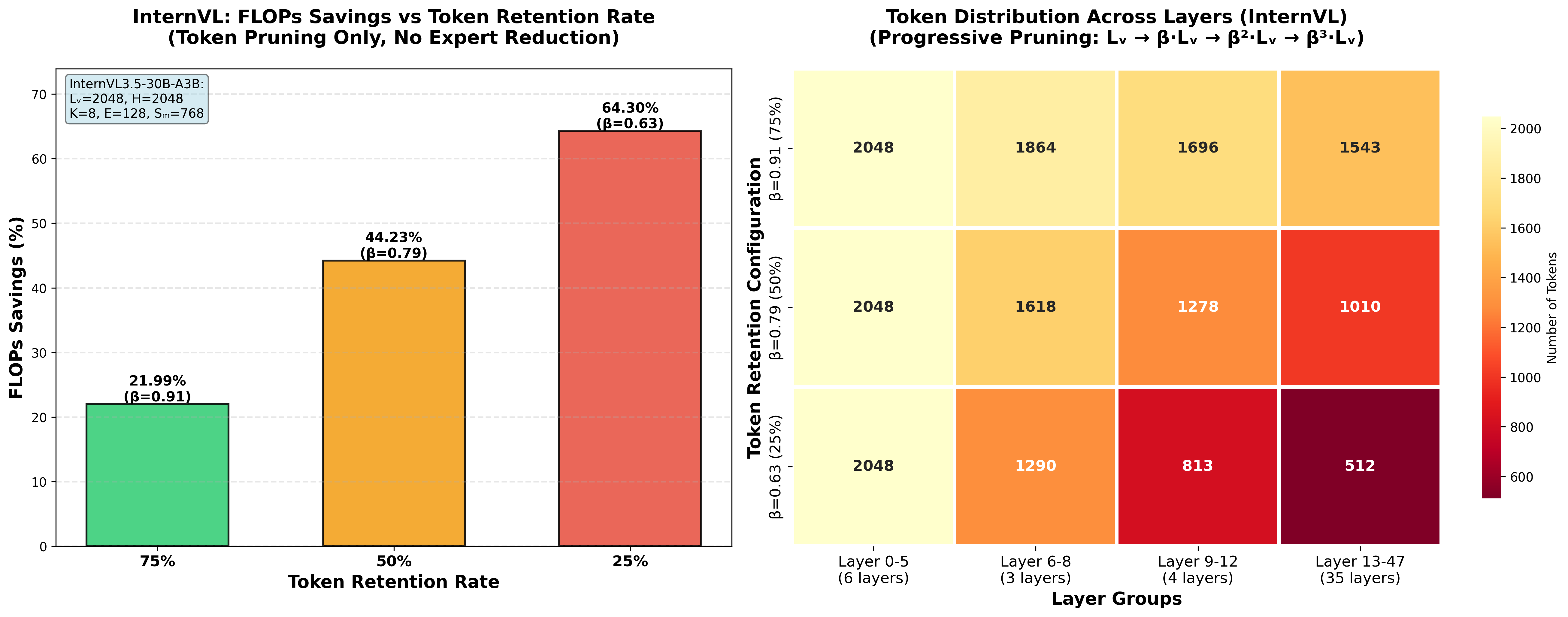}
  \caption{\textbf{FLOPs savings for InternVL3.5-30B-A3B with token pruning only.} 
  (\emph{Left}) FLOPs savings at three retention rates: 
  $\beta=0.91$ (75\%), $\beta=0.79$ (50\%), and $\beta=0.63$ (25\%). 
  (\emph{Right}) Token distribution across four layer groups (0-5, 6-8, 9-12, 13-47) 
  under progressive pruning. 
  InternVL's deeper architecture (48 layers) benefits from aggressive pruning, 
  achieving up to 64.30\% FLOPs reduction at 25\% retention.}
  \label{fig:internvl_flops_formula1}
\end{figure}

\begin{figure}[H]
\centering
  \includegraphics[width=\columnwidth,keepaspectratio]{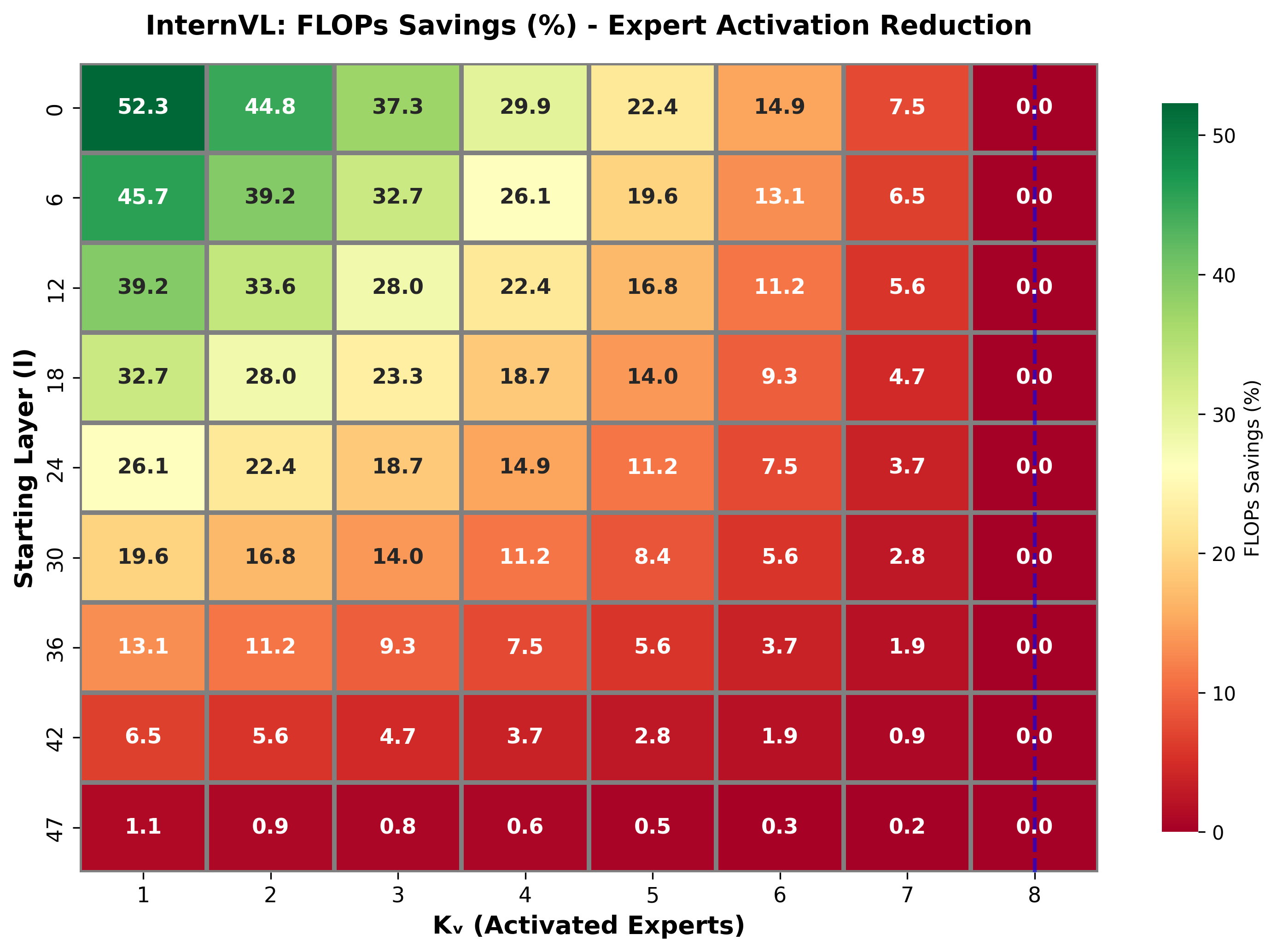}
  \caption{\textbf{FLOPs savings for InternVL3.5-30B-A3B with expert activation reduction only.} 
  Heatmap shows FLOPs savings (\%) for different starting layers $l$ and 
  number of activated experts $K_v \in [1,8]$. 
  Unlike DeepSeek-VL2, InternVL has no shared experts, 
  allowing all experts to be freely reduced. 
  Best case ($l=0, K_v=1$): 63.91\% FLOPs reduction; 
  the longer layer span (48 vs.\ 30) amplifies the impact of early-layer activation reduction.}
  \label{fig:internvl_flops_formula2}
\end{figure}

\subsection{Empirical FLOPs Analysis and Visualization}
\label{sec:flops_visualization}

To validate the theoretical derivations and provide intuitive insights into the computational savings, 
we visualize the FLOPs reduction under different configurations for both DeepSeek-VL2 and InternVL3.5-30B-A3B. 
\cref{fig:deepseek_flops_formula1,fig:deepseek_flops_formula2,fig:internvl_flops_formula1,fig:internvl_flops_formula2} 
present the results for token pruning only and expert activation reduction only, respectively. 
These visualizations reveal several key observations about the effectiveness of each strategy 
and how they interact with different model architectures.

\paragraph{Token Pruning Strategy.}
\cref{fig:deepseek_flops_formula1,fig:internvl_flops_formula1} illustrate the FLOPs savings 
achieved through progressive vision-token pruning at three retention rates: 
$\beta=0.91$ (75\%), $\beta=0.79$ (50\%), and $\beta=0.63$ (25\%). 
For DeepSeek-VL2, the savings range from 21.99\% at 75\% retention to 64.30\% at 25\% retention. 
InternVL3.5-30B-A3B exhibits a similar trend but with slightly different magnitudes due to its distinct layer structure 
(48 layers with pruning at layers 5, 8, 12 versus DeepSeek's 30 layers with pruning at layers 2, 5, 8). 
The right panels of both figures show how tokens are progressively reduced across layer groups 
($L_v \to \beta L_v \to \beta^2 L_v \to \beta^3 L_v$), 
with the majority of layers (29 out of 30 for DeepSeek; 35 out of 48 for InternVL) 
operating on heavily pruned sequences ($\beta^3 L_v$). 
This cubic reduction in the deepest layers is the primary driver of computational savings, 
as the quadratic attention cost scales with the square of sequence length.

\paragraph{Expert Activation Reduction Strategy.}
\cref{fig:deepseek_flops_formula2,fig:internvl_flops_formula2} present heatmaps of FLOPs savings 
as a function of the starting layer $l$ and the number of activated experts $K_v$. 
A critical architectural difference emerges here: 
DeepSeek-VL2 has 2 shared experts that are always activated, 
so $K_v$ represents \emph{additional} experts beyond the shared ones (ranging from 0 to 6 in the visualization); 
InternVL3.5-30B-A3B has \emph{no} shared experts, allowing $K_v$ to vary freely from 1 to 8. 
Both models exhibit a clear gradient: 
earlier activation reduction (smaller $l$) and fewer activated experts (smaller $K_v$) yield higher savings. 
For DeepSeek, applying activation reduction from layer 1 with $K_v=0$ (only shared experts) 
achieves 43.58\% FLOPs reduction, 
while delaying to layer 29 results in negligible savings. 
InternVL shows a similar pattern, with best-case savings of 63.91\% at $l=0, K_v=1$. 
The steeper gradient in InternVL's heatmap reflects its longer layer span (48 vs.\ 30), 
which amplifies the impact of early-layer intervention.

\paragraph{Combined Strategy and Trade-offs.}
While the visualizations focus on individual strategies, 
our experimental results in \cref{tab:internvl_main_table} demonstrate that 
\emph{combining} token pruning and expert activation reduction can yield complementary benefits. 
For instance, FastMMoE$^\dagger$ with $l_v=6, K_v=5$ on InternVL achieves 55.02\% FLOPs savings 
at 50\% token retention while maintaining 75.63\% average accuracy (only 3.53 points drop). 
The theoretical formulas in \cref{sec:flops} account for this synergy: 
token pruning reduces the sequence length $L_v$ in the attention and MoE computations, 
while expert activation reduction decreases the effective $K$ in the MoE term ($6BLHS_mK$). 
The heatmaps reveal that \emph{modest} activation reduction (e.g., $K_v=4$ or $5$ out of 8) 
from intermediate layers (e.g., $l_v=6$ or $10$) provides a favorable accuracy-efficiency trade-off, 
as overly aggressive reduction (e.g., $K_v=1$ from $l=0$) can degrade routing quality for vision tokens. 
This underscores the importance of tuning $l_v$ and $K_v$ based on the specific model architecture 
and downstream task requirements.

\paragraph{Summary.}
The visualizations confirm that both token pruning and expert activation reduction 
are effective mechanisms for reducing FLOPs in MoE-based MLLMs, 
with savings ranging from 20\% to over 60\% depending on the aggressiveness of the configuration. 
Token pruning is particularly impactful due to the quadratic scaling of attention, 
while expert activation reduction offers fine-grained control over MoE computation. 
The architectural differences between DeepSeek-VL2 and InternVL3.5-30B-A3B 
(e.g., shared experts, layer count, pruning schedules) 
lead to quantitatively different savings profiles, 
but the qualitative trends remain consistent across both models.

\begin{table*}[!t]
  \centering
  \small
  \caption{\textbf{Real-world Prefill Latency on MMMU (Batch Size=8).} Evaluated on a single NVIDIA A800 GPU. FastMMoE (75\% retain, 4 experts) reduces latency from 2796.5ms to 1800.3ms (\textbf{1.55$\times$ speedup}), outperforming pruning-only baselines.}
  \label{tab:latency_appendix}
  \begin{tabular}{lcccc}
    \toprule
    \textbf{Method} & \textbf{Retain} & \textbf{$K_v$ (Experts)} & \textbf{Latency (ms)} & \textbf{Speedup} \\
    \midrule
    Baseline & 100\% & 8 (Full) & 2796.5 & 1.00$\times$ \\
    Pruning Only & 75\% & 8 & 1827.2 & 1.53$\times$ \\
    \textbf{FastMMoE} & \textbf{75\%} & \textbf{4} & \textbf{1800.3} & \textbf{1.55$\times$} \\
    Pruning Only & 50\% & 8 & 1817.4 & 1.54$\times$ \\
    Pruning Only & 25\% & 8 & 1797.4 & 1.56$\times$ \\
    \bottomrule
  \end{tabular}
\end{table*}

\subsection{Real-world Latency Analysis}
\label{sec:latency_analysis}

While theoretical FLOPs reduction provides a hardware-agnostic measure of efficiency, it does not always perfectly translate to wall-clock latency improvements due to memory bandwidth constraints and kernel overheads. To evaluate the practical acceleration of \textbf{FastMMoE}, we conducted real-world prefill latency measurements on a single \textbf{NVIDIA A800 GPU} for InternVL3.5. 

Table~\ref{tab:latency_appendix} reports the prefill latency on the MMMU benchmark with a batch size of 8. We observe the following key findings:

\begin{itemize}
    \item \textbf{Significant Speedup:} Our complete FastMMoE framework (75\% token retention, 4 active experts) significantly reduces the prefill latency from 2796.5~ms to 1800.3~ms, achieving a \textbf{1.55$\times$ speedup} over the full-expert baseline. This demonstrates the practical effectiveness of our method in real-world deployment scenarios.
    \item \textbf{Diminishing Returns at Extreme Compression:} When applying token pruning only, we observe that decreasing the retention ratio below 75\% (e.g., to 50\% or 25\%) hits a latency plateau (around 1797~ms to 1827~ms). This indicates that at highly compressed sequence lengths, the inference becomes memory-bound, and fixed kernel overheads dominate the execution time. Consequently, the 75\% retention configuration offers the optimal trade-off between latency reduction and performance preservation.
    \item \textbf{Potential of Expert Reduction:} Comparing FastMMoE (75\% retain, 4 experts) with the Pruning-Only baseline (75\% retain, 8 experts), we observe an additional latency reduction (from 1827.2~ms to 1800.3~ms). While the current lack of highly optimized, fused MoE kernels masks some of the latency gains from expert reduction, the FLOPs savings remain substantial. Future developments in fused MoE kernels are expected to further unlock the wall-clock acceleration potential of activating fewer experts.
\end{itemize}

\section{Acknowledgments}
The work is supported by Beijing Natural Science Foundation (L247010), National Key R\&D Program of China (No. 2024YFF0907003, No.2020YFF0305302), and the Science and Technology Project of State Grid Corporation of China (No. 5700-202058480A-0-0-00).

% \usepackage{booktabs}
% \usepackage{longtable}
% Note: It may be necessary to compile the document several times to get a multi-page table to line up properly
\onecolumn
\begin{longtable}[c]{@{}cc|cccccc|c@{}}
\caption{Comparison on different strategy of reducing activation in InternVL3.5.}
\label{tab:internvl_reduce_act_all}\\
\toprule
\multicolumn{1}{l}{\textbf{$K_v$}} &
  \multicolumn{1}{l|}{\textbf{$l_v$}} &
  \textbf{MMMU} &
  \textbf{SQA$^\text{I}$} &
  \textbf{MMBench} &
  \textbf{OCRBench} &
  \textbf{HallusionBench} &
  \textbf{AI2D} &
  \textbf{Avg.$\uparrow$} \\* \midrule
\endfirsthead
\endhead
\bottomrule
\endfoot
\endlastfoot
1 & 2  & 54.22          & 85.37 & 76.63 & 27.70 & 39.91 & 73.12 & 59.49 \\
1 & 5  & 55.22          & 86.32 & 76.55 & 29.70 & 41.38 & 74.06 & 60.54 \\
1 & 10 & 54.11          & 87.70 & 78.52 & 32.10 & 40.50 & 75.65 & 61.43 \\
1 & 15 & 57.00          & 89.89 & 78.78 & 36.90 & 42.09 & 78.01 & 63.78 \\
1 & 20 & 57.56          & 93.60 & 81.27 & 46.30 & 44.96 & 81.61 & 67.55 \\
1 & 25 & 59.78          & 98.56 & 84.88 & 64.50 & 50.20 & 85.36 & 73.88 \\
1 & 30 & 60.33 & 98.91 & 85.91 & 75.90 & 52.28 & 86.66 & 76.67 \\
1 & 35 & 61.67          & 99.01 & 86.34 & 82.80 & 53.41 & 86.95 & 78.36 \\
1 & 40 & 60.89          & 98.96 & 86.43 & 87.00 & 53.43 & 87.05 & 78.96 \\
1 & 45 & 60.89          & 98.96 & 86.43 & 88.00 & 53.46 & 87.18 & 79.15 \\
2 & 2  & 58.78          & 96.38 & 84.71 & 79.30 & 47.56 & 84.36 & 75.18 \\
2 & 5  & 58.33          & 96.88 & 84.28 & 81.30 & 47.02 & 84.33 & 75.36 \\
2 & 10 & 58.33          & 97.37 & 84.88 & 81.20 & 47.21 & 84.91 & 75.65 \\
2 & 15 & 59.33 & 97.87 & 84.71 & 81.40 & 48.57 & 84.91 & 76.13 \\
2 & 20 & 60.11          & 98.31 & 85.22 & 83.70 & 50.69 & 85.75 & 77.30 \\
2 & 25 & 61.11          & 98.71 & 85.91 & 83.90 & 53.65 & 86.24 & 78.25 \\
2 & 30 & 60.44          & 99.01 & 86.08 & 85.80 & 53.48 & 86.88 & 78.62 \\
2 & 35 & 61.00          & 98.96 & 86.43 & 86.60 & 53.66 & 86.92 & 78.93 \\
2 & 40 & 60.56          & 98.96 & 86.34 & 87.60 & 53.56 & 87.11 & 79.02 \\
2 & 45 & 61.00          & 98.96 & 86.34 & 88.20 & 53.25 & 87.08 & 79.14 \\
3 & 2  & 60.44          & 98.86 & 85.82 & 85.70 & 50.02 & 86.04 & 77.82 \\
3 & 5  & 60.56          & 98.66 & 85.65 & 87.00 & 50.84 & 85.82 & 78.09 \\
3 & 10 & 59.78          & 98.86 & 86.43 & 85.90 & 49.67 & 86.24 & 77.81 \\
3 & 15 & 60.67          & 98.91 & 85.48 & 84.80 & 51.27 & 86.27 & 77.90 \\
3 & 20 & 60.22          & 98.81 & 86.25 & 86.10 & 53.02 & 86.66 & 78.51 \\
3 & 25 & 60.11          & 98.86 & 86.43 & 86.80 & 52.24 & 86.66 & 78.52 \\
3 & 30 & 60.78          & 99.01 & 86.25 & 86.40 & 54.01 & 86.76 & 78.87 \\
3 & 35 & 61.11          & 99.01 & 86.25 & 87.40 & 53.86 & 86.98 & 79.10 \\
3 & 40 & 60.56          & 98.96 & 86.43 & 87.80 & 53.56 & 87.24 & 79.09 \\
3 & 45 & 61.00          & 99.01 & 86.43 & 88.20 & 53.14 & 87.11 & 79.15 \\
4 & 2  & 60.11          & 98.91 & 86.34 & 87.20 & 52.41 & 86.46 & 78.57 \\
4 & 5  & 60.11          & 98.96 & 86.51 & 86.90 & 52.35 & 86.24 & 78.51 \\
4 & 10 & 59.78          & 99.01 & 86.34 & 86.90 & 52.34 & 86.82 & 78.53 \\
4 & 15 & 60.78          & 98.91 & 86.25 & 87.60 & 53.02 & 86.46 & 78.84 \\
4 & 20 & 60.89          & 98.91 & 86.25 & 87.50 & 52.95 & 86.85 & 78.89 \\
4 & 25 & 60.67          & 98.91 & 86.25 & 87.40 & 53.26 & 86.82 & 78.88 \\
4 & 30 & 60.33          & 99.06 & 86.60 & 87.20 & 53.69 & 87.21 & 79.01 \\
4 & 35 & 60.44          & 98.96 & 86.34 & 87.40 & 53.66 & 87.08 & 78.98 \\
4 & 40 & 61.00          & 98.96 & 86.43 & 88.00 & 53.77 & 87.11 & 79.21 \\
4 & 45 & 61.11          & 98.96 & 86.43 & 88.40 & 53.14 & 87.18 & 79.20 \\
5 & 2  & 61.11          & 98.86 & 86.43 & 87.50 & 53.06 & 86.72 & 78.95 \\
5 & 5  & 60.56          & 99.06 & 86.60 & 88.20 & 53.07 & 86.82 & 79.05 \\
5 & 10 & 61.00          & 98.91 & 86.60 & 87.80 & 53.57 & 86.88 & 79.13 \\
5 & 15 & 61.78          & 98.96 & 86.17 & 87.60 & 53.77 & 86.79 & 79.18 \\
5 & 20 & 61.33          & 98.86 & 86.60 & 87.50 & 53.54 & 87.21 & 79.17 \\
5 & 25 & 60.67          & 98.86 & 86.34 & 87.70 & 53.84 & 86.88 & 79.05 \\
5 & 30 & 60.33          & 99.01 & 86.34 & 87.50 & 53.59 & 87.21 & 79.00 \\
5 & 35 & 60.78          & 98.96 & 86.34 & 87.90 & 53.66 & 87.11 & 79.12 \\
5 & 40 & 60.89          & 98.96 & 86.51 & 88.20 & 53.25 & 87.11 & 79.15 \\
5 & 45 & 61.00          & 98.96 & 86.43 & 88.50 & 53.03 & 87.18 & 79.18 \\
6 & 2  & 61.33          & 99.01 & 86.43 & 88.00 & 53.32 & 86.95 & 79.17 \\
6 & 5  & 61.44          & 98.91 & 86.25 & 87.90 & 53.66 & 86.95 & 79.19 \\
6 & 10 & 61.56          & 98.91 & 86.43 & 88.30 & 53.78 & 86.88 & 79.31 \\
6 & 15 & 61.22          & 98.86 & 86.00 & 88.00 & 53.17 & 86.95 & 79.03 \\
6 & 20 & 61.44          & 98.96 & 86.43 & 87.70 & 52.90 & 86.88 & 79.05 \\
6 & 25 & 60.89          & 98.96 & 86.60 & 88.00 & 53.08 & 86.95 & 79.08 \\
6 & 30 & 61.00          & 98.96 & 86.43 & 88.20 & 53.46 & 87.01 & 79.18 \\
6 & 35 & 61.22          & 98.96 & 86.43 & 88.60 & 53.56 & 87.18 & 79.32 \\
6 & 40 & 60.78          & 98.96 & 86.51 & 88.40 & 53.36 & 87.18 & 79.20 \\
6 & 45 & 60.89          & 99.01 & 86.34 & 88.50 & 53.03 & 87.08 & 79.14 \\
7 & 2  & 61.67          & 98.91 & 86.51 & 87.90 & 53.70 & 87.05 & 79.29 \\
7 & 5  & 61.33          & 98.91 & 86.08 & 88.10 & 53.35 & 86.85 & 79.10 \\
7 & 10 & 61.33          & 98.91 & 86.17 & 88.00 & 53.53 & 86.95 & 79.15 \\
7 & 15 & 61.67          & 98.86 & 86.25 & 88.20 & 53.66 & 86.85 & 79.25 \\
7 & 20 & 61.33          & 98.96 & 86.25 & 88.10 & 53.60 & 86.76 & 79.17 \\
7 & 25 & 60.56          & 98.96 & 86.25 & 87.90 & 53.12 & 87.24 & 79.00 \\
7 & 30 & 60.67          & 99.01 & 86.51 & 88.00 & 53.27 & 87.11 & 79.10 \\
7 & 35 & 60.78          & 99.01 & 86.43 & 87.90 & 53.56 & 87.18 & 79.14 \\
7 & 40 & 61.11          & 98.96 & 86.51 & 88.50 & 53.25 & 87.18 & 79.25 \\
7 & 45 & 61.00          & 98.96 & 86.43 & 88.50 & 53.14 & 87.05 & 79.18 \\* \midrule
8 & 48 & 60.67          & 98.96 & 86.43 & 88.60 & 53.14 & 87.14 & 79.16 \\* \bottomrule
\end{longtable}

% \usepackage{longtable}
% Note: It may be necessary to compile the document several times to get a multi-page table to line up properly
\onecolumn
% \usepackage{booktabs}
% \usepackage{longtable}
% Note: It may be necessary to compile the document several times to get a multi-page table to line up properly
\begin{longtable}[c]{@{}cc|cccccc|c@{}}
\caption{Comparison on different strategy of reducing activation in DeepSeek-VL2.}
\label{tab:deepseek_reduce_act_all}\\
\toprule
\multicolumn{1}{l}{\textbf{$K_v$}} &
  \multicolumn{1}{l|}{\textbf{$l_v$}} &
  \textbf{MMMU} &
  \textbf{SQA$^\text{I}$} &
  \textbf{MMBench} &
  \textbf{OCRBench} &
  \textbf{HallusionBench} &
  \textbf{AI2D} &
  \textbf{Avg.$\uparrow$} \\* \midrule
\endfirsthead
\endhead
\bottomrule
\endfoot
\endlastfoot
0 & 2  & 47.67          & 90.18 & 79.04 & 62.20 & 32.54 & 78.59 & 65.04 \\
0 & 5  & 50.11          & 91.62 & 80.67 & 71.30 & 34.10 & 80.83 & 68.11 \\
0 & 10 & 51.56          & 96.78 & 82.56 & 80.00 & 41.19 & 82.32 & 72.40 \\
0 & 15 & 51.11          & 96.83 & 83.16 & 81.20 & 41.49 & 82.32 & 72.68 \\
0 & 20 & 51.89          & 96.88 & 83.08 & 81.00 & 41.47 & 82.06 & 72.73 \\
0 & 25 & 51.78          & 96.83 & 83.16 & 81.00 & 41.12 & 82.12 & 72.67 \\
1 & 2  & 49.89 & 93.55 & 81.53 & 70.80 & 35.87 & 80.15 & 68.63 \\
1 & 5  & 50.33          & 94.10 & 81.62 & 75.20 & 37.32 & 80.86 & 69.91 \\
1 & 10 & 51.67          & 96.53 & 82.73 & 78.80 & 39.76 & 81.54 & 71.84 \\
1 & 15 & 51.89          & 96.73 & 83.16 & 79.20 & 41.98 & 82.29 & 72.54 \\
1 & 20 & 51.89          & 96.78 & 83.25 & 81.00 & 41.57 & 82.32 & 72.80 \\
1 & 25 & 51.78          & 96.83 & 83.25 & 80.30 & 41.05 & 82.12 & 72.55 \\
2 & 2  & 51.22          & 96.38 & 82.73 & 78.10 & 38.99 & 81.90 & 71.55 \\
2 & 5  & 51.00 & 96.48 & 83.16 & 79.50 & 39.51 & 81.74 & 71.90 \\
2 & 10 & 51.67          & 96.63 & 82.90 & 80.60 & 40.40 & 81.93 & 72.35 \\
2 & 15 & 52.00          & 96.83 & 83.33 & 80.10 & 41.27 & 82.35 & 72.65 \\
2 & 20 & 51.56          & 96.83 & 83.33 & 80.50 & 41.05 & 82.35 & 72.60 \\
2 & 25 & 51.67          & 96.88 & 83.16 & 81.00 & 40.92 & 82.35 & 72.66 \\
3 & 2  & 52.00          & 96.58 & 83.33 & 79.60 & 40.52 & 82.35 & 72.40 \\
3 & 5  & 51.67          & 96.93 & 82.99 & 80.30 & 40.92 & 82.03 & 72.47 \\
3 & 10 & 52.00          & 97.03 & 83.08 & 80.80 & 40.35 & 81.87 & 72.52 \\
3 & 15 & 52.11          & 96.88 & 83.16 & 81.10 & 41.44 & 82.29 & 72.83 \\
3 & 20 & 51.56          & 96.83 & 83.08 & 80.90 & 41.42 & 82.32 & 72.68 \\
3 & 25 & 51.44          & 96.88 & 83.25 & 81.30 & 40.81 & 82.35 & 72.67 \\
4 & 2  & 51.89          & 96.93 & 83.51 & 80.20 & 41.09 & 81.90 & 72.58 \\
4 & 5  & 52.44          & 96.73 & 83.33 & 80.60 & 41.70 & 82.32 & 72.85 \\
4 & 10 & 51.78          & 96.88 & 83.16 & 80.80 & 41.08 & 82.09 & 72.63 \\
4 & 15 & 51.89          & 96.88 & 83.33 & 81.40 & 40.94 & 82.29 & 72.79 \\
4 & 20 & 51.56          & 96.93 & 83.25 & 81.00 & 41.24 & 82.25 & 72.70 \\
4 & 25 & 51.89          & 96.88 & 83.25 & 81.60 & 40.81 & 82.22 & 72.77 \\
5 & 2  & 52.33          & 96.88 & 83.08 & 80.90 & 41.60 & 81.70 & 72.75 \\
5 & 5  & 51.78          & 96.93 & 83.08 & 81.00 & 40.93 & 82.35 & 72.68 \\
5 & 10 & 52.56          & 96.88 & 83.16 & 80.90 & 41.06 & 82.12 & 72.78 \\
5 & 15 & 52.11          & 96.88 & 83.08 & 81.50 & 41.14 & 82.06 & 72.79 \\
5 & 20 & 51.56          & 96.88 & 83.25 & 81.40 & 40.90 & 82.22 & 72.70 \\
5 & 25 & 51.67          & 96.83 & 83.25 & 81.60 & 40.97 & 82.29 & 72.77 \\* \midrule
6 & 29 & 51.89          & 96.88 & 83.25 & 81.40 & 40.83 & 82.38 & 72.77 \\* \bottomrule
\end{longtable}

% \usepackage{booktabs}
% \usepackage{multirow}
% \usepackage{longtable}
% Note: It may be necessary to compile the document several times to get a multi-page table to line up properly
\begin{longtable}[c]{@{}c|cc|cccccc|c@{}}
\caption{\textbf{Comprehensive ablation study of routing-similarity weighting parameter $\alpha$ and merge rate $\gamma$ on InternVL3.5.} 
We report results across six benchmarks under three different vision-token retain ratios (75\%, 50\%, and 25\%). 
$\alpha$ controls the balance between routing-probability similarity and cross-modal attention in redundancy scoring, while $\gamma$ controls the proportion of retained tokens produced through merging high-redundancy windows. 
The table enumerates all $(\alpha,\gamma)$ combinations tested, showing that optimal values of $\alpha$ and $\gamma$ vary with the pruning intensity: moderate $\alpha$ values and merge rates close to their theoretical upper bound yield the best trade-off between accuracy and compression.}
\label{tab:internvl_pruning_ablation}\\
\toprule
\textbf{Retain Ratio} &
  \multicolumn{1}{l}{\textbf{$\alpha$}} &
  \multicolumn{1}{l|}{\textbf{$\gamma$}} &
  \textbf{MMMU} &
  \textbf{SQA$^\text{I}$} &
  \textbf{MMBench} &
  \textbf{OCRBench} &
  \textbf{HallusionBench} &
  \textbf{AI2D} &
  \textbf{Avg.$\uparrow$} \\* \midrule
\endfirsthead
\endhead
\bottomrule
\endfoot
\endlastfoot
\multirow{11}{*}{75\%} & 1   & 0.025 & 60.44          & 98.96 & 86.25 & 80.00 & 52.21 & 86.33 & 77.37 \\
                       & 0.9 & 0.025 & 60.78          & 99.01 & 86.25 & 79.30 & 51.40 & 86.43 & 77.20 \\
                       & 0.8 & 0.025 & 60.89          & 98.91 & 86.51 & 79.90 & 51.45 & 86.30 & 77.33 \\
                       & 0.7 & 0.025 & 60.78          & 98.91 & 86.60 & 80.40 & 51.26 & 86.08 & 77.34 \\
                       & 0.6 & 0.025 & 60.33          & 98.86 & 86.17 & 79.70 & 52.29 & 86.20 & 77.26 \\
                       & 0.5 & 0.025 & 60.67          & 98.91 & 86.60 & 80.70 & 52.21 & 86.27 & 77.56 \\
                       & 0.4 & 0.025 & \textbf{60.78} & 99.01 & 86.25 & 80.50 & 52.46 & 86.20 & 77.53 \\
                       & 0.3 & 0.025 & 60.78          & 98.91 & 86.25 & 80.50 & 51.73 & 86.08 & 77.37 \\
                       & 0.2 & 0.025 & 60.78          & 99.01 & 86.00 & 80.70 & 51.95 & 86.20 & 77.44 \\
                       & 0.1 & 0.025 & 60.78          & 98.86 & 86.00 & 79.50 & 52.26 & 86.08 & 77.24 \\
                       & 0   & 0.025 & 60.78          & 98.96 & 86.17 & 79.70 & 51.41 & 86.08 & 77.18 \\* \midrule
\multirow{22}{*}{50\%} & 1   & 0.05  & 59.89          & 98.56 & 85.57 & 76.80 & 50.92 & 84.91 & 76.11 \\
                       & 1   & 0.025 & 60.56          & 97.87 & 85.91 & 75.70 & 50.62 & 85.07 & 75.95 \\
                       & 0.9 & 0.05  & \textbf{60.56} & 98.56 & 85.22 & 75.70 & 52.58 & 84.78 & 76.23 \\
                       & 0.9 & 0.025 & 60.44          & 98.22 & 85.91 & 74.30 & 50.56 & 84.13 & 75.59 \\
                       & 0.8 & 0.05  & 60.89          & 98.41 & 85.31 & 75.70 & 51.23 & 84.94 & 76.08 \\
                       & 0.8 & 0.025 & 60.44          & 98.17 & 85.14 & 75.40 & 49.98 & 84.07 & 75.53 \\
                       & 0.7 & 0.05  & 60.00          & 98.26 & 85.31 & 75.30 & 53.25 & 84.49 & 76.10 \\
                       & 0.7 & 0.025 & 59.89          & 98.17 & 85.31 & 74.30 & 49.73 & 84.29 & 75.28 \\
                       & 0.6 & 0.05  & 61.00          & 98.36 & 85.22 & 76.50 & 52.54 & 84.49 & 76.35 \\
                       & 0.6 & 0.025 & 60.44          & 98.22 & 85.05 & 74.10 & 48.57 & 84.26 & 75.11 \\
                       & 0.5 & 0.05  & 59.56          & 98.07 & 85.65 & 76.00 & 50.00 & 84.59 & 75.64 \\
                       & 0.5 & 0.025 & 60.56          & 98.12 & 84.97 & 72.70 & 48.04 & 83.48 & 74.64 \\
                       & 0.4 & 0.05  & 59.89          & 98.22 & 85.40 & 76.00 & 50.52 & 84.72 & 75.79 \\
                       & 0.4 & 0.025 & 59.56          & 98.26 & 84.97 & 74.10 & 48.61 & 83.48 & 74.83 \\
                       & 0.3 & 0.05  & 60.22          & 98.36 & 85.40 & 75.70 & 49.66 & 84.81 & 75.69 \\
                       & 0.3 & 0.025 & 60.89          & 98.12 & 84.88 & 74.10 & 47.03 & 83.42 & 74.74 \\
                       & 0.2 & 0.05  & 60.89          & 98.07 & 85.74 & 74.80 & 49.72 & 84.13 & 75.56 \\
                       & 0.2 & 0.025 & 59.89          & 98.02 & 84.79 & 73.70 & 48.66 & 83.81 & 74.81 \\
                       & 0.1 & 0.05  & 60.22          & 97.97 & 85.22 & 75.80 & 50.56 & 84.07 & 75.64 \\
                       & 0.1 & 0.025 & 60.89          & 98.02 & 85.14 & 74.10 & 48.61 & 83.39 & 75.02 \\
                       & 0   & 0.05  & 60.22          & 98.22 & 85.48 & 75.30 & 49.72 & 83.65 & 75.43 \\
                       & 0   & 0.025 & 60.00          & 97.87 & 85.31 & 73.50 & 49.00 & 83.03 & 74.78 \\* \midrule
\multirow{44}{*}{25\%} & 1   & 0.15  & 59.33          & 96.18 & 83.76 & 69.20 & 46.77 & 80.99 & 72.71 \\
                       & 1   & 0.1   & 60.78          & 96.98 & 83.93 & 66.80 & 50.52 & 82.12 & 73.52 \\
                       & 1   & 0.05  & 59.00          & 97.03 & 82.73 & 62.30 & 44.84 & 80.44 & 71.06 \\
                       & 1   & 0.025 & 59.44          & 96.18 & 82.56 & 60.00 & 45.59 & 78.72 & 70.42 \\
                       & 0.9 & 0.15  & 59.22          & 96.93 & 84.28 & 66.90 & 46.18 & 81.57 & 72.51 \\
                       & 0.9 & 0.1   & 60.00          & 96.63 & 83.33 & 64.20 & 50.33 & 81.02 & 72.59 \\
                       & 0.9 & 0.05  & 58.67          & 96.58 & 82.30 & 61.50 & 47.48 & 78.82 & 70.89 \\
                       & 0.9 & 0.025 & 59.44          & 96.13 & 81.79 & 56.80 & 44.05 & 78.11 & 69.39 \\
                       & 0.8 & 0.15  & 59.67          & 96.73 & 83.85 & 66.00 & 47.15 & 81.06 & 72.41 \\
                       & 0.8 & 0.1   & 59.44          & 96.28 & 82.99 & 63.80 & 49.91 & 81.09 & 72.25 \\
                       & 0.8 & 0.05  & 59.56          & 96.13 & 82.90 & 58.60 & 46.06 & 78.37 & 70.27 \\
                       & 0.8 & 0.025 & 57.89          & 95.54 & 82.73 & 55.80 & 46.02 & 78.04 & 69.34 \\
                       & 0.7 & 0.15  & 60.00          & 96.68 & 83.59 & 65.60 & 47.72 & 80.99 & 72.43 \\
                       & 0.7 & 0.1   & 60.33          & 96.63 & 83.51 & 63.70 & 47.44 & 80.31 & 71.99 \\
                       & 0.7 & 0.05  & 58.44          & 96.13 & 83.08 & 59.40 & 45.24 & 78.21 & 70.08 \\
                       & 0.7 & 0.025 & 58.78          & 95.64 & 82.90 & 56.30 & 44.85 & 77.95 & 69.40 \\
                       & 0.6 & 0.15  & 60.00          & 96.73 & 83.33 & 64.90 & 47.20 & 80.96 & 72.19 \\
                       & 0.6 & 0.1   & 60.00          & 95.98 & 83.68 & 62.50 & 46.97 & 80.28 & 71.57 \\
                       & 0.6 & 0.05  & 58.44          & 95.98 & 82.73 & 58.20 & 44.74 & 77.91 & 69.67 \\
                       & 0.6 & 0.025 & 58.56          & 95.88 & 82.13 & 55.30 & 45.63 & 78.01 & 69.25 \\
                       & 0.5 & 0.15  & 58.33          & 96.23 & 82.82 & 64.50 & 46.10 & 81.41 & 71.57 \\
                       & 0.5 & 0.1   & 58.78          & 95.74 & 83.51 & 61.30 & 47.27 & 79.92 & 71.09 \\
                       & 0.5 & 0.05  & 58.33          & 95.49 & 82.82 & 57.80 & 45.71 & 78.53 & 69.78 \\
                       & 0.5 & 0.025 & 58.78          & 95.79 & 82.47 & 55.70 & 44.73 & 77.30 & 69.13 \\
                       & 0.4 & 0.15  & 59.67          & 96.28 & 83.33 & 65.10 & 46.80 & 80.28 & 71.91 \\
                       & 0.4 & 0.1   & 59.78          & 95.88 & 82.99 & 60.80 & 48.75 & 79.37 & 71.26 \\
                       & 0.4 & 0.05  & 58.11          & 96.03 & 83.16 & 56.90 & 45.92 & 78.40 & 69.75 \\
                       & 0.4 & 0.025 & 57.67          & 95.64 & 82.39 & 56.10 & 44.53 & 77.88 & 69.03 \\
                       & 0.3 & 0.15  & 59.89          & 96.33 & 83.16 & 64.90 & 47.39 & 79.89 & 71.93 \\
                       & 0.3 & 0.1   & 59.67          & 95.98 & 83.16 & 61.70 & 47.57 & 79.08 & 71.19 \\
                       & 0.3 & 0.05  & 58.33          & 96.13 & 82.90 & 57.00 & 45.99 & 78.17 & 69.76 \\
                       & 0.3 & 0.025 & 58.56          & 95.54 & 82.30 & 56.20 & 44.17 & 77.49 & 69.04 \\
                       & 0.2 & 0.15  & 58.56          & 96.48 & 82.13 & 63.10 & 46.85 & 80.18 & 71.22 \\
                       & 0.2 & 0.1   & 59.22          & 96.08 & 83.59 & 60.80 & 47.23 & 79.83 & 71.13 \\
                       & 0.2 & 0.05  & 58.22          & 95.98 & 83.08 & 56.90 & 45.12 & 78.01 & 69.55 \\
                       & 0.2 & 0.025 & 59.00          & 95.49 & 82.22 & 56.30 & 45.82 & 77.66 & 69.41 \\
                       & 0.1 & 0.15  & 59.33          & 96.33 & 82.82 & 63.00 & 47.37 & 80.34 & 71.53 \\
                       & 0.1 & 0.1   & 59.33          & 96.13 & 82.99 & 60.90 & 46.12 & 79.83 & 70.88 \\
                       & 0.1 & 0.05  & 57.89          & 95.79 & 82.56 & 56.00 & 44.86 & 78.47 & 69.26 \\
                       & 0.1 & 0.025 & 57.89          & 95.49 & 81.96 & 55.70 & 45.49 & 77.49 & 69.00 \\
                       & 0   & 0.15  & 59.00          & 96.23 & 82.47 & 62.90 & 46.81 & 79.73 & 71.19 \\
                       & 0   & 0.1   & 59.56          & 96.38 & 82.99 & 60.00 & 44.81 & 78.82 & 70.43 \\
                       & 0   & 0.05  & 58.00          & 96.03 & 83.08 & 57.60 & 45.36 & 78.34 & 69.73 \\
                       & 0   & 0.025 & 57.78          & 95.54 & 82.04 & 56.60 & 46.38 & 78.24 & 69.43 \\* \bottomrule
\end{longtable}

% \usepackage{booktabs}
% \usepackage{multirow}
% \usepackage{longtable}
% Note: It may be necessary to compile the document several times to get a multi-page table to line up properly
\begin{longtable}[c]{@{}c|cccccccc|c@{}}
\caption{\textbf{Comprehensive ablation study of routing-similarity weighting parameter $\alpha$ and merge rate $\gamma$ on DeepSeek-VL2.} 
Results are reported for three retain ratios (75\%, 50\%, and 25\%) across six benchmarks. 
$\alpha$ and $\gamma$ are defined as in Table~\ref{tab:internvl_pruning_ablation}. 
Compared to InternVL3.5, DeepSeek-VL2 shows smaller performance fluctuations across $(\alpha,\gamma)$ settings, partly due to its two always-activated shared experts, which stabilize performance under pruning. 
Nonetheless, the trends remain consistent: balanced $\alpha$ values and moderate $\gamma$ provide the best retention-performance trade-off.}
\label{tab:deepseek_pruning_ablation}\\
\toprule
\textbf{Retain Ratio} &
  \multicolumn{1}{l}{\textbf{$\alpha$}} &
  \multicolumn{1}{l}{\textbf{$\gamma$}} &
  \textbf{MMMU} &
  \textbf{SQA$^\text{I}$} &
  \textbf{MMBench} &
  \textbf{OCRBench} &
  \textbf{HallusionBench} &
  \textbf{AI2D} &
  \textbf{Avg.$\uparrow$} \\* \midrule
\endfirsthead
\endhead
\bottomrule
\endfoot
\endlastfoot
\multirow{8}{*}{75\%}  & 1   & \multicolumn{1}{c|}{0.05} & 51.44          & 97.03 & 82.47 & 80.00 & 41.07 & 81.99 & 72.33 \\
                       & 0.9 & \multicolumn{1}{c|}{0.05} & 51.11          & 96.93 & 82.65 & 79.50 & 40.76 & 81.90 & 72.14 \\
                       & 0.8 & \multicolumn{1}{c|}{0.05} & 51.56          & 96.78 & 82.99 & 79.90 & 40.89 & 81.99 & 72.35 \\
                       & 0.7 & \multicolumn{1}{c|}{0.05} & 51.67          & 96.83 & 82.90 & 80.40 & 41.08 & 81.74 & 72.44 \\
                       & 0.6 & \multicolumn{1}{c|}{0.05} & 51.44          & 96.83 & 82.73 & 80.60 & 40.89 & 81.67 & 72.36 \\
                       & 0.5 & \multicolumn{1}{c|}{0.05} & 52.00          & 97.03 & 82.90 & 80.70 & 40.69 & 81.90 & 72.54 \\
                       & 0.4 & \multicolumn{1}{c|}{0.05} & 51.11 & 96.98 & 83.08 & 80.10 & 41.23 & 81.77 & 72.38 \\
                       & 0.3 & \multicolumn{1}{c|}{0.05} & 51.56          & 96.98 & 83.25 & 80.80 & 40.30 & 82.06 & 72.49 \\* \midrule
\multirow{16}{*}{50\%} & 1   & \multicolumn{1}{c|}{0.1}  & 50.89          & 96.38 & 82.82 & 77.80 & 39.80 & 81.70 & 71.56 \\
                       & 1   & \multicolumn{1}{c|}{0.05} & 50.11          & 96.53 & 82.82 & 79.70 & 40.68 & 81.22 & 71.84 \\
                       & 0.9 & \multicolumn{1}{c|}{0.1}  & 50.56          & 96.38 & 83.08 & 79.40 & 40.06 & 81.25 & 71.79 \\
                       & 0.9 & \multicolumn{1}{c|}{0.05} & 51.11          & 96.33 & 82.99 & 79.30 & 40.55 & 81.35 & 71.94 \\
                       & 0.8 & \multicolumn{1}{c|}{0.1}  & 50.67          & 96.38 & 82.99 & 79.30 & 40.20 & 81.25 & 71.80 \\
                       & 0.8 & \multicolumn{1}{c|}{0.05} & 50.22 & 96.48 & 83.08 & 79.20 & 40.07 & 81.15 & 71.70 \\
                       & 0.7 & \multicolumn{1}{c|}{0.1}  & 50.11          & 96.18 & 82.65 & 78.80 & 39.32 & 81.06 & 71.35 \\
                       & 0.7 & \multicolumn{1}{c|}{0.05} & 51.00          & 96.18 & 82.73 & 79.20 & 40.23 & 81.09 & 71.74 \\
                       & 0.6 & \multicolumn{1}{c|}{0.1}  & 49.89          & 96.63 & 82.73 & 79.20 & 39.12 & 81.31 & 71.48 \\
                       & 0.6 & \multicolumn{1}{c|}{0.05} & 50.44          & 96.43 & 82.99 & 79.80 & 39.94 & 80.83 & 71.74 \\
                       & 0.5 & \multicolumn{1}{c|}{0.1}  & 50.00          & 96.43 & 82.90 & 79.00 & 40.79 & 81.09 & 71.70 \\
                       & 0.5 & \multicolumn{1}{c|}{0.05} & 50.78          & 96.48 & 83.16 & 80.00 & 40.13 & 80.99 & 71.92 \\
                       & 0.4 & \multicolumn{1}{c|}{0.1}  & 50.22          & 96.43 & 82.99 & 79.00 & 39.56 & 81.06 & 71.54 \\
                       & 0.4 & \multicolumn{1}{c|}{0.05} & 50.67          & 96.63 & 82.82 & 79.90 & 39.20 & 81.19 & 71.73 \\
                       & 0.3 & \multicolumn{1}{c|}{0.1}  & 50.78          & 96.38 & 83.08 & 79.10 & 39.69 & 80.51 & 71.59 \\
                       & 0.3 & \multicolumn{1}{c|}{0.05} & 50.67          & 96.38 & 82.82 & 79.70 & 39.82 & 80.99 & 71.73 \\* \midrule
\multirow{39}{*}{25\%} & 1   & \multicolumn{1}{c|}{0.25} & 50.00          & 95.19 & 80.33 & 70.20 & 39.23 & 79.27 & 69.04 \\
                       & 1   & \multicolumn{1}{c|}{0.2}  & 49.89          & 95.24 & 81.53 & 71.40 & 38.96 & 79.50 & 69.42 \\
                       & 1   & \multicolumn{1}{c|}{0.15} & 51.00          & 95.74 & 82.13 & 72.70 & 38.61 & 79.18 & 69.89 \\
                       & 1   & \multicolumn{1}{c|}{0.1}  & 50.44          & 95.09 & 81.62 & 73.10 & 39.35 & 79.08 & 69.78 \\
                       & 1   & \multicolumn{1}{c|}{0.05} & 50.78          & 95.34 & 81.96 & 74.50 & 39.18 & 79.31 & 70.18 \\
                       & 0.9 & \multicolumn{1}{c|}{0.25} & 50.00          & 95.09 & 80.93 & 73.50 & 38.36 & 78.95 & 69.47 \\
                       & 0.9 & \multicolumn{1}{c|}{0.2}  & 50.33          & 95.19 & 81.96 & 73.90 & 39.14 & 79.08 & 69.93 \\
                       & 0.9 & \multicolumn{1}{c|}{0.15} & 50.00          & 95.29 & 81.53 & 74.50 & 39.02 & 78.98 & 69.89 \\
                       & 0.9 & \multicolumn{1}{c|}{0.1}  & 50.33          & 95.59 & 81.96 & 74.20 & 39.02 & 78.85 & 69.99 \\
                       & 0.9 & \multicolumn{1}{c|}{0.05} & 50.33          & 95.59 & 82.22 & 74.90 & 38.84 & 78.95 & 70.14 \\
                       & 0.8 & \multicolumn{1}{c|}{0.25} & 51.00          & 95.24 & 80.84 & 73.00 & 38.83 & 78.85 & 69.63 \\
                       & 0.8 & \multicolumn{1}{c|}{0.2}  & 51.11          & 95.29 & 82.04 & 74.10 & 38.36 & 79.08      & 70.00 \\
                       & 0.8 & \multicolumn{1}{c|}{0.15} & 49.78          & 95.49 & 82.04 & 74.20 & 38.89 & 78.95 & 69.89 \\
                       & 0.8 & \multicolumn{1}{c|}{0.1}  & 50.67          & 95.79 & 81.79 & 73.80 & 39.71 & 79.40 & 70.19 \\
                       & 0.8 & \multicolumn{1}{c|}{0.05} & 50.56          & 95.49 & 82.04 & 73.70 & 39.06 & 78.82 & 69.95 \\
                       & 0.7 & \multicolumn{1}{c|}{0.25} & 50.44          & 95.14 & 81.19 & 73.60 & 38.61 & 78.95 & 69.66 \\
                       & 0.7 & \multicolumn{1}{c|}{0.2}  & 50.56          & 95.34 & 82.04 & 74.70 & 40.11 & 79.15 & 70.32 \\
                       & 0.7 & \multicolumn{1}{c|}{0.15} & 49.89          & 95.74 & 82.13 & 74.10 & 38.92 & 79.21 & 70.00 \\
                       & 0.7 & \multicolumn{1}{c|}{0.1}  & 49.78          & 95.44 & 81.79 & 74.10 & 39.09 & 79.18 & 69.90 \\
                       & 0.7 & \multicolumn{1}{c|}{0.05} & 49.56          & 95.49 & 82.04 & 74.30 & 38.76 & 78.98 & 69.86 \\
                       & 0.6 & \multicolumn{1}{c|}{0.25} & 49.89          & 95.14 & 81.27 & 73.70 & 38.51 & 78.76 & 69.54 \\
                       & 0.6 & \multicolumn{1}{c|}{0.2}  & 49.89          & 95.49 & 81.44 & 74.10 & 39.30 & 78.76 & 69.83 \\
                       & 0.6 & \multicolumn{1}{c|}{0.15} & 50.44          & 95.29 & 81.53 & 74.40 & 39.48 & 78.98 & 70.02 \\
                       & 0.6 & \multicolumn{1}{c|}{0.1}  & 49.78          & 95.24 & 82.13 & 73.60 & 39.19 & 79.27 & 69.87 \\
                       & 0.5 & \multicolumn{1}{c|}{0.25} & 50.33          & 95.34 & 81.79 & 73.10 & 38.61 & 78.82 & 69.67 \\
                       & 0.5 & \multicolumn{1}{c|}{0.2}  & 50.11          & 95.69 & 81.36 & 74.90 & 39.46 & 79.15 & 70.11 \\
                       & 0.5 & \multicolumn{1}{c|}{0.15} & 49.78          & 95.59 & 82.39 & 74.10 & 39.29 & 78.98 & 70.02 \\
                       & 0.5 & \multicolumn{1}{c|}{0.1}  & 50.11          & 95.39 & 82.13 & 73.80 & 38.66 & 79.05 & 69.86 \\
                       & 0.5 & \multicolumn{1}{c|}{0.05} & 50.78          & 95.44 & 81.79 & 73.70 & 39.16 & 78.85 & 69.95 \\
                       & 0.4 & \multicolumn{1}{c|}{0.25} & 49.78          & 95.19 & 81.10 & 73.20 & 38.63 & 79.08 & 69.50 \\
                       & 0.4 & \multicolumn{1}{c|}{0.2}  & 50.67          & 95.39 & 81.79 & 74.80 & 38.70 & 78.95 & 70.05 \\
                       & 0.4 & \multicolumn{1}{c|}{0.15} & 50.22          & 95.24 & 82.39 & 74.20 & 38.44 & 79.18 & 69.95 \\
                       & 0.4 & \multicolumn{1}{c|}{0.1}  & 49.89          & 95.39 & 82.13 & 73.60 & 38.87 & 79.15 & 69.84 \\
                       & 0.4 & \multicolumn{1}{c|}{0.05} & 50.78          & 95.79 & 81.70 & 74.30 & 39.36 & 79.08 & 70.17 \\
                       & 0.3 & \multicolumn{1}{c|}{0.25} & 49.89          & 95.19 & 81.53 & 74.40 & 38.41 & 78.82 & 69.71 \\
                       & 0.3 & \multicolumn{1}{c|}{0.2}  & 50.22          & 95.44 & 81.70 & 74.50 & 38.90 & 79.08 & 69.97 \\
                       & 0.3 & \multicolumn{1}{c|}{0.15} & 50.00          & 95.44 & 81.87 & 74.30 & 38.81 & 79.08 & 69.92 \\
                       & 0.3 & \multicolumn{1}{c|}{0.1}  & 49.78          & 95.39 & 82.04 & 74.10 & 38.94 & 79.02 & 69.88 \\
                       & 0.3 & \multicolumn{1}{c|}{0.05} & 50.00          & 95.54 & 82.04 & 74.10 & 38.96 & 79.05 & 69.95 \\* \bottomrule
\end{longtable}

\end{document}